\documentclass[a4paper,fleqn]{cas-sc}

\usepackage[numbers]{natbib}
\usepackage{listings}
\usepackage[shortlabels]{enumitem}
\usepackage{siunitx}
\usepackage{subcaption}
\usepackage[ruled,vlined]{algorithm2e}
\usepackage{amsmath}
\usepackage{amssymb}
\usepackage{bm}
\def\tsc#1{\csdef{#1}{\textsc{\lowercase{#1}}\xspace}}
\tsc{WGM}
\tsc{QE}
\tsc{EP}
\tsc{PMS}
\tsc{BEC}
\tsc{DE}
\newcommand{\red}{\color{red}}

\newcommand{\vu}{\mathbf{u}}

\newcommand{\vsigma}{\boldsymbol{\sigma}}
\newcommand{\vtau}{\boldsymbol{\tau}}

\definecolor{arsenic}{rgb}{0.23, 0.27, 0.29}
\definecolor{charcoal}{rgb}{0.21, 0.27, 0.31}
\definecolor{hanblue}{rgb}{0.27, 0.42, 0.81}
\definecolor{blue-ncs}{rgb}{0.0, 0.53, 0.74}
\definecolor{awesome}{rgb}{1.0, 0.13,0.32}
\definecolor{darkgreen}{rgb}{0, .4,0}

\newcommand{\be}{\begin{equation}}
\newcommand{\ee}{\end{equation}}
\newcommand{\bd}{\begin{displaymath}}
\newcommand{\ed}{\end{displaymath}}
\newcommand{\BE}{\begin{eqnarray}}
\newcommand{\EE}{\end{eqnarray}}

\def\c#1{\textcolor{blue}{#1}}

\begin{document}
\let\WriteBookmarks\relax
\def\floatpagepagefraction{1}
\def\textpagefraction{.001}
\shorttitle{A Physics-Based Continuum Model for Versatile, Scalable, and Fast Terramechanics Simulation}
\shortauthors{Huzaifa Mustafa Unjhawala et~al.}
%\begin{frontmatter}

\title [mode = title]{A Physics-Based Continuum Model for Versatile, Scalable, and Fast Terramechanics Simulation}

% \tnotetext[1]{This document is the results of the research
%    project funded by the National Science Foundation.}

% \tnotetext[2]{The second title footnote which is a longer text matter
%    to fill through the whole text width and overflow into
%    another line in the footnotes area of the first page.}

\author{Huzaifa Mustafa Unjhawala}[orcid=0009-0004-4273-1212]\cormark[1]\ead{unjhawala@wisc.edu}
\author{Luning Bakke}[orcid=0000-0002-9650-6959]\ead{lfang9@wisc.edu}
\author{Harry Zhang}[orcid=0009-0003-1903-274X]\ead{hzhang699@wisc.edu}
\author{Michael Taylor}[orcid=0000-0002-6416-3154]\ead{mtaylor22@wisc.edu}
\author{Ganesh Arivoli}[orcid=0009-0004-9385-4271]\ead{arivoli@wisc.edu}
\author{Radu Serban}[orcid=0000-0002-4219-905X]\ead{serban@wisc.edu}
\author{Dan Negrut}[orcid=0000-0003-1565-2784]\ead{negrut@wisc.edu}

\affiliation[]{organization={Department of Mechanical Engineering, University of Wisconsin-Madison},
	addressline={2107, 1513 University Ave}, 
	city={Madison},
	%               citysep={}, % Uncomment if no comma needed between city and postcode
	postcode={53706}, 
	state={Wisconsin},
	country={United States}}

\cortext[cor1]{Corresponding author}

\begin{abstract}
This paper discusses Chrono's Continuous Representation Model (called herein Chrono::CRM), a general-purpose, scalable, and efficient simulation solution for terramechanics problems. Built on Chrono's Smoothed Particle Hydrodynamics (SPH) framework, Chrono::CRM moves beyond semi-empirical terramechanics approaches, e.g., Bekker-Wong/Janosi-Hanamoto, to provide a physics-based model able to address complex tasks such as digging, grading, as well as interaction with deformable wheels and complex grouser/lug patterns. The terramechanics model is versatile in that it allows the terrain to interact with both rigid and flexible implements simulated via the Chrono dynamics engine. We validate Chrono::CRM against experimental data from three physical tests, including one involving NASA's MGRU3 rover. In addition, the simulator is benchmarked against a high-fidelity Discrete Element Method (DEM) simulation of a digging scenario involving the Regolith Advanced Surface Systems Operations Robot (RASSOR). Being GPU-accelerated, Chrono::CRM achieves computational efficiency comparable to that of semi-empirical simulation approaches for terramechanics problems. Through an ``active domains'' implementation, Chrono::CRM can handle terrain stretches up to \SI{10}{km} long with 100 million SPH particles at near interactive rates, making high-fidelity off-road simulations at large scales feasible. As a component of the Chrono package, the CRM model is open source and released under a BSD-3 license. All models and simulations used in this contribution are available in a public GitHub repository for reproducibility studies and further research.
\end{abstract}


\maketitle

\section{Introduction}
\label{sec:intro}
\subsection{Simulation in engineering}
Simulation has become an increasingly integral tool in the engineering of vehicles and robots~\cite{Johnson2023SimulationFidelity}.
As noted by Choi et al.~\cite{PNASsimRobotics2021}, in robotics applications, simulation can shorten development cycles, enhance safety, and lead to better designs.
Moreover, simulation serves as an effective mechanism for generating data to train and test AI-driven controllers. It also allows researchers to create synthetic datasets, thereby reducing the dependence on extensive real-world data acquisition while maintaining learned model performance~\cite{softserve2024synthetic}.
Simulation offers a systematic methodology for investigating complex physical interactions that are challenging or impractical to study through traditional experimentation.
For extra-terrestrial design projects, in particular, simulation is often the primary feasible means of assessing system performance before construction and deployment.
This is particularly relevant when addressing deformable terrain interactions on extraterrestrial surfaces, where conditions such as reduced gravity and distinct material properties are not readily replicable on Earth~\cite{weiGravOffset2025}.

\subsection{Existing Simulation Solutions}
Simulators for off-road vehicle and robot design require a balance between \textit{accuracy} and \textit{speed}. Accuracy ensures design decisions reached in simulation yield quality real-world solutions, while speed permits rapid iteration and short development cycles. Achieving a balance is usually challenging --- high-accuracy modeling, which uses high-fidelity physics to capture rigid and flexible multi-body interactions with granular media, can yield transferable results but incurs significant computational costs.

Many existing simulation solutions compromise on one requirement or the other. For example, Chrono's Soil Contact Model (Chrono::SCM)~\cite{chronoOverview2016,chronoSCM_JCND_2023} uses semi-empirical theories like Bekker-Wong/Janosi-Hanamoto, allowing for fast performance but lacking accuracy and requiring extensive experimental calibration~\cite{weiVirtualBevameter2024}. Moreover, SCM does not capture certain complex soil interaction and therefore cannot support tasks such as digging or grading. Similarly, popular robotics simulators such as Gazebo~\cite{KoenigDesign2004}, Carla~\cite{carlaAVsim2017}, AirSim~\cite{airsim2018}, Webots~\cite{webots2004}, and Isaac Sim~\cite{IssacNvidia2023}, while performant and proven invaluable for developing and validating autonomy stacks in a wide range of robots~\cite{bousmalis2018using,farchyStone2013,duriez2018,bewley2019learning,Xuemin2023Sim2RealSurvey,voogd2023reinforcementpathfollowing,HarryIROSPathFollowing}, have limited, if any, terramechanics support and thus are less relevant for off-road applications.

On the other hand, Discrete Element Method (DEM)-based simulators, such as the Chrono Discrete Element Method Engine (DEME)~\cite{ruochunGRC-DEM2023}, resolve each grain of granular material individually, achieving high fidelity at the expense of computational cost. For instance, simulating three seconds of dynamics for a problem involving 10 million grains can take up to seven hours on two NVIDIA A100 GPUs~\cite{ruochunGRC-DEM2023}. GranularGym~\cite{millard2023granulargym} attempts to optimize DEM for near-real-time performance, but its benchmarks with \num{100000} particles and a relatively large time step ($1\times 10^{-3}\,\mathrm{s}$) raise questions about simulation accuracy and scalability. In addition, GranularGym is written in Julia, complicating integration with typical multi-body dynamics and sensor simulation frameworks.

Continuum models for granular media offer a middle ground between accuracy and speed. These models capture macro-scale behavior by relating stress and strain-rate fields, which can be discretized using a hybrid Eulerian-Lagrangian approach (Material Point Method, MPM)~\cite{bardenhagen2000material} or a purely Lagrangian approach (Smoothed Particle Hydrodynamics, SPH)~\cite{bui08lagrangian,weiGranularSPH2021}. Since the discretization is not tied to individual grain size, continuum models typically have a much lower degree of freedom count and thus run more efficiently than DEM simulations. Taichi MPM~\cite{hu2019taichi} is a popular GPU-accelerated MPM solver, but it is standalone and has not been validated extensively against experimental data or tested for large-scale, real-world simulations.

\subsection{Contributions}
This work answers the need for a general simulation framework that can handle a wide range of robot-terrain interactions—especially in off-road environments—while still delivering the speed needed for large-scale scenarios. Striking this balance between accuracy and computational efficiency is critical for advancing both vehicle design and robotic autonomy.
To that end, we introduce Chrono::CRM, a physics-based, versatile, fast, and scalable simulation framework built on top of Chrono's existing Fluid-Solid Interaction (FSI) and SPH framework for modeling granular media as a continuum~\cite{RaduFSI2025}. Our main contributions are:

\begin{itemize}
    \item We validate the SPH-based continuum approach against a suite of experimental tests, including system-level rover trials. We also compare rover digging scenarios, using a Regolith Advanced Surface Systems Operations Robot (RASSOR) drum~\cite{RASSOR2013}, against high-fidelity DEM simulations, highlighting our the versatility of the proposed approach in various off-road tasks.

    \item We achieve a 3$\times$ speedup through optimized neighbor lists construction and update frequency control, enabling near real-time simulations for vehicles and robots. To illustrate these gains, we present five scenarios that include rigid and flexible bodies, tracked and wheeled vehicles, and a full-scale rover.

    \item We introduce an ``active domains'' technique that supports off-road terrain traverses of up to \SI{6}{km} on consumer-grade hardware and up to \SI{29}{km} on NVIDIA H100 GPUs. This renders Chrono::CRM suitable for large-scale, real-world applications. Additionally, we show that using ``active domains'' where applicable leads to no loss in accuracy while delivering up to 10$\times$ speedups.

    \item The close integration of Chrono::CRM with the Chrono multi-body dynamics library is demonstrated through two case studies:
    \begin{enumerate}
        \item A performance comparison of a Polaris RZR traversing a \SI{50}{m} course. This study evaluates scenarios involving both rigid and deformable terrain, combined with the use of rigid tires and deformable tires (the latter modeled using the Absolute Nodal Coordinate Formulation, ANCF~\cite{mikeANCF-implementationAspects2023}).
        \item Simulation of an autonomous Gator, equipped with a blade, undertaking a leveling task on a hilly, deformable terrain patch. The vehicle's control policy for this task was trained within a simulation environment using Chrono::CRM.
    \end{enumerate}
\end{itemize}

\subsection{Organization}
The paper is organized as follows. In Sec.~\ref{sec:method} we discuss details about the continuum model for granular media, the SPH solver, and the novel FSI framework that couples the Chrono::CRM with the Chrono multi-body dynamics library. Then, in Sec.~\ref{sec:validation}, we validate the framework against a suite of experimental tests showing the generality and accuracy of the software. In Sec.~\ref{sec:performance} we illustrate the computational efficiency and resulting performance gains of the proposed solution methodology, based on a direct comparison with semi-empirical approaches. The scalability of the framework is discussed in Sec.~\ref{sec:scalability} and the two case studies are demonstrated in Sec.~\ref{sec:demOfTech}. We conclude the paper with a discussion of the results and directions for future work in Sec.~\ref{sec:conclusion}.

\section{Methodology}\label{sec:method}
The granular material is modeled using the same approach as described in~\cite{KamrinFluidMechanics2015,weiGranularSPH2021}, baring minor differences and additional features. For completeness, we describe the entire model here and point out the key differences.

%===============================================================

\subsection{Equations of Motion for the Granular Material}\label{subsec:equationsOfMotionGranularMaterial}
\noindent In Chrono::CRM, discrete granular flows are treated as a continuum ``fluid'' and granular flow interactions with solids is posed as an FSI problem (see Sec.~\ref{subsec:FSI}). The density $\rho$ and the ``fluid'' velocity $\vu$ are governed by the continuity and momentum balance equations which, in Lagrangian form, are given by:
\begin{align}
\label{eq:CRM_const}
\frac{d\rho}{dt} &= -\rho \nabla \cdot \mathbf{u},\\
\label{eq:CRM_eq}
\frac{d\mathbf{u}}{dt} &= \frac{\nabla \cdot \vsigma}{\rho} + \mathbf{f}_b,
\end{align}
where $\mathbf{f}_b$ denotes the external force per unit mass. 
The stress can be decomposed into a pressure term $p \coloneq -\mathrm{tr}(\vsigma)/3$ and a deviatoric stress tensor $\vtau$, so that $\vsigma = -p\mathbf{I} + \vtau$. 

For materials undergoing finite deformation, the stress rate must be expressed in an objective (frame-independent) form. One commonly used objective stress rate is the \textit{Jaumann rate}. Under the assumption of linear hypoelasticity, a generalization of Hooke’s law provides a linear relation between the Jaumann stress rate tensor and the elastic strain rate tensor:
\begin{equation}\label{equ:stress_rate_hypoelastic}
\overset{\vartriangle}{\bm{\sigma}} = 2G \left( \dot{\bm{\varepsilon}} - \frac{1}{3} \text{tr}(\dot{\bm{\varepsilon}}) \bm{I} \right) + K \, \text{tr}(\dot{\bm{\varepsilon}}) \bm{I},
\end{equation}
where \( G \) and \( K \) are the shear and bulk moduli, respectively, \( \dot{\bm{\varepsilon}} \) is the rate of deformation tensor (symmetric part of the velocity gradient), and \( \bm{I} \) is the second-order identity tensor.
The Jaumann rate of the Cauchy stress tensor is also defined as:
\begin{equation}\label{equ:stress_rate_jaumann}
\overset{\vartriangle}{\bm{\sigma}} = \frac{d\bm{\sigma}}{dt} - \dot{\bm{\phi}} \cdot \bm{\sigma} + \bm{\sigma} \cdot \dot{\bm{\phi}},
\end{equation}
where \( \dot{\bm{\phi}} = \frac{1}{2}(\nabla \mathbf{u} - (\nabla \mathbf{u})^\mathrm{T}) \) is the spin tensor (also called the rotation rate tensor). Using the material constitutive relation of linear hypoelasticity (Eq.~\ref{equ:stress_rate_hypoelastic}) and the definition of the Jaumann rate (Eq.~\ref{equ:stress_rate_jaumann}), we can write the material time derivative of the Cauchy stress tensor as:
\begin{equation}\label{equ:stress_rate}
\frac{d\bm{\sigma}}{dt} = \dot{\bm{\phi}} \cdot \bm{\sigma} - \bm{\sigma} \cdot \dot{\bm{\phi}} + \overset{\vartriangle}{\bm{\sigma}}.
\end{equation}

The elastic strain rate tensor $\dot{\boldsymbol{\varepsilon}}$ is evaluated by taking the total strain rate and subtracting the plastic strain rate due to flow:
\begin{equation}\label{equ:strain_rate}
\dot{\boldsymbol{\varepsilon}} = 
\underbrace{\frac{1}{2}(\nabla\textbf{u} + \nabla\textbf{u}^\intercal)}_{\text{total}} 
- 
\underbrace{\frac{1}{\sqrt{2}}\dot{\lambda} \frac{\boldsymbol{\tau}}{\bar{\tau}}}_{\text{plastic}} \; ,
\end{equation}
$\dot{\lambda}$ and $\bar{\tau}$ are the plastic strain rate and equivalent shear stress, respectively. Thus, once the phase of the granular material is determined, the stress tensor can be computed using Eq.~\ref{equ:stress_rate} and the density and velocity can be updated using Eqs.~\ref{eq:CRM_const} and~\ref{eq:CRM_eq}. We discuss the computation of $\dot{\lambda}$ and $\bar{\tau}$ in Sec.~\ref{subsec:PlasticFlow}.

%===============================================================

\subsection{SPH Discretization of the Equations of Motion}\label{subsec:SPHDiscretizationGranularMaterial}
The Partial Differential Equations (PDEs) in Eqs.~\ref{eq:CRM_const}, \ref{eq:CRM_eq}, and \ref{equ:stress_rate} are discretized in space using the SPH method, in which the domain of interest is represented by particles that move (or ``advect'') with the flow field. The SPH spatial discretization is based on two main approximations: the kernel approximation, and the particle approximation~\cite{Monaghan2005a, bui2021smoothed, fatehi2011}. For an arbitrary field function $f$, the interpolated value $\langle f(\mathbf{r})\rangle$ can be written in terms of its values at neighboring particles, $f_j$, as
\begin{equation}
\langle f(\mathbf{r})\rangle = \sum_{j=1}^{N} f_j \, W(\mathbf{r} - \mathbf{r}_j, h)\,\mathcal{V}_j \; ,
\end{equation}
where $\mathcal{V}_j$ is the volume associated with particle $j$, and $W$ is the smoothing (kernel) function, which approximates the Dirac delta and has compact support for $|\mathbf{r} - \mathbf{r}_j| < h$. 

To obtain the first-order spatial derivatives required by Eqs.~\ref{eq:CRM_const},~\ref{eq:CRM_eq}, and~\ref{equ:stress_rate}, several discretization schemes have been proposed~\cite{fatehi2011}:
\begin{description}
\item{F1} (Monaghan's Standard Scheme~\cite{Monaghan2005a}):
$\langle \nabla f \rangle_i = \sum_j \mathcal{V}_j f_j \nabla_i W_{ij}$;
\item{F2} (Zero-Order Consistent Scheme):
$\langle \nabla f \rangle_i = \sum_j \mathcal{V}_j (f_j - f_i)\nabla_i W_{ij}$; and
\item{F3} (Symmetric Form, ensuring local momentum conservation~\cite{fatehi2011}):
$\langle \nabla f \rangle_i = \sum_j \mathcal{V}_j (f_j + f_i)\nabla_i W_{ij}$.
\end{description}

In Chrono::CRM, we discretize Eqs.~\ref{eq:CRM_const} and \ref{equ:stress_rate} using the scheme F2 for zero-order consistency. However, unlike~\cite{weiGranularSPH2021}, we discretize Eq.~\ref{eq:CRM_eq} using scheme F3 to enforce local momentum conservation. The discretized form of Eqs.~\ref{eq:CRM_const}, \ref{eq:CRM_eq}, and~\ref{equ:stress_rate} is given by:
\begin{align}
    \frac{d\rho_i}{dt} &= -\rho_i  \sum\limits_j (\mathbf{u}_j- \mathbf{u}_i) \cdot \nabla_i W_{ij}\,\mathcal{V}_{j} \label{equ:continuity_dis} \\
    \frac{d\mathbf{u}_i}{dt} &= \frac{1}{\rho_i} \sum\limits_j (\boldsymbol{\sigma}_j + \boldsymbol{\sigma}_i) \cdot \nabla_i W_{ij} \,\mathcal{V}_{j} + \mathbf{f}_{b,i} + \Pi_a \label{equ:momentum_dis} \\
    \frac{d\boldsymbol{\sigma}_i}{dt} &= 
    \begin{aligned}[t]
        &\frac{1}{2}\left\{
        \left [\sum\limits_j (\mathbf{u}_{ji} - \mathbf{u}_{ji}^\intercal)  \nabla_i W_{ij}\,\mathcal{V}_{j} \right ]\boldsymbol{\sigma}_i
        -\boldsymbol{\sigma}_i  \left [\sum\limits_j (\mathbf{u}_{ji} - \mathbf{u}_{ji}^\intercal)  \nabla_i W_{ij}\,\mathcal{V}_{j}\right]
        \right\} \\
        &+G\left\{
        \left[ \sum\limits_j (\mathbf{u}_{ji} + \mathbf{u}_{ji}^\intercal)  \nabla_i W_{ij}\,\mathcal{V}_{j}\right]-\frac{1}{3}\textrm{tr}\left( \sum\limits_j (\mathbf{u}_{ji} + \mathbf{u}_{ji}^{\mathrm{T}})  \nabla_i W_{ij}\,\mathcal{V}_{j}\right)\mathbf{I}
        \right\} \\
        &+\frac{1}{6}K\left\{
        \textrm{tr}\left( \sum\limits_j (\mathbf{u}_{ji} + \mathbf{u}_{ji}^\intercal)  \nabla_i W_{ij}\,\mathcal{V}_{j}\right)\mathbf{I}
        \right\} \; ,
    \end{aligned}
    \label{equ:stress_rate_dis}
\end{align}
where $\mathbf{u}_{ji} = \mathbf{u}_j - \mathbf{u}_i$. Note that in Eq.~\ref{equ:stress_rate_dis}, $\dot{\boldsymbol{\varepsilon}}$ is expressed using the total strain rate tensor as $\dot{\boldsymbol{\varepsilon}} = \frac{1}{2}(\nabla\textbf{u} + \nabla\textbf{u}^\intercal)$. This is because when plastic flow is present, the predicted stress tensor is corrected after the integration step as in~\cite{weiGranularSPH2021} (see Sec.~\ref{subsec:PlasticFlow} for more details).
Additionally, we add to the momentum equation an ``artificial viscosity'' term ($\Pi_a$) that is commonly used in SPH codes for numerical stability reasons~\cite{Monaghan1983,crespo2015dualsphysics}. The artificial viscosity force term is given by:
\begin{equation}
  \label{eq:artificial_viscosity_bilateral}
   \Pi_a = -\gamma_a h \sum_j \frac{m_j}{\bar{\rho}_{ij}} c_s \frac{\mathbf{v}_{ij} \cdot \mathbf{r}_{ij}}{r_{ij}^2 + \xi^2} \nabla_i W_{ij} \; .
\end{equation}
Above, $\gamma_a$ is the empirically determined artificial viscosity parameter, $h$ is the smoothing length, and $c_s = \sqrt{K/\rho}$ is the speed of sound in the granular material.
The artificial viscosity force term thus causes close approaching particles to repel each other with a force which increases with the approach velocity. Additionally, it also causes particles moving apart from each other to slow down, thus maintaining a uniform particle distribution resulting in a more stable simulation. Some SPH codes only allow particles approaching each other to mutually repel by modifying the artificial viscosity force term as
\begin{equation}
  \label{eq:artificial_viscosity_unilateral}
\Pi_{ab} = \begin{cases} -\gamma_a h \sum_j \frac{m_j}{\bar{\rho}_{ij}} c_s \frac{\mathbf{v}_{ij} \cdot \mathbf{r}_{ij}}{r_{ij}^2 + \xi^2} \nabla_i W_{ij} & \mathbf{v}_{ij} \cdot \mathbf{r}_{ij} < 0; \\ 0 & \mathbf{v}_{ij} \cdot \mathbf{r}_{ij} \ge 0; \end{cases}
\end{equation}
We provisioned for both approaches in Chrono::CRM and use them both in our validation studies in Sec.~\ref{sec:validation}. We henceforth refer to Eq.~\ref{eq:artificial_viscosity_bilateral} as the ``bilateral'' approach and Eq.~\ref{eq:artificial_viscosity_unilateral} as the ``unilateral'' approach.

% ===========================================

\subsection{Time Integration: Second-Order Runge--Kutta Scheme}\label{subsec:TimeIntegration}
Collecting the particle positions, velocities, and stress tensor in a system state vector $\mathbf{y} = \left[ \mathbf{x} , \mathbf{u} , \mathbf{\sigma} \right]$, the semi-discretized equations of motion can be written as a first-order Ordinary Differential Equation (ODE) system with right-hand side $\mathbf{f}(t, \mathbf{y}) = \left[  \mathbf{u} , \mathbf{a} , \mathbf{g} \right]$, with the particle accelerations and stress-rates given by Eqs.~\ref{equ:momentum_dis} and \ref{equ:stress_rate_dis}, respectively.

We integrate this ODE using a two-stage Runge-Kutta (RK2) scheme, namely the explicit midpoint method, which provides second-order temporal accuracy. The scheme advances the state $\mathbf{y}_{n}$ at time $t_n$ using:
\begin{equation}
\mathbf{y}_{n+1} = \mathbf{y}_n + h \mathbf{f} \left(
t_n + \frac{1}{2}h , \mathbf{y}_n + \frac{1}{2}hf(t_n, \mathbf{y}_n)
\right) \; , 
\end{equation}
where $h = t_{n+1} - t_n$ is the integration step size. This RK2 method requires two evaluations of $\mathbf{a}$ and $\mathbf{g}$, at the current time and at the midpoint.

\subsection{The prediction-correction step of the stress tensor to incorporate plastic flow}
\label{subsec:PlasticFlow}
To represent the many phases of soil flow, we employ a rheology proposed in conjunction with the material point method (MPM)~\cite{KamrinFluidMechanics2015,weiGranularSPH2021} with a modification to incorporate a cohesion like behavior when desired. 
The RK2 integrator of Sec.~\ref{subsec:TimeIntegration} advances the stress tensor from $t_n$ to an \textit{elastic trial} value $\boldsymbol{\sigma}^\ast=\boldsymbol{\sigma}(t_{n+1})$, where we consider only the elastic strain rate.
From $\boldsymbol{\sigma}^\ast$ we extract the pressure and the deviatoric stress:
\begin{equation}
p^\ast = -\tfrac13\,\mathrm{tr}\,\boldsymbol{\sigma}^\ast, 
\qquad
\boldsymbol{\tau}^\ast = \boldsymbol{\sigma}^\ast + p^\ast\mathbf{I},
\qquad
\bar\tau^\ast
   = \sqrt{\tfrac12 \,\boldsymbol{\tau}^\ast\!:\!\boldsymbol{\tau}^\ast } .
\label{eq:trial_quantities}
\end{equation}

The final stress $\boldsymbol{\sigma}^{\,n+1}$ is obtained with the four-step return-mapping procedure below, which augments the original method of~\cite{KamrinFluidMechanics2015} with a cohesive intercept~$c$.
\begin{description}

\item{\bf{Step 1}}.
The Mohr-Coulomb envelope with cohesion, $\tau_{\max}= \mu_{\mathrm{s}}\,p + c$, intersects the $p$‑axis at the \emph{critical pressure}
\begin{equation}
p_{\mathrm{cri}}=-\dfrac{c}{\mu_{\mathrm{s}}}.
\label{eq:pcrit}
\end{equation}
\noindent If the trial state lies to the left of this cut-off ($p^\ast < p_{\mathrm{cri}}$), the soil is essentially being pulled more than the cohesive strength and thus the material cannot carry any shear load, and we therefore correct the trial stress tensor as
\begin{equation}
\boldsymbol{\sigma}^{\,n+1}=\mathbf{0}
\quad\text{and advance to the next time step}.
\end{equation}

\item{\bf{Step 2}}.
For $p^\ast\ge p_{\mathrm{cri}}$ we evaluate the $\mu(I)$-rheology\,\cite{KamrinFluidMechanics2015}
\begin{equation}
\mu(I)=\mu_{\mathrm{s}}
        +\frac{\mu_2-\mu_{\mathrm{s}}}{1 + I_0/I},
\qquad
I = \dot{\gamma}\,d\sqrt{\frac{\rho_0}{p^\ast}}, 
\qquad
\dot{\gamma}= \frac{\bar\tau^\ast-\bar\tau^{\,n}}{G\,\Delta t},
\label{eq:muI}
\end{equation}
where $d$ is the particle diameter, $\rho_0$ the reference density, $G$ the elastic shear modulus, and the constants $(\mu_{\mathrm{s}},\mu_{2},I_0)$ are material parameters.

\item{\bf{Step 3}}.
The cohesive $\mu(I)$ criterion reads
\begin{equation}
\tau_{\max}= \mu(I)\,p^\ast + c .
\label{eq:yieldsurf}
\end{equation}
If $\bar\tau^\ast\le \tau_{\max}$, the trial stress is admissible and we simply keep it:
\[
\boldsymbol{\sigma}^{\,n+1}= \boldsymbol{\sigma}^\ast.
\]

\item{\bf{Step 4}}.
Otherwise ($\bar\tau^\ast>\tau_{\max}$) plastic flow occurs. We correct the stress via a radial return in deviatoric space:
\begin{subequations}
\begin{align}
\boldsymbol{\tau}^{\,n+1}
 &= \frac{\tau_{\max}}{\bar\tau^\ast}\,\boldsymbol{\tau}^\ast,
\\[2pt]
p^{\,n+1} &= p^\ast,
\qquad
\boldsymbol{\sigma}^{\,n+1}
   = -p^{\,n+1}\mathbf{I} + \boldsymbol{\tau}^{\,n+1}.
\end{align}
\end{subequations}

\end{description}

% =====================================================================

\subsection{Fluid-Solid Interaction Framework}\label{subsec:FSI}
Coupling of Chrono::CRM with a Chrono multibody system is done in an explicit force-displacement co-simulation framework implement through the Chrono::FSI framework for fluid-solid interaction~\cite{RaduFSI2025}. A CRM simulation is thus set up as an FSI system which manages the Chrono multibody system (modeling for example a vehicle), a Chrono::SPH fluid system (configured here to solve the CRM equations), and an FSI interface which controls the data exchange between the co-simulated systems for the solid and fluid phases, respectively. The Chrono multibody system communicates full state information of the solid phase (position and velocity level) and the fluid solver sends resultant fluid forces and torques acting on rigid bodies and on nodes of FEA meshes.  The FSI system controls the co-simulation meta-step (i.e., the frequency of inter-system data communication) and advances the dynamics of the two phases simultaneously and in parallel, using non-blocking concurrent threads. While the full architecture is described in detail in~\cite{RaduFSI2025}, here we focus on the Boundary Condition Enforcing (BCE) markers that intermediate the transfer of states and forces between the multibody and fluid solvers. 

%Chrono::CRM interfaces with Chrono's multibody dynamics engine through an abstract fluid-structure interaction (FSI) framework. An FSI system manages references to a Chrono multibody system, and a fluid solver, along with an abstract coupling interface that handles the data exchange between the two phases. Chrono::CRM serves as a concrete implementation of a ``fluid'' solver within this modular FSI framework. The FSI problem is solved in an explicit force-displacement co-simulation setting, with the Chrono system communicating full state information of the solid phase (position and velocity level) and the fluid solver sending fluid forces and torques acting on rigid bodies or on nodes of the FEA meshes. The FSI framework controls the co-simulation time stepping function, first invoking the data exchange functions through the interface and then advancing the dynamics of the two phases simultaneously and in parallel (using non-blocking concurrent threads). While the full architecture is described in detail in~\cite{RaduFSI2025}, here we focus on the Boundary Condition Enforcing (BCE) markers that help transfer state and force between the multibody and fluid solvers. 

\subsubsection{Boundary Condition Enforcing Markers}
In the Chrono::FSI framework, for the purpose of phase coupling, both rigid and flexible solids are represented by their collision geometry. In the case of rigid bodies, the collision model is a set of simple geometric primitives (spheres, boxes, cylinders, capsules, or cones) and, if needed, higher-resolution triangular surface meshes. The collision geometry of FEA meshes used to model flexible solids is a deformable surface with vertices at the FEA nodes, either a segment set for beam-like elements, or a triangular mesh for shell and solid finite elements. The Chrono::FEA module provides support for extraction of these generic collision surfaces from any FEA mesh (as the external envelope surface in case of solid finite elements), and represent them in data structures that are independent of the particular type of finite element used in modeling the flexible solid.

The interaction between particle-based fluids and solid bodies is managed through BCE markers~\cite{Adami2012,Holmes2011Smooth,weiGranularSPH2021}. These markers are distributed over the solid surfaces in distinct layers with an initial uniform spacing $d_0$ in all directions. As the solid body moves and/or deforms under the multibody dynamics simulation, the BCE markers move correspondingly, acting as specialized SPH particles whose kinematics are dictated by the solid phase. 
The BCE markers serve two main purposes: enforcing boundary conditions at the fluid-solid interface (such as the no-slip condition) and ensuring adequate kernel support for nearby fluid SPH particles. Because Chrono::CRM uses a single resolution scheme, the BCE markers share the same discretization spacing $d_0$ and kernel smoothing length $h$ as the fluid particles. To guarantee that all fluid particles whose kernel radius ($\mathcal{K}h$, with $\mathcal{K}=2.0$ in this study) intersects the solid surface interact with sufficient BCE markers, the required number of BCE marker layers is calculated as $N_{\text{layers}} = \left\lceil \frac{\mathcal{K}h}{d_0} \right\rceil$.

For rigid bodies, the first layer of BCE markers is placed on the collision surfaces with subsequent parallel layers inside the volume of the collision shapes. For flexible solids, BCE markers are similarly placed on and around the FEA collision geometry, with different options available for placement of subsequent BCE layers relative to the collision surface (see~\cite{RaduFSI2025}). While the BCE marker positions are fixed relative to their corresponding rigid body, the BCE markers associated with flexible solids follow the deformation of the corresponding finite elements.

BCE markers impose the no-slip boundary condition and facilitate force transfer between fluid and solid phases. The procedure involves several key steps. First, the state variables (e.g., velocity, stress) of the BCE markers are linearly extrapolated from neighboring fluid SPH particles. Chrono::CRM supports two methods for velocity extrapolation: Adami~\cite{Adami2012} and Holmes~\cite{Holmes2011Smooth}. For example, using the Adami method the extrapolated velocity at a BCE marker $a$ is computed as $\tilde{\mathbf{u}}_a = \left({\sum_b \mathbf{u}_b W_{ab}}\right)/\left({\sum_b W_{ab}}\right)$, where $b$ denotes nearby fluid SPH particles and $W_{ab}$ is the kernel weight function between marker $a$ and particle $b$. To enforce the no-slip boundary condition, the velocity at each BCE marker is corrected to $\mathbf{u}_a = 2\mathbf{u}_\text{body} - \tilde{\mathbf{u}}_a$, where $\mathbf{u}_\text{body}$ is the velocity of the solid body at the BCE marker location.

Similarly, the stress tensor is linearly extrapolated from the SPH particles as
\begin{equation*}
	\boldsymbol{\sigma}_s^{ij} = 
\frac{
  \sum_f \boldsymbol{\sigma}_f^{ij} W(\mathbf{x}_{sf}) \left( \mathbf{g}^i - \mathbf{a}_s^i \right)
  - \sum_f \boldsymbol{\rho}_f^i W(\mathbf{r}_{sf}) \delta^{ij}
}
{
  \sum_f W(\mathbf{x}_{sf})
} \, ,
\end{equation*} 
where subscripts $s$ and $f$ indicate solid BCE markers and SPH particles, respectively, and $\delta^{ij}$ is the Kronecker delta function~\cite{zhan2019three,weiGranularSPH2021}. This extrapolation ensures a balanced force distribution at the fluid-solid interface, effectively preventing SPH particles from penetrating solid surfaces defined by BCE markers~\cite{Adami2012,zhan2019three}.

The states evaluated at the BCE marker locations allow calculation of their accelerations via the momentum balance equation (Eq.~\ref{equ:momentum_dis}). These accelerations are subsequently used to calculate the resultant total force and torque acting on rigid bodies. The procedure for flexible bodies is analogous, with forces summed over BCE markers within each finite element and then distributed as nodal forces. These fluid forces and torques are then provided to the multibody system, while new position and velocity information received from the multibody system are used to update the kinematic state of the BCE markers. For more details on the BCE markers, see~\cite{RaduFSI2025}.

% given by: \(\textbf{F}_{body} = \sum\limits_{i\in \Omega_s} m_i \frac{d \mathbf{u}_i}{dt} \quad \text{and} \quad \mathbf{T}_{body} = \sum\limits_{i\in \Omega_s} \mathbf{r}_{i} \times (m_i \frac{d \mathbf{u}_i}{dt})\), where $i\in\Omega_s$ is the set of BCE markers and $ \frac{d \mathbf{u}_i}{dt}$ is the computed acceleration of the BCE marker $i$ and $m_i$ is the mass of the BCE marker $i$. 

% \begin{figure}[tbp]
%   \centering
%   \begin{minipage}[b]{0.45\textwidth}
%     \centering
%     \includegraphics[width=1.8in]{figs/bce_filledRigid.png}
%     \\[-1ex]
%     \small Filled rigid body
%   \end{minipage}%
%   \hspace{0.04\textwidth}%
%   \begin{minipage}[b]{0.45\textwidth}
%     \centering
%     \includegraphics[width=1.8in]{figs/bce_filledRigid.png} % Placeholder for second image
%     \\[-1ex]
%     \small Flexible bodies \SBELcomment{TODO: add image from Radu}
%   \end{minipage}
%   \caption{\SBELcomment{TODO: modify caption for flexible bodies too} SPH particles (red) and BCE markers (blue) near the ``fluid"-solid boundary of a filled rigid body. $\mathbf{u}_j$ is the extrapolated velocity on the BCE marker (evaluated either using Adami's method~\cite{Adami2012} or Holmes's method~\cite{Holmes2011Smooth}) so as to impose the no-slip boundary condition for SPH particles close to the solid surface.}
%   \label{fig:BCE_markers_filledRigid}
% \end{figure}

\section{Validation Studies}\label{sec:validation}
In three validation studies, we compare Chrono::CRM results against physical test data. In a fourth study, we compare the results of Chrono::CRM against DEM simulation results. These four validation studies are described below.

% ============================================

\subsection{Sphere Cratering test}
\label{sec:sphere_cratering}
In the sphere cratering test, we drop a sphere with radius $R_{\text{sphere}} = $ \SI{0.0125}{m} onto a granular medium characterized by a density $\rho =$ \SI{1510}{kg/m^3} and a static friction coefficient $\mu_s = 0.3$. We perform drops from three initial heights $H = \{ 0.05, \, 0.1, \, 0.2\}$ \SI{}{m}. Following the experiments by Uehara et al.~\cite{uehara2003low} and Ambroso et al.~\cite{ambroso2005penetration}, we repeat these drops using two different sphere densities, $\rho_{\text{sphere}} = $ \SI{700}{kg/m^3} and \SI{2200}{kg/m^3}. These experimental studies found that the sphere penetration depth, $D$, can be predicted by the empirical relationship:
\begin{equation}
\label{eq:ballDropEquation}
D = 0.14 \cdot \frac{1}{\mu_s}\left( \frac{\rho_{sphere}}{\rho_{granular}}\right)^{\frac{1}{2}}(2R_{sphere})^{\frac{2}{3}}H^{\frac{1}{3}}.
\end{equation}
Furthermore, Ambroso et al.~\cite{ambroso2005penetration} observed that different granular particle sizes resulted in identical crater shapes and penetration depths, suggesting particle size is not a significant factor in this test. Based on this finding, we use a consistent particle diameter of \SI{1}{mm} for the rheology model parameter $d$ (see Eq.~\ref{eq:muI}) throughout our simulations. The simulations are conducted within a container of size \SI{0.14}{m} $\times$ \SI{0.1}{m} $\times$ \SI{0.15}{m}. These dimensions are chosen to be sufficiently large to minimize boundary effects. The specific SPH parameters used are detailed in Table~\ref{tab:sph_params}.

\begin{figure}[htbp]
  \captionsetup[subfigure]{justification=centering}
  \centering
  \begin{subfigure}[t]{0.35\textwidth}
    \centering
    \includegraphics[width=\textwidth]{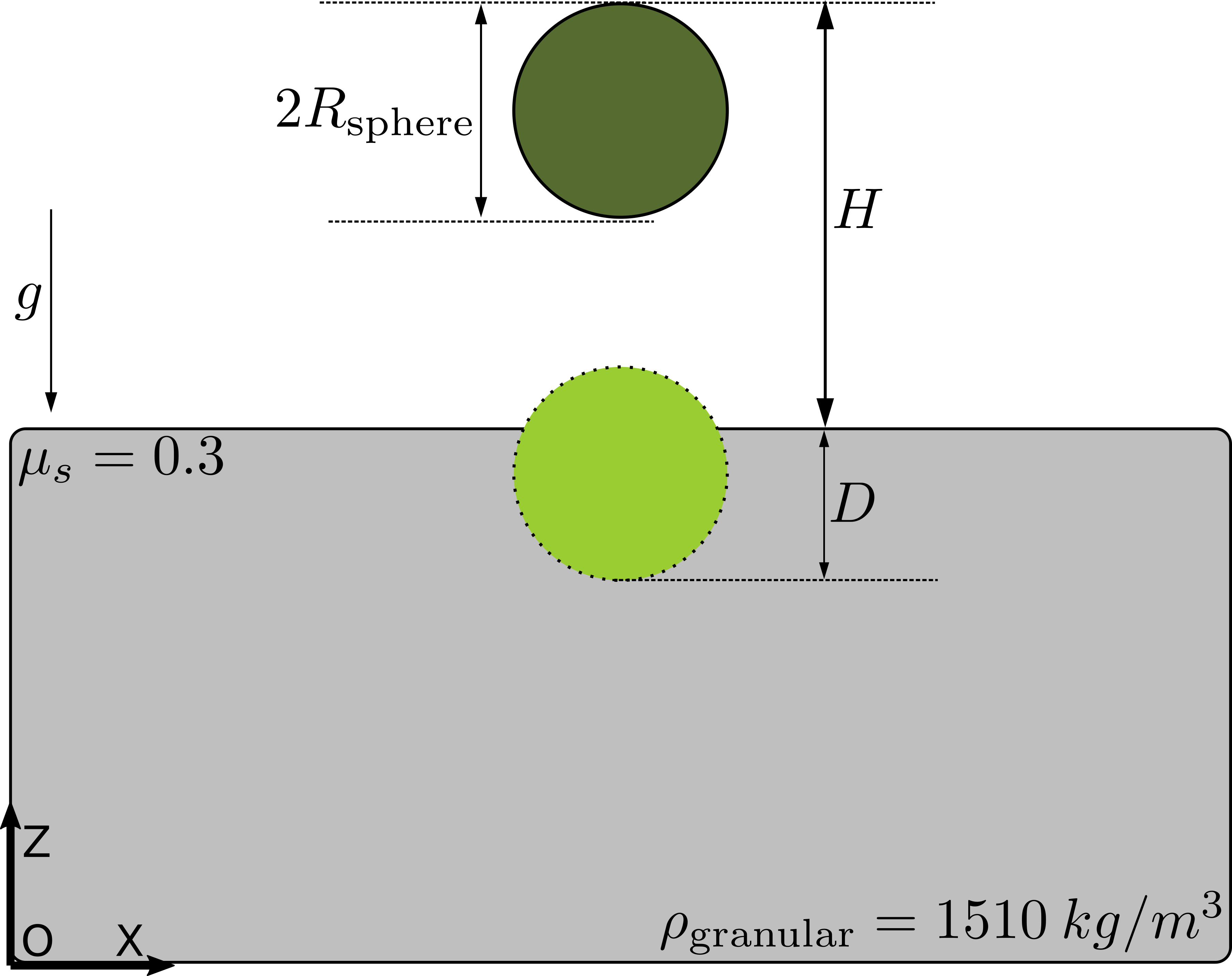}
    \caption{Sphere Cratering test setup.}
    \label{fig:sphere_cratering}
  \end{subfigure}
  \hfill
  \begin{subfigure}[t]{0.35\textwidth}
    \centering
    \includegraphics[width=\textwidth]{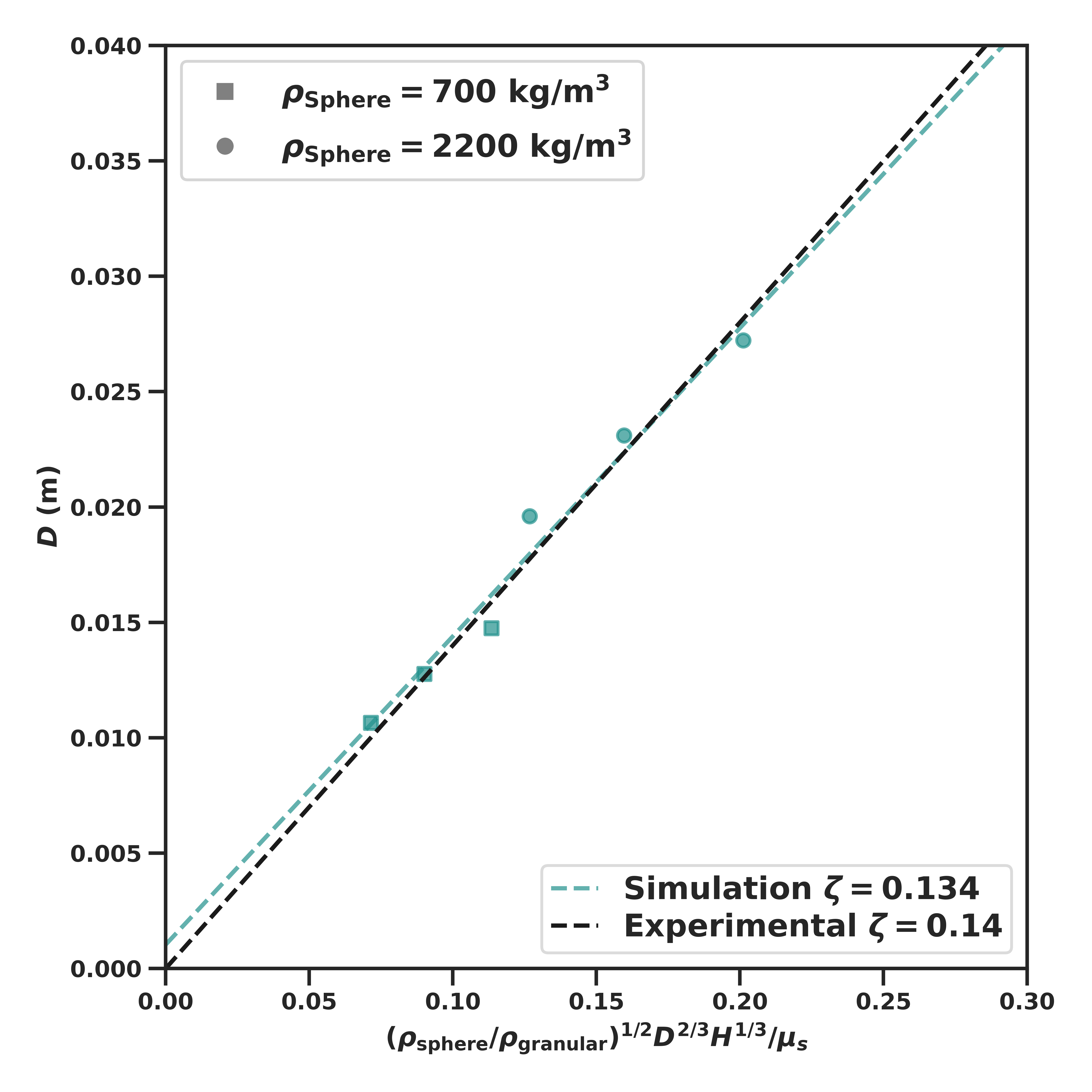}
    \caption{Variations in penetration depth across different sphere densities.}
    \label{fig:sphere_cratering_plot}
  \end{subfigure}
  \hfill
  \begin{subfigure}[t]{0.27\textwidth}
    \centering
    \includegraphics[width=\textwidth]{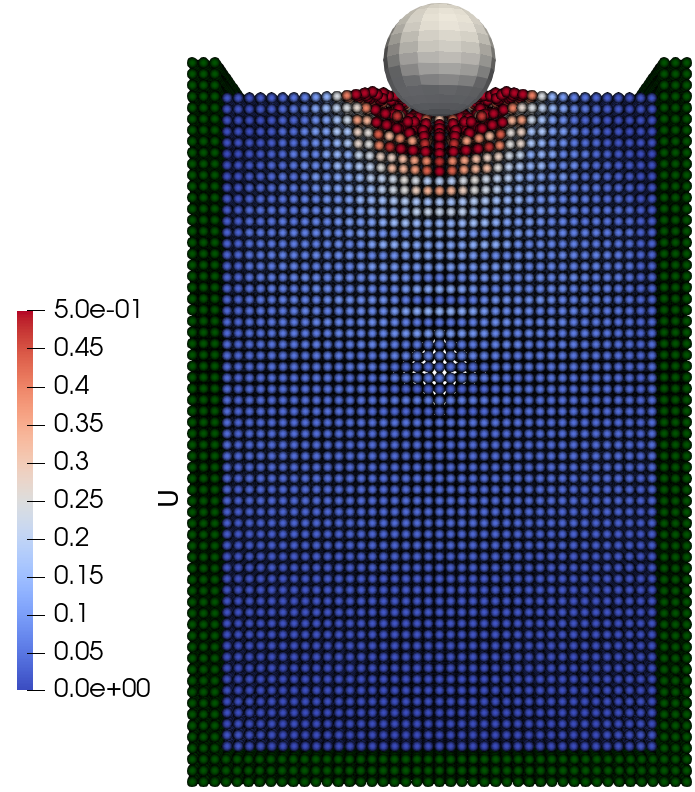}
    \caption{Velocity field at impact\\($t = $ \SI{2.5e-3}{s}).}
    \label{fig:sphere_cratering_velocity_1}
  \end{subfigure}
  \caption{Sphere Cratering test: (a) Schematic showing a sphere of radius $R_{\text{sphere}} = $ \SI{0.0125}{m} dropped from a height $H$ onto granular media. (b) Results showing penetration depth correlation with empirical model. (c) Velocity field visualization for a sphere with density $\rho_{\text{sphere}} = $ \SI{2200}{kg/m^3} dropped from $H = $ \SI{0.2}{m}. SPH particles are rendered as spheres with diameter $d = d_0 = $ \SI{2.5}{mm}. Similarly, the BCE markers are rendered with a green color.}
  \label{fig:sphere_cratering_combined}
\end{figure}

\begin{table}[H]
    \centering
    \caption{SPH parameters used for the different validation tests.}
    \label{tab:sph_params}
    \begin{tabular}{|c|c|c|c|c|c|c|c|}
      \hline
      \textbf{Test} & \textbf{Time Step ($\Delta t$)} & \textbf{Spacing ($d_0$)} & \textbf{Smoothing ($h$)} & \textbf{Viscosity ($\gamma_a$)} & \textbf{Boundary} & \textbf{Kernel}\\
      \hline
      Sphere Cratering & \SI{5e-5}{s} & \SI{0.0025}{m} & \SI{0.00325}{m} & Bilateral, 0.01 & Adami & Cubic\\
      \hline
      Cone Penetration & \SI{2e-5}{s} & \SI{0.0001}{m} & \SI{0.0013}{m} & Bilateral, 0.2 & Adami & Cubic\\
      \hline
      MGRU3 Single Wheel & \SI{2.5e-4}{s} & \SI{0.01}{m} & \SI{0.012}{m} & Unilateral, 0.02 & Adami & Cubic\\
      \hline
      RASSOR & \SI{5e-5}{s} & \SI{0.005}{m} & \SI{0.006}{m} & Unilateral, 0.02 & Adami & Cubic\\
      \hline
    \end{tabular}
  \end{table}

We compute the penetration depth $D$ for each simulation case. Figure~\ref{fig:sphere_cratering_plot} compares these simulated depths to the values predicted by the empirical relationship of Eq.~\ref{eq:ballDropEquation}. A linear regression fit applied to our simulated data points (shown as the blue dotted line in the figure) yields a slope of $0.1336$ with an $R^2$ value of $0.9714$. This slope is close to the experimentally measured value of $0.14$. Additionally, we calculate the Mean Squared Error (MSE) between our simulated penetration depths and the empirical predictions, resulting in a value of \SI{1e-7}{m^2}, which indicates good agreement between the simulation and the experimental results.

% ====================================================

\subsection{Cone Penetration test}
\label{sec:cone_penetration}
We next validate the solver using the cone penetration experiments described in~\cite{KyleExpConePenetration-TR2016}. These experiments involved two distinct scenarios using different granular materials and cone geometries.

In the first scenario, a \SI{60}{\degree} cone with a \SI{19.8}{mm} diameter was dropped into \SI{3}{mm} glass beads having a bulk density of \SI{1500}{kg/m^3}.
In the second scenario, a \SI{30}{\degree} cone with a \SI{9.2}{mm} diameter was dropped into 20-30 Ottawa sand having a bulk density of \SI{1780}{kg/m^3}.

For both of these scenarios, the cones were dropped from three different initial heights, defined relative to the cone's length $L$: $0$, $L/2$, and $L$.

\begin{figure}[htbp]
  \centering
  \begin{subfigure}[t]{0.44\textwidth}
    \centering
    \includegraphics[width=\textwidth]{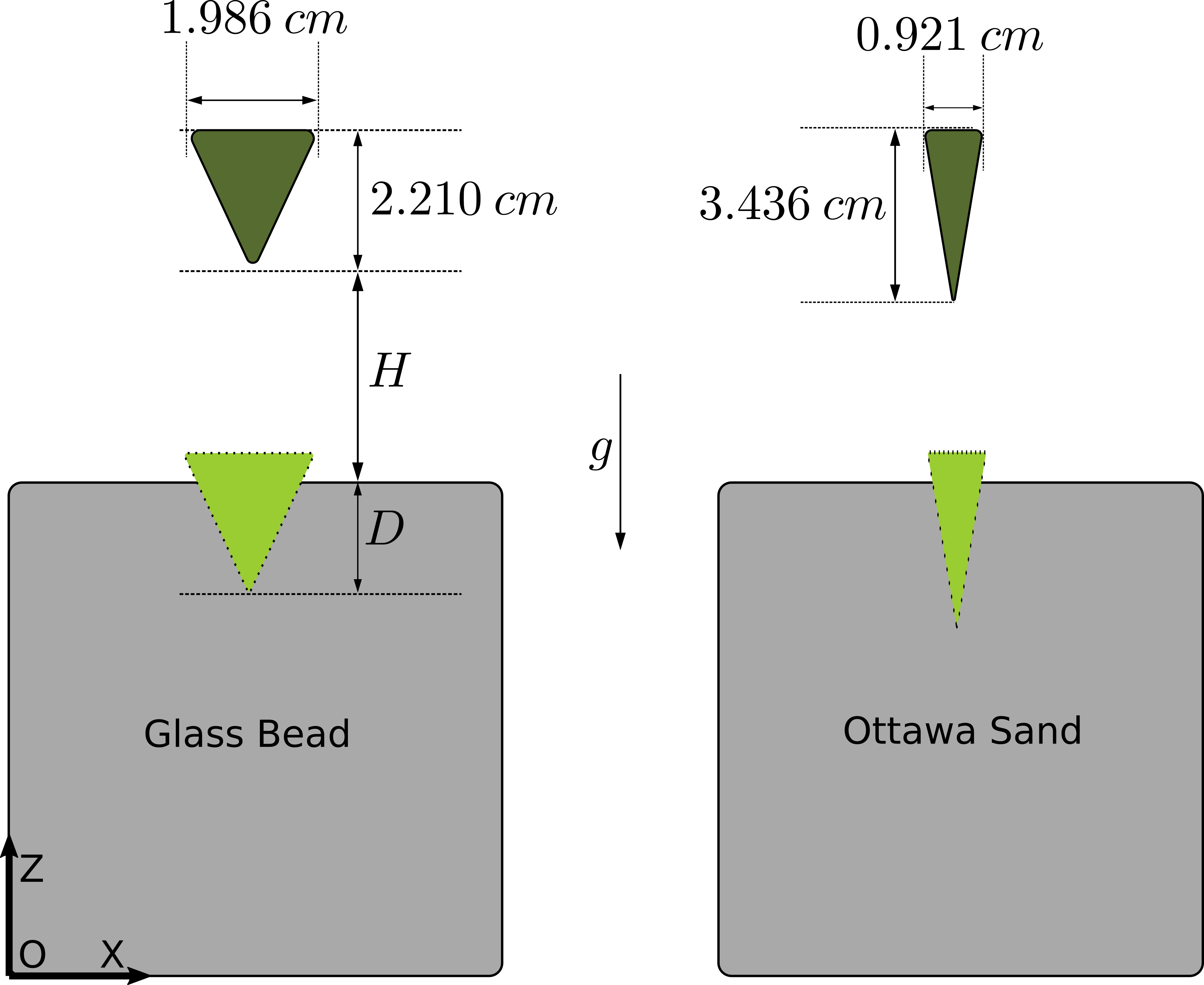}
    \caption{Schematic of the Cone Penetration test setup.}
    \label{fig:cone_penetration_setup}
  \end{subfigure}
  \hfill
  \begin{subfigure}[t]{0.54\textwidth}
    \centering
    \includegraphics[width=\textwidth]{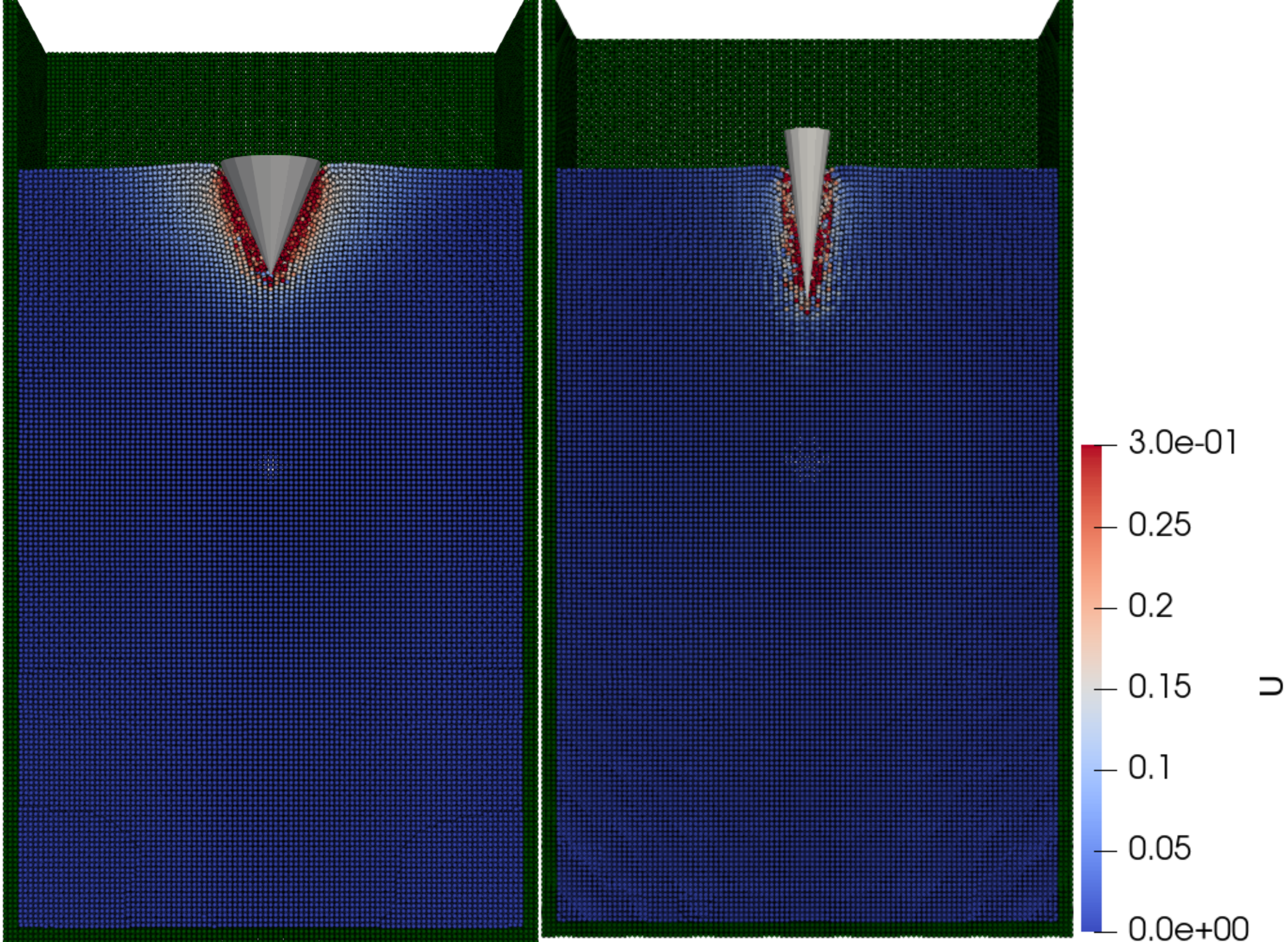}
    \caption{Velocity field at impact ($t = $ \SI{0.03}{s})}
    \label{fig:cone_penetration_vel_combined}
  \end{subfigure}
  \caption{Cone Penetration test: (a) Schematic of the test setup with cone geometries. (b) Velocity field visualization for the 60$^\circ$ cone on glass beads (Left) and the 30$^\circ$ cone on Ottawa sand (Right).}
  \label{fig:cone_penetration_combined}
\end{figure}

\begin{figure}[htbp]
  \centering
  \includegraphics[width=0.8\textwidth]{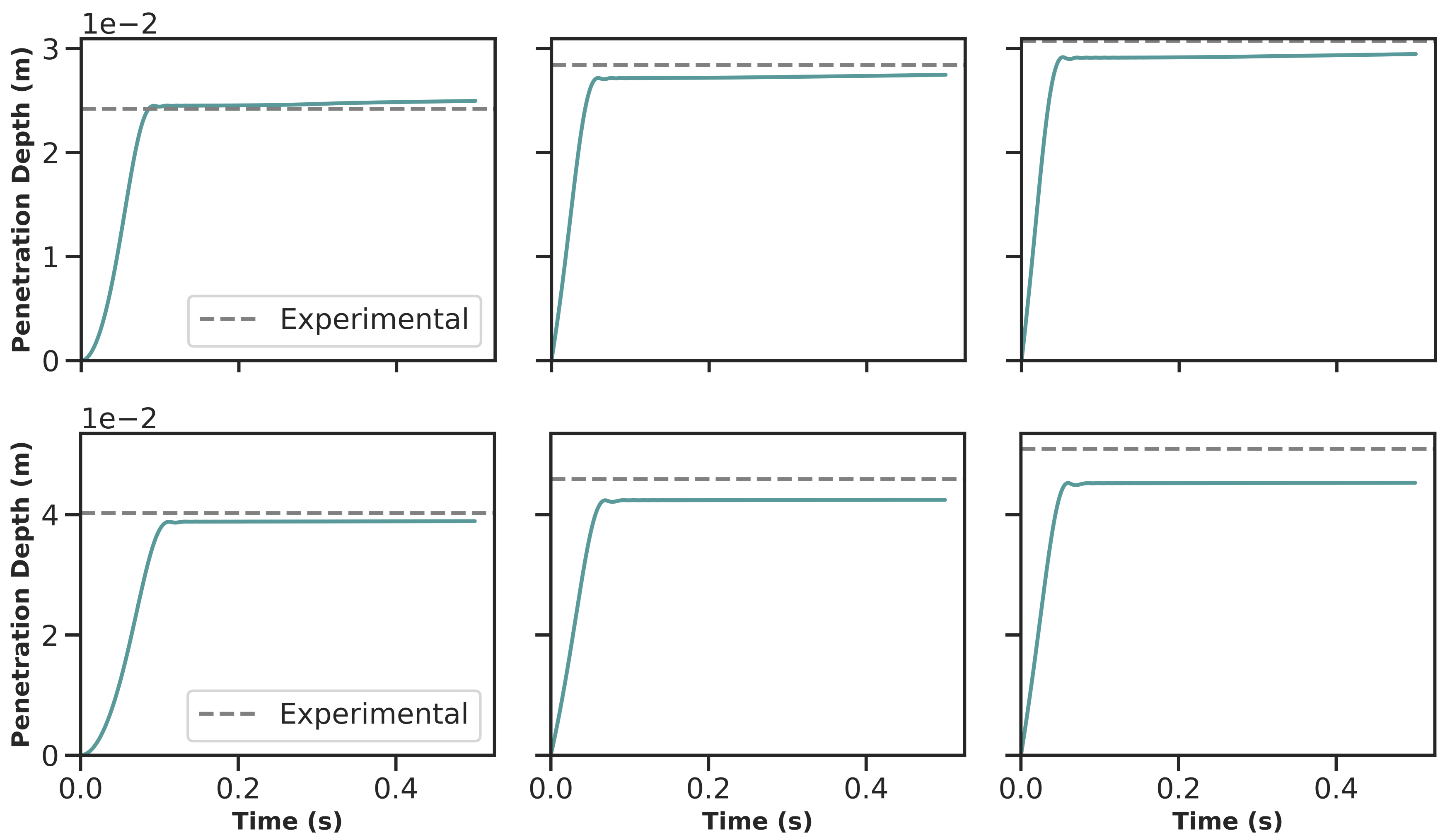}
  \caption{Simulation results for cone penetration depth versus time show good agreement with experiments~\cite{KyleExpConePenetration-TR2016}, with relative errors below 10\% for all cases. (Top) Data for a $60^\circ$ cone in glass beads. (Bottom) Data for a $30^\circ$ cone in Ottawa sand. Results cover three drop heights ($0$, $L/2$, $L$, left to right) for each scenario, where $L$ is the cone length.}
  \label{fig:cone_penetration_plot}
\end{figure}

For our simulations, we set the average particle diameter parameter in the rheology model ($d$ in Eq.~\ref{eq:muI}) to \SI{3}{mm} for the glass beads and \SI{7}{mm} for the Ottawa sand, based on information derived from the reference (specifically Fig.~1 in~\cite{KyleExpConePenetration-TR2016}).
We simulated the granular medium within a cubic container with dimensions \SI{0.1}{m} $\times$ \SI{0.1}{m} $\times$ \SI{0.1}{m}, which mimics the 4-inch diameter Proctor mold used in the experiments~\cite{KyleExpConePenetration-TR2016}.
For the rheology model, we set the parameters as follows: $\mu_s = 0.70$, $\mu_2 = 0.80$, and $I_0 = 0.08$ for the glass beads; and $\mu_s=0.80$, $\mu_2=1.00$, and $I_0=0.08$ for the Ottawa sand with cohesion ($c$ in Eq.~\ref{eq:pcrit}) set to \SI{0}{Pa}. These parameters were determined empirically, as specific experimental material data was not available. Both simulations utilized the same SPH parameters, summarized in Table~\ref{tab:sph_params}. Figure~\ref{fig:cone_penetration_combined} provides a schematic illustration of the simulation setup and includes velocity fields observed at the time of cone impact.
Since the physical experiments were conducted three times for each test case, we compare our simulation results to the average penetration depth measured across these three experimental trials, presented in Figure~\ref{fig:cone_penetration_plot}.

We obtain mean relative errors across all drop heights $H$ of 3.5\% for glass beads and 7.4\% for Ottawa sand. These values are comparable to the inherent variability observed in the repeated experimental trials~\cite{KyleExpConePenetration-TR2016}, indicating agreement between the simulation and the experimental results. This demonstrates the solver's ability to capture dynamics involved when non-trivial geometries, such as sharp cones, cut into a granular medium.

\subsection{MGRU3 single wheel test}
\label{sec:mgru3}
Here we compare data from physical rover testing conducted at the NASA SLOPE lab~\cite{movieMGRU3Tilt}, with simulation data generated using Chrono::CRM. While the underlying SPH solver was previously validated against a range of wheel and rover tests~\cite{weiCRM2024}, our goal is twofold: ($i$) confirm that our enhanced solver maintains accuracy against these experimental benchmarks using a representative test case; and ($ii$) compare its performance against recently published DEM results~\cite{ruochunGRC-DEM2023}, which is detailed in Sec.~\ref{sec:performance}. Through ($i$) and ($ii$), we confirm that Chrono::CRM improves computational speed without compromising the solution accuracy.

The experiments utilized the GRC-1 lunar simulant prepared to a bulk density of \SI{1760}{kg/m^3} (corresponding to 60\% relative density) with an associated friction angle $\phi = $ \SI{38.4}{\degree}. The tests featured the MGRU3 rover, which is a 1/6th scale model of NASA's VIPER rover. In each test run, the rover wheels were driven at a constant angular velocity $\omega = $ \SI{0.8}{rad/s} while the rover ascended a ramp inclined at an angle $\theta$ (with $\theta$ varying from \SI{0}{\degree} to \SI{30}{\degree}) relative to the horizontal. The steady-state average translational velocity of the rover, $\bar{v}$, was measured to calculate the wheel slip $s$, defined as $s = 1 - {\bar{v}}/\left({\omega r_{g}}\right)$, where $r_{g}$ represents the effective radius of the wheel with grousers.

\begin{figure}[htbp]
  \centering
  \includegraphics[width=0.8\textwidth]{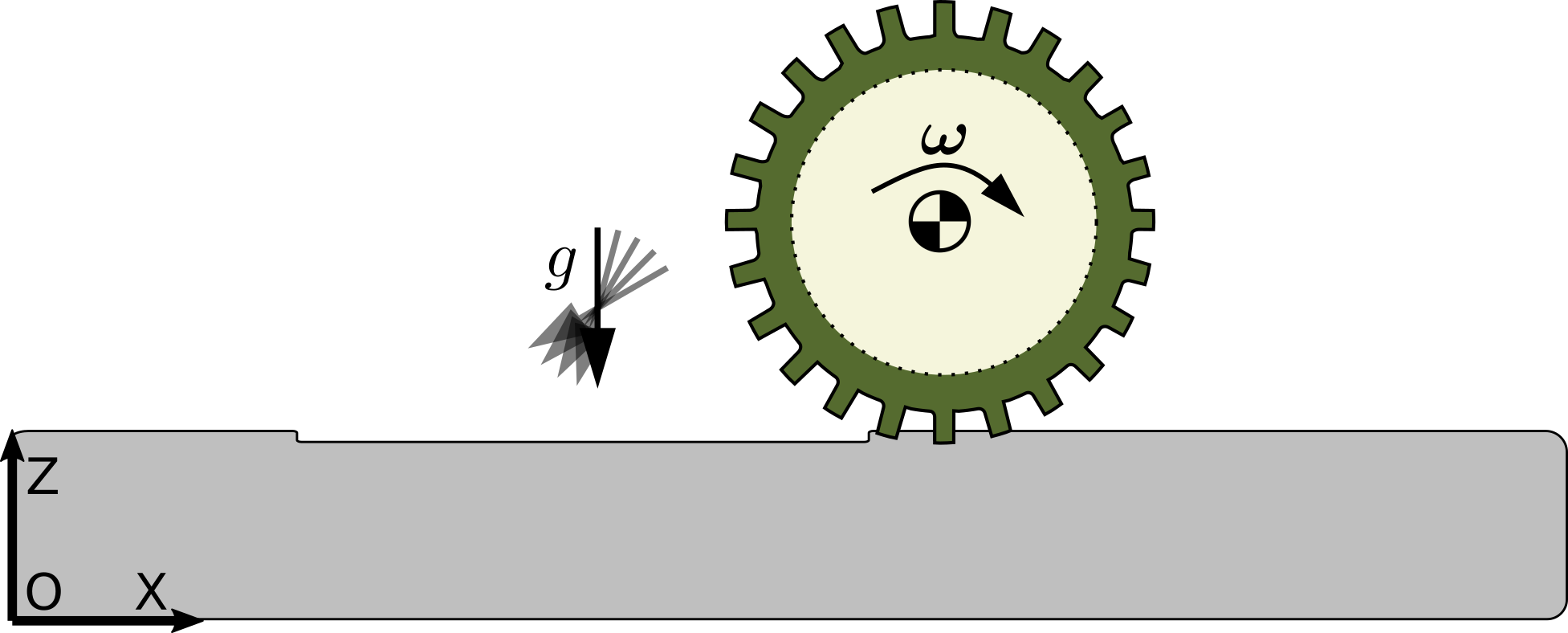}
  \caption{Schematic of the MGRU3 single wheel tests. The gravity direction is modified to simulate ramps of angles from $0^\circ$ to $30^\circ$. The wheel is given a constant angular velocity of $\omega = $ \SI{0.8}{rad/s} and is constrained to move in the $x$ direction.}
  \label{fig:mgru3_setup}
\end{figure}

For computational efficiency and based on prior findings that single-wheel results are representative of full rover behavior~\cite{weiCRM2024}, we simulated a single wheel subjected to one-quarter of the total rover mass. To model the inclined terrain, we modified the direction of the gravitational acceleration vector relative to the simulation domain instead of tilting all the elements of the model. The simulated soil bin has dimensions \SI{5}{m} $\times$ \SI{0.8}{m} $\times$ \SI{00.25}{m}. A schematic overview of this setup is shown in Figure~\ref{fig:mgru3_setup}. The MGRU3 wheel has an outer radius of \SI{0.25}{m}, a width of \SI{0.2}{m}, and 24 distinct grousers, each with a height of \SI{0.025}{m} and a width of \SI{0.01}{m}.

The simulation employs the specified GRC-1 material properties: bulk density \SI{1760}{kg/m^3} and friction angle $\phi =$ \SI{38.4}{\degree}. Owing to the large physical scale of the simulation domain (\SI{5}{m} length), a coarser SPH particle resolution and a correspondingly larger simulation time step were used compared to the previous validation cases. The complete set of SPH parameters used for this simulation is detailed in Table~\ref{tab:sph_params}.

\begin{figure}[htbp]
  \centering
  \begin{subfigure}[t]{0.46\textwidth}
    \centering
    \includegraphics[width=\textwidth]{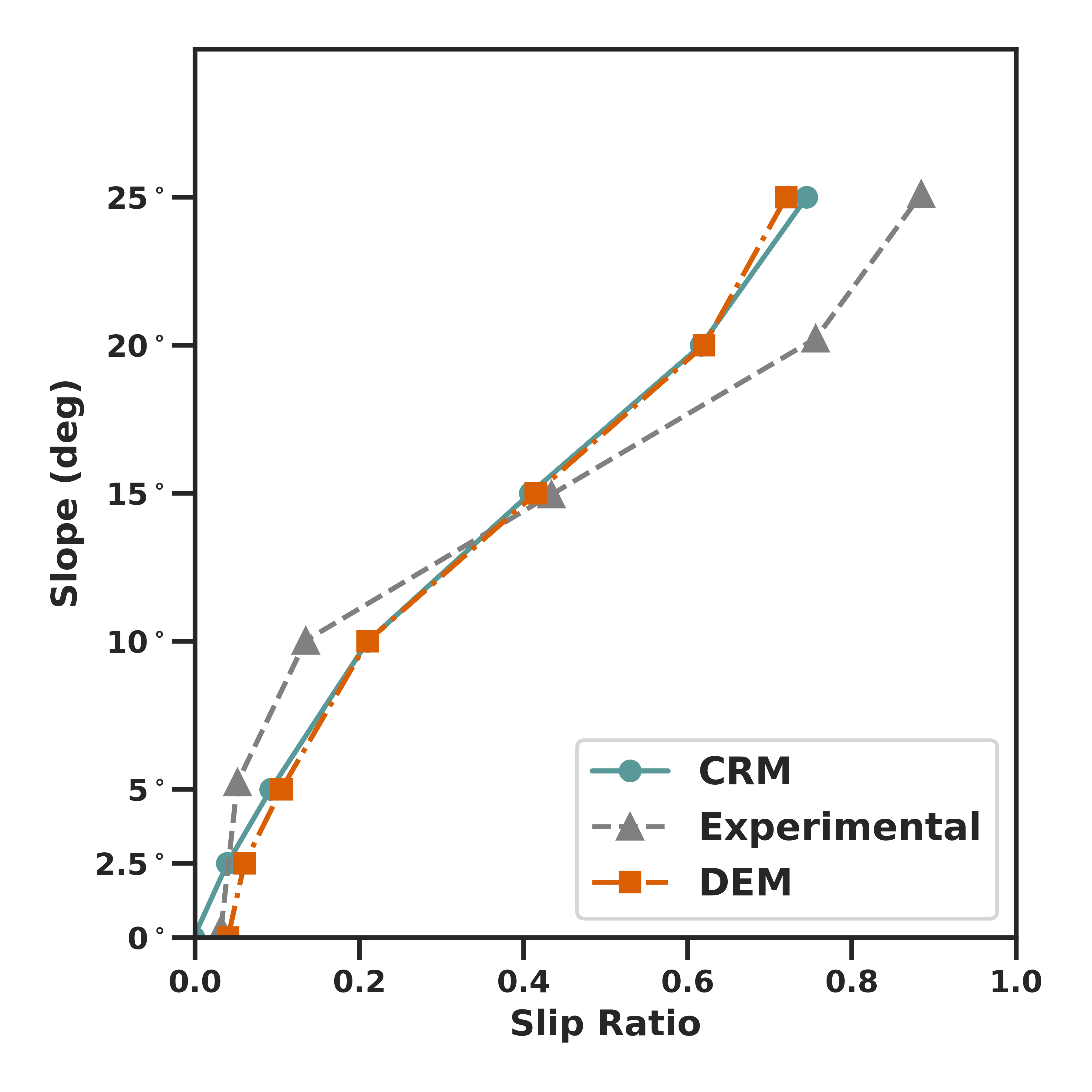}
    \caption{Slip vs slope curve for the MGRU3 single wheel test.}
    \label{fig:mgru3_slip_vs_slope}
  \end{subfigure}
  \hfill
  \begin{subfigure}[t]{0.48\textwidth}
    \centering
    \includegraphics[width=\textwidth]{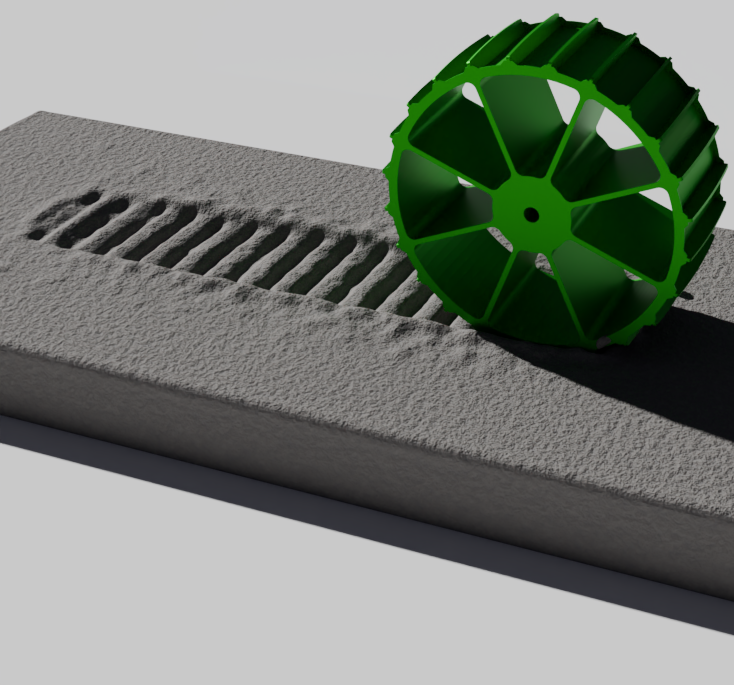}
    \caption{Treads created by the MGRU3 wheel.}
    \label{fig:mgru3_treads}
  \end{subfigure}
  \caption{Results obtained for the MGRU3 single wheel test: (a) The slip vs slope curve obtained compares the CRM results with the DEM results from~\cite{ruochunGRC-DEM2023} and the experimental results from the NASA SLOPE lab. The CRM results show close agreement with the DEM results and the experimental results. (b) Rendered views of the treads created by the MGRU3 wheel.}
  \label{fig:mgru3_slip_slope_treads}
\end{figure}

Figure~\ref{fig:mgru3_slip_vs_slope} reports the wheel slip ($s$) against the ramp inclination angle ($\theta$). The Chrono::CRM simulation results show good agreement with the DEM predictions published in~\cite{ruochunGRC-DEM2023}. Moreover, the CRM simulations align well with the experimental measurements, reinforcing the findings previously reported in~\cite{weiCRM2024}. Figure~\ref{fig:mgru3_treads} provides visual examples of the wheel tracks generated by the simulation.

% ======================================================

\subsection{RASSOR single drum test}
\label{sec:rassor}
\begin{figure}[htbp]
  \centering
  \begin{subfigure}[t]{0.7\textwidth}
    \centering
    \includegraphics[width=\textwidth]{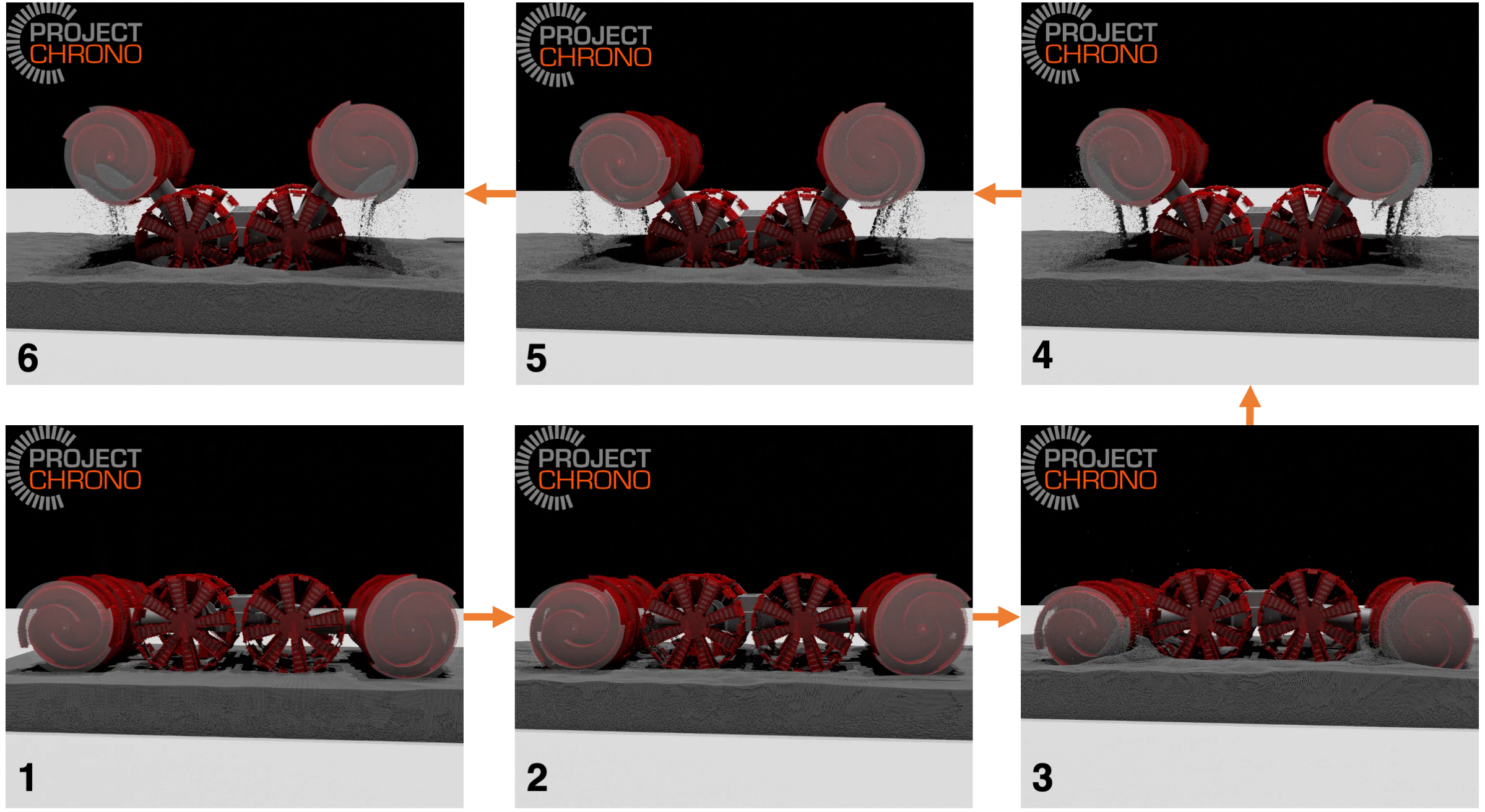}
    \caption{RASSOR's operation mechanism.}
    \label{fig:rassor_mechanism}
  \end{subfigure}
  \hfill
  \begin{subfigure}[t]{0.28\textwidth}
    \centering
    \includegraphics[width=\textwidth]{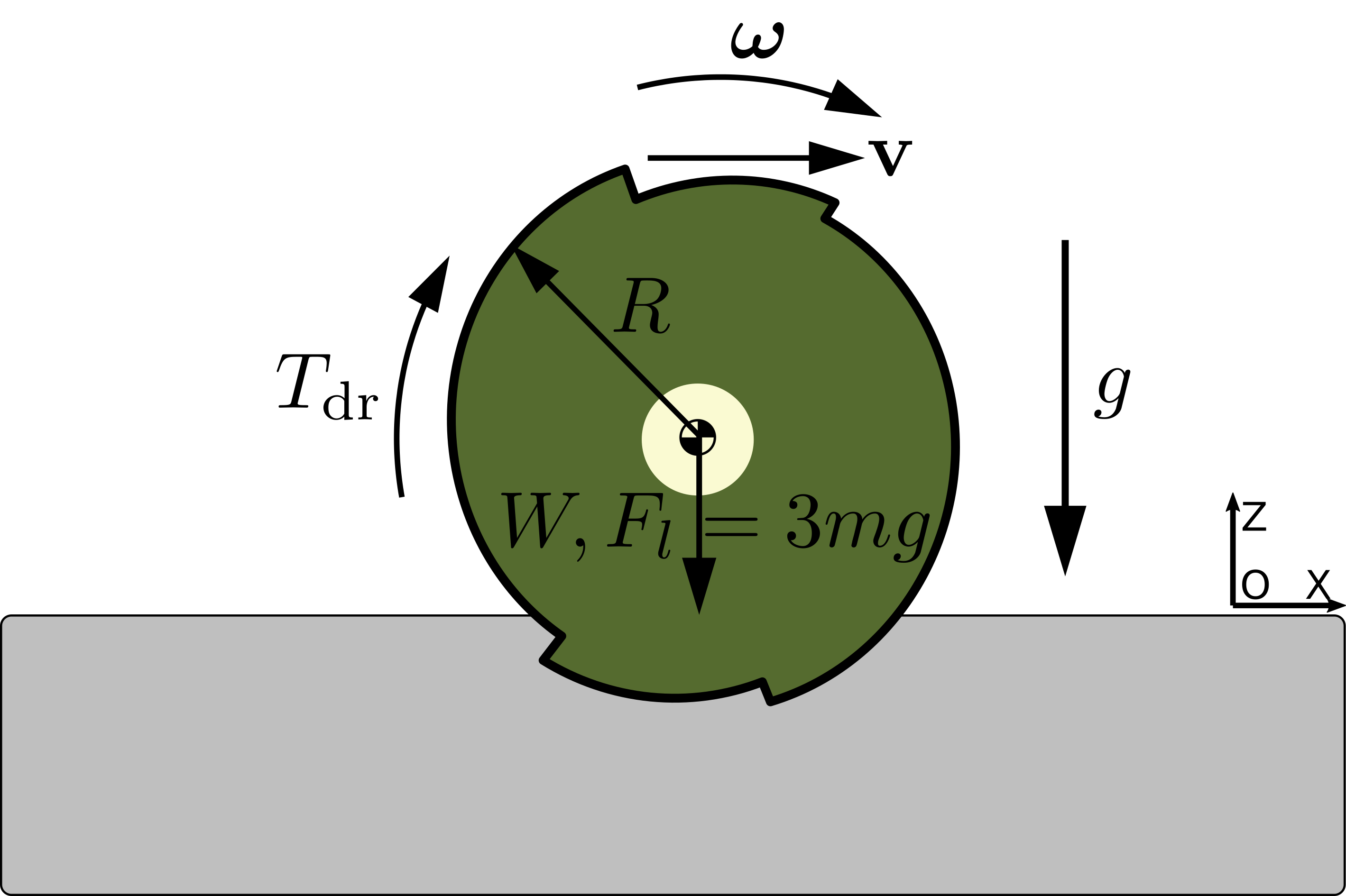}
    \caption{Schematic of the RASSOR single drum test setup done in this work.}
    \label{fig:rassor_setup}
  \end{subfigure}
  \caption{RASSOR single drum tests: (a) Figures (1) to (6) depict the mechanism with which the RASSOR is expected to excavate lunar regolith. (1) to (3) show the left drum rotating counter-clockwise and the right drum rotating clockwise to fill the RASSOR drum. (4) to (6) show RASSOR's arm lifted to empty the drum by rotating the left drum clockwise and the right drum counter-clockwise. (b) A schematic of the single drum test setup compares the Chrono::CRM solver with a DEM simulation. The drum is prescribed a constant angular and linear velocity. The driving torque $T_\text{dr}$ is measured and used as the metric for comparison.}
  \label{fig:rassor_mech_sch}
\end{figure}

The RASSOR~\cite{mueller2016design}, see Figure~\ref{fig:rassor_mechanism}, is a compact, lightweight planetary excavator developed by NASA. It is designed to autonomously mine regolith on the Moon, facilitating resource extraction in low-gravity environments where traction is limited. A key design feature is its use of counter-rotating bucket drums, a solution aimed at managing the net-zero horizontal reaction forces during excavation, which allows RASSOR to operate effectively despite strict mass constraints inherent in extraterrestrial operations. Simulating the interaction between RASSOR's bucket drums and lunar regolith is challenging, particularly in accurately capturing the cutting action of the drum edges as they engage the soil. The vast majority of existing terramechanics models, inspired by the Bekker-Wong and Wong-Reece theory~\cite{bekker69,wong67a}, are not suitable for analyzing this specific excavation process.

To validate the Chrono::CRM capability of simulating such excavation tasks, we model a single RASSOR drum interacting with granular terrain. The drum is prescribed a constant angular velocity $\omega = $ \SI{2.09}{rad/s} and simultaneously moved forward with a constant linear velocity $v = $ \SI{0.15}{m/s}. Similar to the MGRU3 wheel test configuration, the drum is allowed to move in both x and z directions. To simulate operational down-pressure, an additional vertical load $F_l = 3mg$ is applied to the drum, where $m$ is the drum mass and $g$ represents Earth's gravitational acceleration. The simulation employs GRC-1 material properties, specifically a bulk density $\rho = $ \SI{1700}{kg/m^3} and a friction angle $\phi =$ \SI{35}{\degree}. For the rheology model, we thus set the friction parameters $\mu_s = \mu_2 = 0.7$. Table~\ref{tab:sph_params} provides a summary of the SPH parameters used in this simulation, and Figure~\ref{fig:rassor_setup} shows a schematic of the overall setup.

For a comparative analysis, we simulate the same scenario using Chrono's DEM solver~\cite{ruochunGRC-DEM2023}. The GRC-1 terrain is prepared to replicate the same bulk density and frictional characteristics specified in the Chrono::CRM simulation. The metric chosen for comparison between the two methods is the driving torque required to rotate the drum over time.

\begin{figure}[htbp]
  \centering
  \begin{subfigure}[t]{0.55\textwidth}
    \centering
    \includegraphics[width=\textwidth]{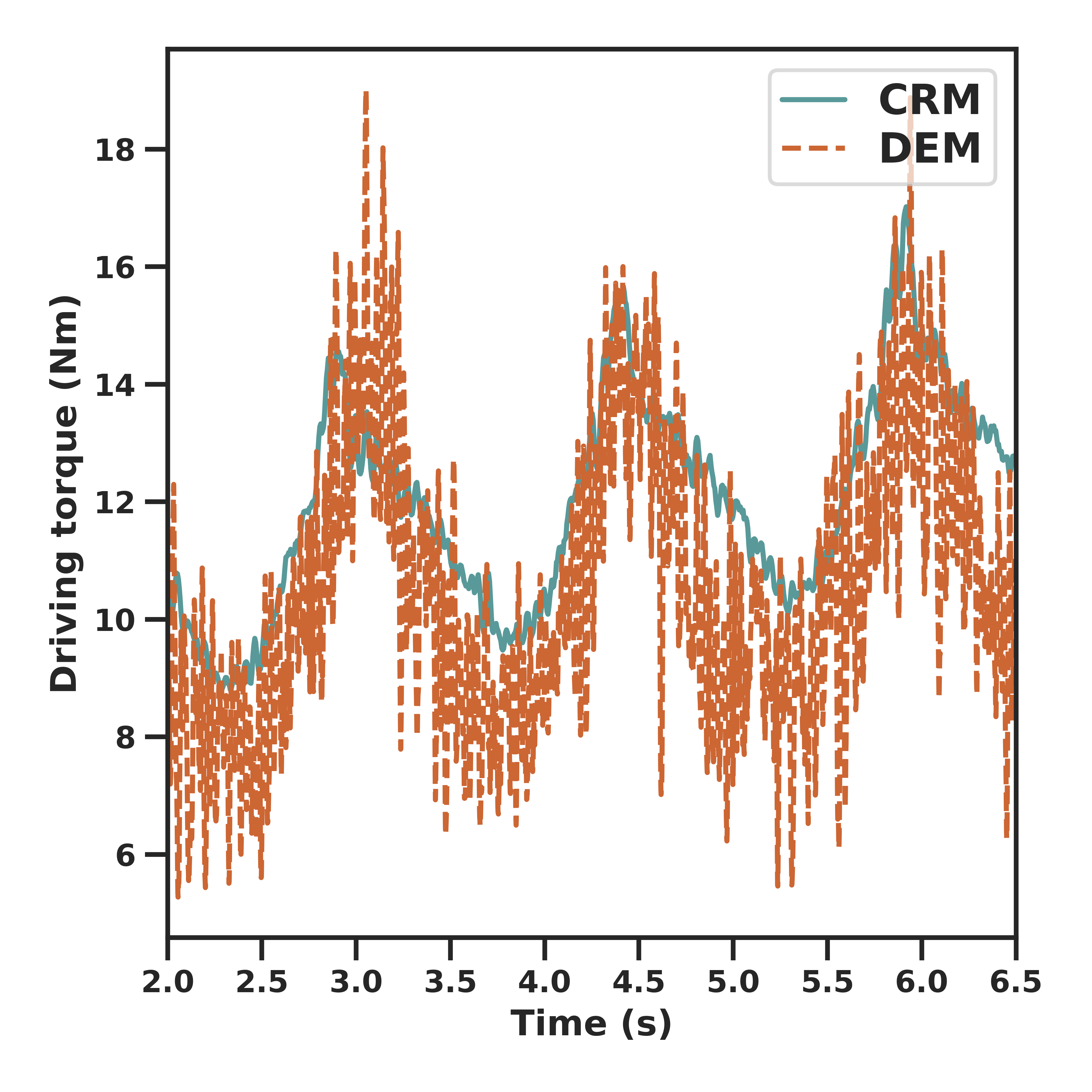}
    \caption{Driving torque for the RASSOR single drum test.}
    \label{fig:rassor_torque}
  \end{subfigure}
  \hfill
  \begin{subfigure}[t]{0.44\textwidth}
    \centering
    \includegraphics[width=\textwidth]{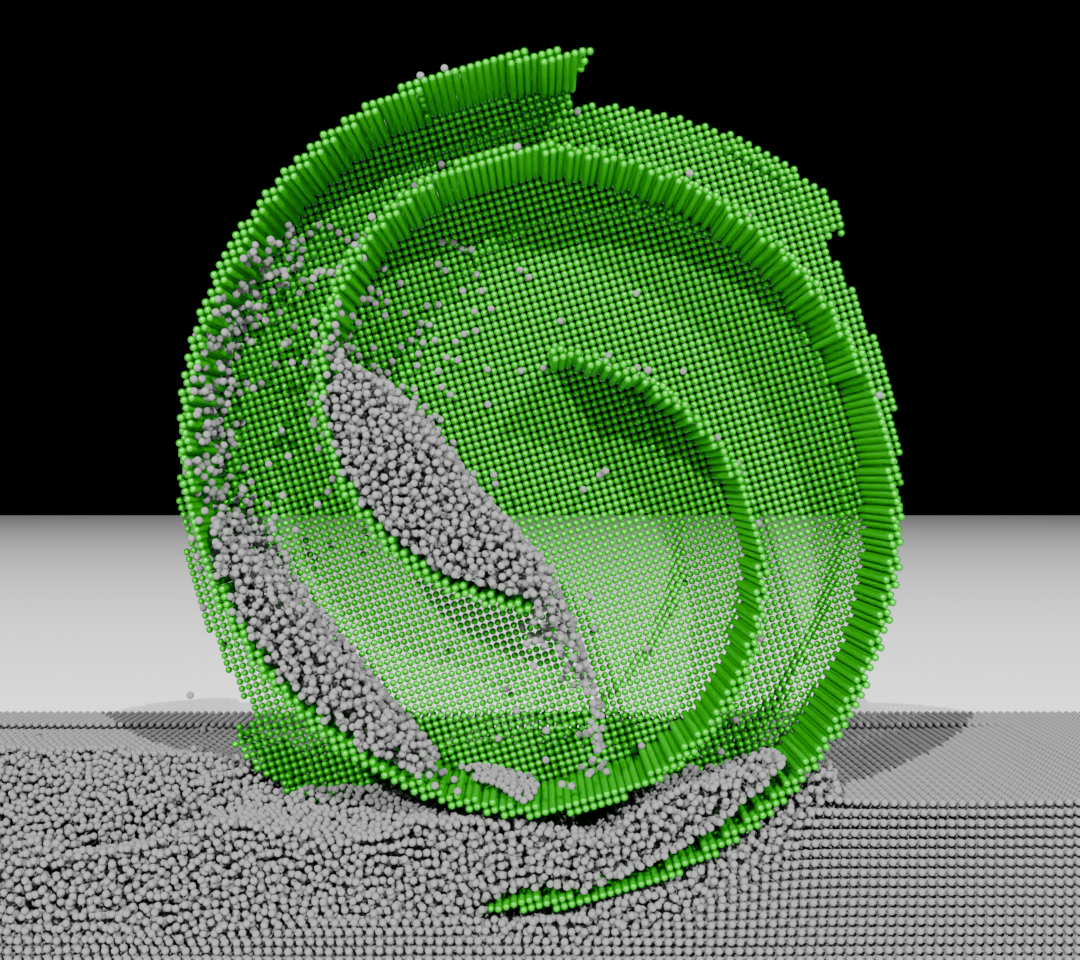}
    \caption{SPH particles collected inside the RASSOR drum at simulation time $t = $ \SI{2.375}{s}.}
    \label{fig:rassor_collection}
  \end{subfigure}
  \caption{Results obtained for the RASSOR single drum test: (a) The driving torque for the RASSOR single drum test compares the CRM results with the benchmark DEM-E results \cite{ruochunGRC-DEM2023}. The CRM results show close agreement with the DEM results. (b) A snapshot of a cross-section view of the RASSOR drum showing the quantity of GRC1 terrain excavated (represented as SPH particles).}
\end{figure}

Figure~\ref{fig:rassor_torque} shows that the driving torque predicted by the Chrono::CRM simulation follows a similar trend and matches the magnitude reasonably well compared to the benchmark DEM simulation. This alignment suggests that the continuum SPH approach effectively captures the overall resistive forces generated during the excavation process. As the drum rotates and cuts through the terrain, SPH particles, which represent the soil continuum, are collected inside the drum (see Fig.~\ref{fig:rassor_collection}). The accumulation of this excavated mass increases the drum's rotational inertia, consequently causing the driving torque to increase over time, as observed in Fig.~\ref{fig:rassor_torque}. It is important to understand that within the SPH framework, these particles function as numerical integration points for the continuum equations; they do not correspond directly to individual soil grains but collectively represent the bulk mass and volume of the excavated soil.

\section{Solver Performance}
\label{sec:performance}
% In the previous sections, we simulated with Chrono::CRM the interaction of varying geometries with many different granular materials, each having different material properties. We found that in most cases, Chrono::CRM produces results that are in close agreement with experimental results or higher-fidelity DEM simulations. However, often times, along with simulations that are indicative of the real-world behavior, engineers are also interested in obtaining these results in a timely manner. Time is of even greater essence when it comes to applications such as controller design or design optimization, where it is often times required to run thousands of simulations for exploration of the design space. In such settings, engineers would greatly benefit from a solver that is not only accurate but also fast and computationally efficient. DN: I COMMENTED THIS OUT SINCE I THOUGTH THIS WAS PRETTY CLEAR AND UNSURPRISING.

In this section, we discuss performance optimizations that resulted in up to 10-fold speedups relative to the previous version of the CRM solver. We show that these speedups don't lead to noticeable losses in accuracy for all validation benchmarks discussed in Sec.~\ref{sec:validation}. We report close to real-time simulation speeds across a range of benchmarks, including those involving tracked vehicles, wheeled vehicles, deformable bodies, and full-scale planetary rovers.

\subsection{Persistent Neighbor Lists and Update Frequency Control}
\label{sec:persistent_neighbor_list}
In the SPH method, properties at a particle's location are interpolated from the values of its neighbors, weighted by a kernel function. Standard kernel functions have a finite support radius, denoted as $\mathcal{K}h$, where $h$ is the smoothing length and the factor $\mathcal{K}$ depends on the specific kernel. Chrono::CRM utilizes Quintic Wendland~\cite{Wendland1995} and Cubic~\cite{Monaghan1985} kernels. For these kernels in 3D simulations, $\mathcal{K} = 2$. Consequently, a core task at each time step is the neighbor search: for every particle $i$, the algorithm must identify all other particles $j$ located within a radius of $2h$. For brevity, ``particle'' hereafter refers collectively to both SPH particles and BCE markers, and simulations are assumed to be 3D. Performing this neighbor search naively by checking all particle pairs requires $\mathcal{O}(N^2)$ operations, where $N$ is the total number of particles. To accelerate this process, Chrono::CRM constructs neighbor lists using a cell-linked list approach, similar to that described in~\cite{dominguez2011neighbour}. This procedure involves the following steps:

\begin{enumerate}
    \item The algorithm assigns each particle to a cell in an axis-aligned grid covering the simulation domain. The grid cell size is set to $2h$ in all dimensions, ensuring that any potential neighbor $j$ of a particle $i$ resides either in $i$'s cell or one of the 26 adjacent cells. A particle's 3D grid coordinates $(x,y,z)$ are mapped to a unique linear index $c$ using the formula $c = z \cdot (Y_{\text{size}} \cdot X_{\text{size}}) + y \cdot X_{\text{size}} + x$, where $X_{\text{size}}$ and $Y_{\text{size}}$ are the grid dimensions. This spatial hashing step requires iterating through all particles and has a time complexity of $\mathcal{O}(N)$, where $N$ is the number of particles.
    \item Next, the particles are sorted based on their cell index $c$. This groups particles within the same or nearby grid cells together in memory, which improves cache performance during the neighbor search by enhancing spatial locality. This sorting step typically has a time complexity of $\mathcal{O}(N \log N)$.
    \item After sorting, helper arrays \texttt{cellStart} and \texttt{cellEnd} are computed. For each cell index $c$, \texttt{cellStart[c]} stores the index of the first particle belonging to that cell in the sorted list, and \texttt{cellEnd[c]} stores the index immediately following the last particle of that cell. These arrays enable efficient access to all particles within a specific grid cell. This step involves a single pass over the sorted particles and has a time complexity of $\mathcal{O}(N+M)$, where $M$ is the total number of grid cells.
    \item Finally, the neighbor list for each particle is constructed as detailed in Algorithm~\ref{alg:neighbor_list}. For each particle $i$, the algorithm examines particles $j$ located within the 27 grid cells adjacent to particle $i$. If the distance between particles $i$ and $j$ is less than the interaction radius $2h$, particle $j$ is added to the neighbor list of $i$. The time complexity of this neighbor search step is $\mathcal{O}(N k)$, where $k$ is the maximum number of particles found within any single grid cell. For reasonably uniform particle distributions, $k$ is typically small and bounded on average (usually around 100), leading to an average time complexity close to $\mathcal{O}(N)$.
\end{enumerate}

This leads to an overall time complexity of $\mathcal{O}(N \log N + N + N k) = \mathcal{O}(N \log N + N k)$ which is better than the naive $\mathcal{O}(N^2)$ complexity. This neighbor lists are then used to compute the SPH properties at each particle location during the force computation step.
\SetKwFunction{KwCalcGridPos}{calcGridPos}
\SetKwFunction{KwCalcGridID}{calcGridID}  
\SetKwFunction{KwDist}{dist}
\SetKwFunction{KwCall}{Call}
\SetKwFunction{KwForceComputation}{ForceComputation}
\SetKwFunction{KwDynamicsUpdate}{DynamicsUpdate}
\SetKwFunction{KwPreprocessNeighborList}{PreprocessNeighborList}
% Algorithm using algorithm2e syntax
\begin{algorithm}
    \caption{Neighbor Lists Construction}\label{alg:neighbor_list}
    \KwIn{vector $pos$, $cellStart$, $cellEnd$, $offset$}
    \KwOut{vector $neighborList$}
    \For{$i \leftarrow 0$ \KwTo $numParticles - 1$}{
        $currGridPos \leftarrow \KwCalcGridPos(pos[i])$\;
        $count \leftarrow 0$\;
        \For{$x \leftarrow -1$ \KwTo $1$}{
            \For{$y \leftarrow -1$ \KwTo $1$}{
                \For{$z \leftarrow -1$ \KwTo $1$}{
                    $neighborGridPos \leftarrow currGridPos + (x, y, z)$\;
                    $gridID \leftarrow \KwCalcGridID(neighborGridPos)$\;
                    $begin \leftarrow cellStart[gridID]$\;
                    $end \leftarrow cellEnd[gridID]$\;
                    \For{$j \leftarrow begin$ \KwTo $end - 1$}{
                        \If{$\KwDist(pos[i]-pos[j]) < 2h$}{
                            \tcp{The $offset$ array is precomputed using a prefix sum of the number of neighbors per particle}
                            $neighborList[offset[i]+count] \leftarrow j$\;
                            $count \leftarrow count+1$\;
                        }
                    }
                }
            }
        }
    }
\end{algorithm}

Our approach introduces ``persistent'' neighbor lists, updated at a frequency controlled by the user. Specifically, the user sets a parameter $ps_{\text{freq}}$, representing the number of time steps before the neighbor lists are updated. If $ps_{\text{freq}} = 1$, the lists are rebuilt every time step, replicating the standard approach. If $ps_{\text{freq}} > 1$, the same neighbor lists are reused for $ps_{\text{freq}}$ steps before being updated, reducing the computational cost associated with the neighbor search. Algorithm~\ref{alg:simulation_loop} illustrates how this conditional update integrates into the main simulation loop.

\begin{algorithm}[H] % Use [H] to suggest placement here (requires float package) or omit for floating
    \caption{Simulation Loop with Persistent Neighbor Lists Update}\label{alg:simulation_loop}
    \KwIn{Initial particle state (e.g., $pos$, $vel$), simulation parameters ($t_{max}$, $dt$, $ps_{freq}$)}
    % Note: neighborList is internal state, computed/updated within the loop.
    % Assumes necessary simulation domain/grid info is accessible.
    \BlankLine
    % Optional: Explicitly build initial neighbor list before loop if needed for t=0 forces.
    % $neighborList \leftarrow \KwCall{Algorithm~\ref{alg:neighbor_list}}{pos_{initial}}$\;
    \For{$t \leftarrow 0$ \KwTo $t_{max}$}{
        \If{$t \pmod{ps_{freq}} = 0$}{
             \tcp{Rebuild neighbor lists based on current positions}
             \tcp{Preprocess the global neighbor list to compute the offset array}
             $cellStart, cellEnd, offset \leftarrow \KwPreprocessNeighborList(pos)$\;
             % Call the full neighbor list construction procedure (Algorithm 1)
             $neighborList \leftarrow \KwCall{Algorithm~\ref{alg:neighbor_list}}{pos, cellStart, cellEnd, offset}$\;
        }
        % If condition is false (t % ps_freq != 0), the existing neighborList is reused.
        \BlankLine
        \tcp{Compute forces using the current (potentially stale) neighbor lists}
        % Assuming ForceComputation takes positions and the neighbor list
        $forces \leftarrow \KwForceComputation(pos, neighborList)$\;
        \BlankLine
        \tcp{Update particle positions and velocities (Integration step)}
        % Assuming DynamicsUpdate takes state, forces, and timestep
        $\KwDynamicsUpdate(pos, vel, forces, dt)$\;
        \BlankLine
    }
    % Implicit Output: Final particle state (pos, vel) at t_max
\end{algorithm}
The concept of a persistent neighbor list, often termed a Verlet list, is described in~\cite{dominguez2011neighbour}. In that approach, particles potentially entering the interaction radius during the list's lifespan are included preemptively. This is achieved by dividing the domain into cells of size $2H = 2h + \delta h$ and constructing the neighbor list using this larger radius $2H$. The safety margin $\delta h$ is typically determined heuristically based on expected maximum particle velocities and the number of time steps the list will be reused. During list construction, the algorithm searches the 27 cells surrounding a target particle's cell and includes all particles found within the distance $2H$. Consequently, the resulting list contains particles that are not necessarily current neighbors but might become neighbors (i.e., come within distance $2h$) in subsequent steps while the list remains valid. This necessitates an additional distance check during the force computation stage to ensure only particle pairs currently within the true SPH interaction radius $2h$ contribute to the force calculation.

For context regarding the underlying SPH implementation in Chrono, the previous version described in~\cite{weiGranularSPH2021} did not employ persistent neighbor lists. Instead, neighbor searches were performed during the force computation step required for evaluating the equations of motion (Eqs.~\ref{equ:continuity_dis}, \ref{equ:momentum_dis}, \ref{equ:stress_rate_dis}).

In this work, we adopt and evaluate a simpler persistent list strategy. We avoid increasing the cell size or the search radius during list construction; instead, the neighbor lists are built using only the standard SPH interaction radius $2h$. This list, containing only confirmed current neighbors, is then kept persistent and reused for $ps_{\text{freq}}$ time steps. We will first demonstrate in Subsec.~\ref{sec:persistent_neighbor_list_accuracy} that this simplification does not compromise accuracy, even when the list is reused for $ps_{\text{freq}} = 10$ steps, across various test cases. Subsequently, in Subsec.~\ref{sec:persistent_neighbor_list_speedup}, we quantify the computational speedup achieved by reusing the list ($ps_{\text{freq}} > 1$) compared to the baseline of rebuilding it at every step ($ps_{\text{freq}} = 1$). 

\subsubsection{Accuracy of Persistent Neighbor Lists}
\label{sec:persistent_neighbor_list_accuracy}
The intuition behind the effectiveness of the simple strategy of reusing neighbor lists over multiple time steps  lies in the nature of particle movement and SPH kernel properties. When a list constructed at time $t$ is reused at a later time $t + \kappa \Delta t$ (where $1 \le \kappa < ps_{\text{freq}}$), it becomes ``stale'' relative to the particle positions at $t+\kappa \Delta t$. Specifically, the list might wrongly include particles that have moved slightly beyond the $2h$ radius and omit particles that have moved slightly inside it. However, since relative particle displacements between consecutive list updates are typically small, the \textit{majority} of the true neighbors remain captured in the stale list. The errors caused by incorrect inclusions or omissions primarily involve particles near the boundary of the $2h$ interaction sphere. SPH kernel functions assign small weights to particles near the edge of their support radius. Therefore, the minor discrepancies at the boundary caused by using the stale list have a negligible impact on the SPH summations for density, forces, and other interpolated quantities. Moreover, the speed of a particle in CRM is typically low, while the integration step is also small, of the order of $10^{-5}$ to $10^{-4}$ seconds. These two factors, small speed and small time step, combine to further reduce the impact of stale neighbor lists on simulation results -- within a few time steps, very few particles enter or exit the $2h$ radius.

\begin{figure}[htbp]
    \centering
    \begin{subfigure}[t]{0.32\textwidth}
      \centering
      \includegraphics[width=\textwidth]{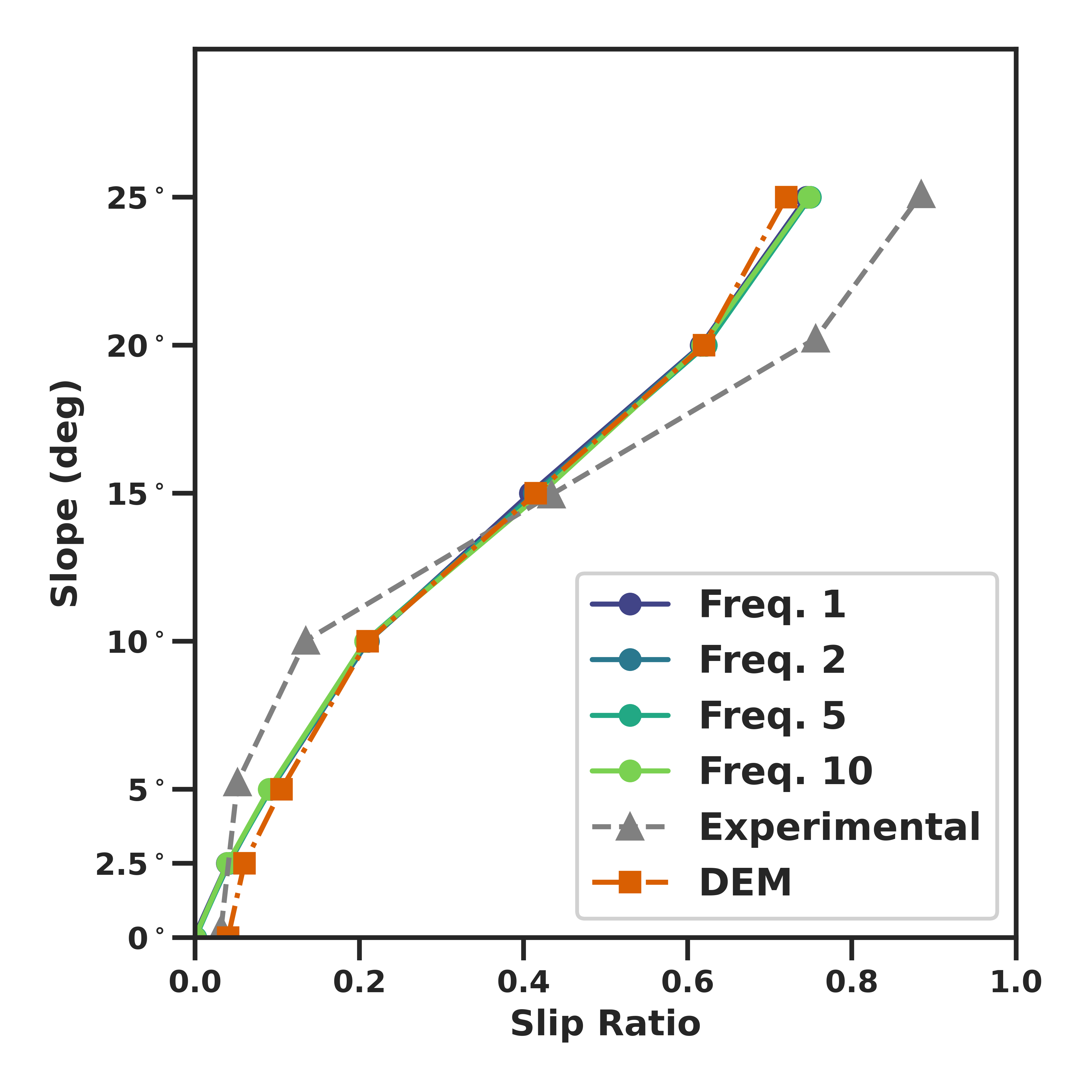}
      \caption{MGRU3 results from Sec.~\ref{sec:mgru3}.}
      \label{fig:mgru3_results_ps_comp}
    \end{subfigure}
    \hfill
    \begin{subfigure}[t]{0.32\textwidth}    
      \centering
      \includegraphics[width=\textwidth]{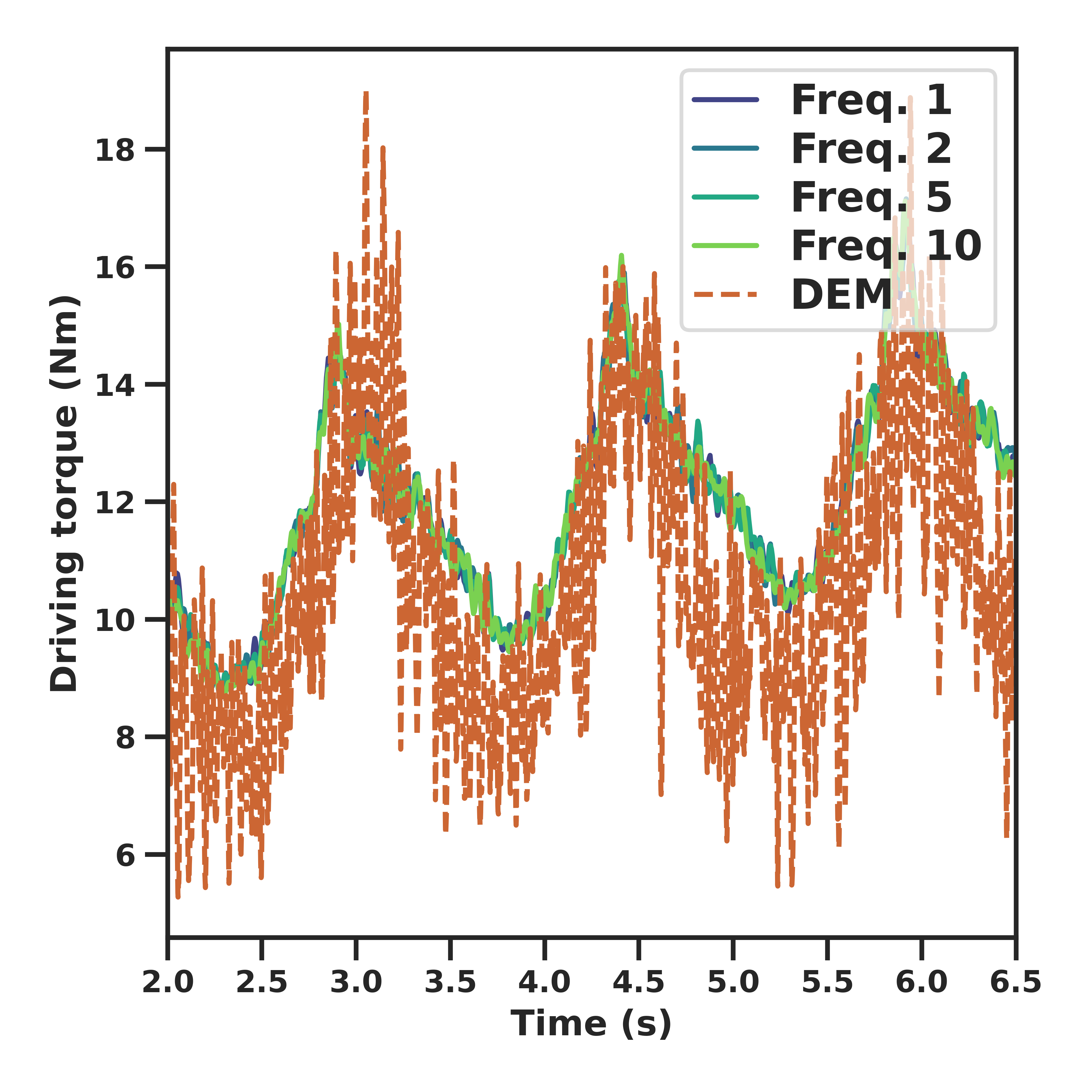}
      \caption{RASSOR results from Sec.~\ref{sec:rassor}.}
      \label{fig:rassor_ps_comp}
    \end{subfigure}
    \hfill
    \begin{subfigure}[t]{0.32\textwidth}
      \centering
      \includegraphics[width=\textwidth]{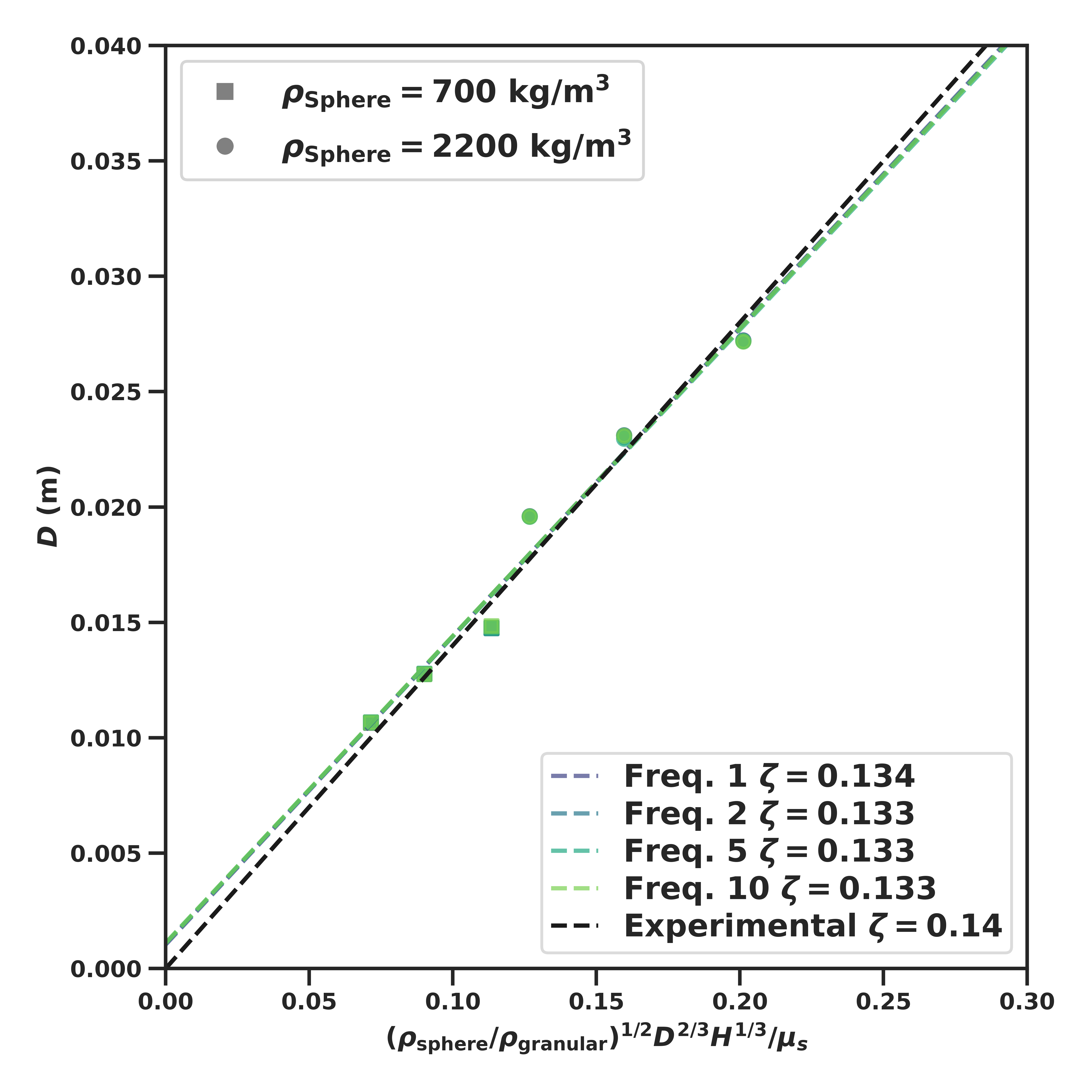}
      \caption{Sphere cratering results from Sec.~\ref{sec:sphere_cratering}.}
      \label{fig:cratering_ps_comp}
    \end{subfigure}
    \caption{Comparison of results using different neighbor list update frequencies ($ps_{\text{freq}}$) ranging from 1 (update every step) to 10. Subfigures show results for: (a) MGRU3 (Sec.~\ref{sec:mgru3}), (b) RASSOR (Sec.~\ref{sec:rassor}), and (c) Sphere Cratering (Sec.~\ref{sec:sphere_cratering}). The close agreement across different $ps_{\text{freq}}$ values demonstrates no loss of accuracy when reusing the neighbor lists.}
    \label{fig:persistent_neighbor_list_accuracy}
\end{figure}

\begin{figure}[htbp]
    \centering
    \begin{subfigure}[t]{0.7\textwidth}
        \centering
        \includegraphics[width=\textwidth]{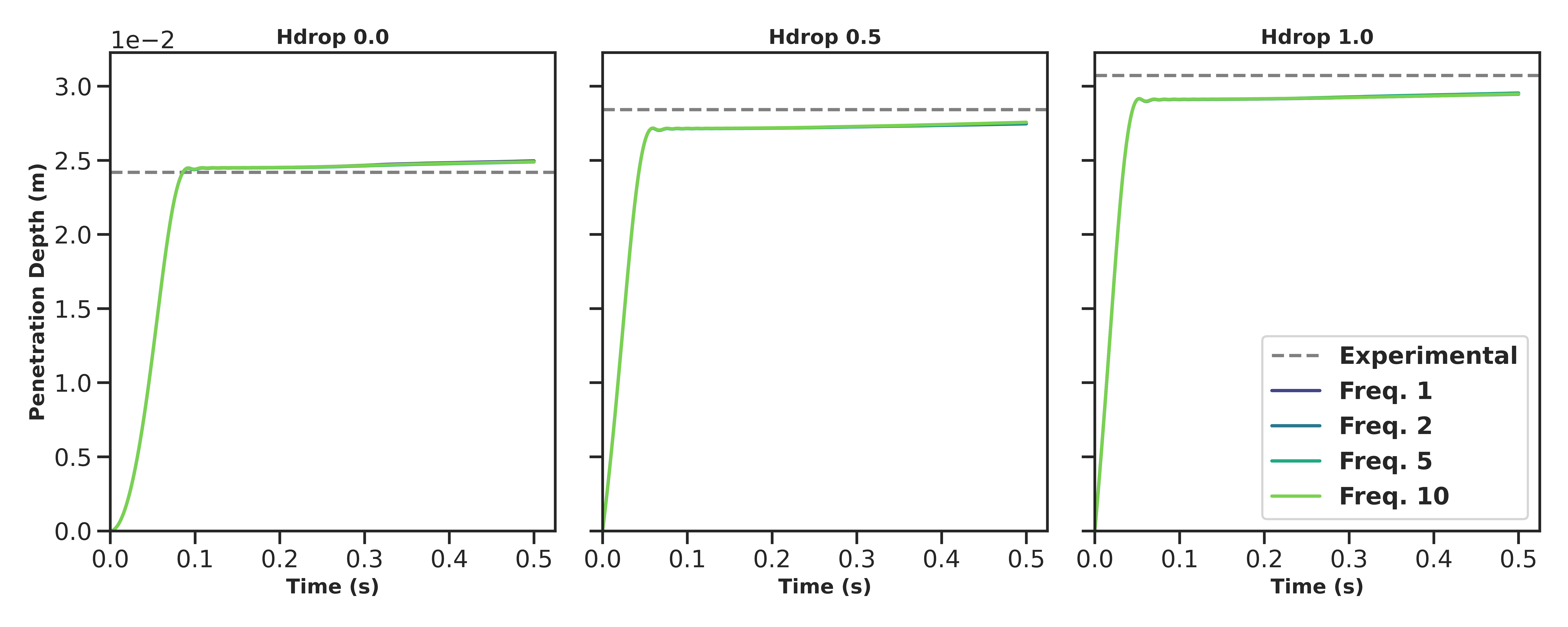}
        \caption{Cone Penetration in Glass Beads}
        \label{fig:bead_ps_comp}
    \end{subfigure}
    \hfill
    \begin{subfigure}[t]{0.7\textwidth}
        \centering
        \includegraphics[width=\textwidth]{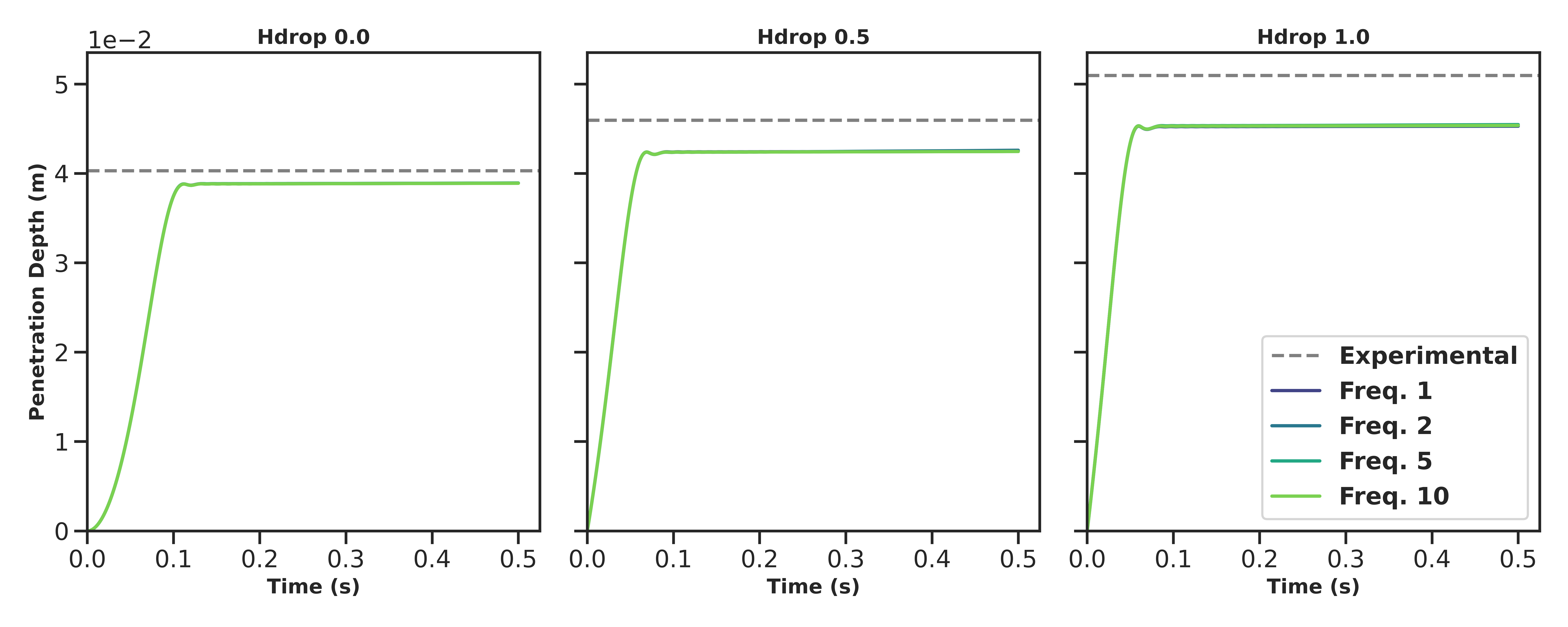}
        \caption{Cone Penetration in Ottawa Sand.}
        \label{fig:sand_ps_comp}
    \end{subfigure}
    \caption{Additional simulation results for the cone penetration benchmarks using different neighbor list update frequencies ($ps_{\text{freq}}$) ranging from 1 (update every step) to 10.}
    \label{fig:persistent_neighbor_list_accuracy_2}
\end{figure}

To demonstrate this, we redo the validation benchmarks discussed in Sec.~\ref{sec:validation} and vary $ps_{\text{freq}}$ from 1 to 10.  Figures~\ref{fig:persistent_neighbor_list_accuracy} and~\ref{fig:persistent_neighbor_list_accuracy_2} show that there is no loss in accuracy in the simulation results even when $ps_{\text{freq}} = 10$, across all test cases.

% ========================================

\subsubsection{Performance Improvements from Persistent Neighbor Lists}
\label{sec:persistent_neighbor_list_speedup}
To quantify the performance benefit of using persistent neighbor lists, we measure the total wall-clock time ($t_{\text{wall}}$) required to complete a fixed duration of simulated time ($t_{\text{sim}}$) and define the Real-Time Factor (RTF) as $RTF = t_{\text{wall}} / t_{\text{sim}}$. An RTF of 1.0 signifies real-time performance; $\text{RTF} < 1.0$ indicates faster-than-real-time execution, while $\text{RTF} > 1.0$ indicates slower-than-real-time execution. Although RTF provides context, the key metric for our purpose is the reduction in wall-clock time by not updating the neighbor lists every time step.

\begin{figure}[htbp]
    \centering
    \includegraphics[width=0.7\textwidth]{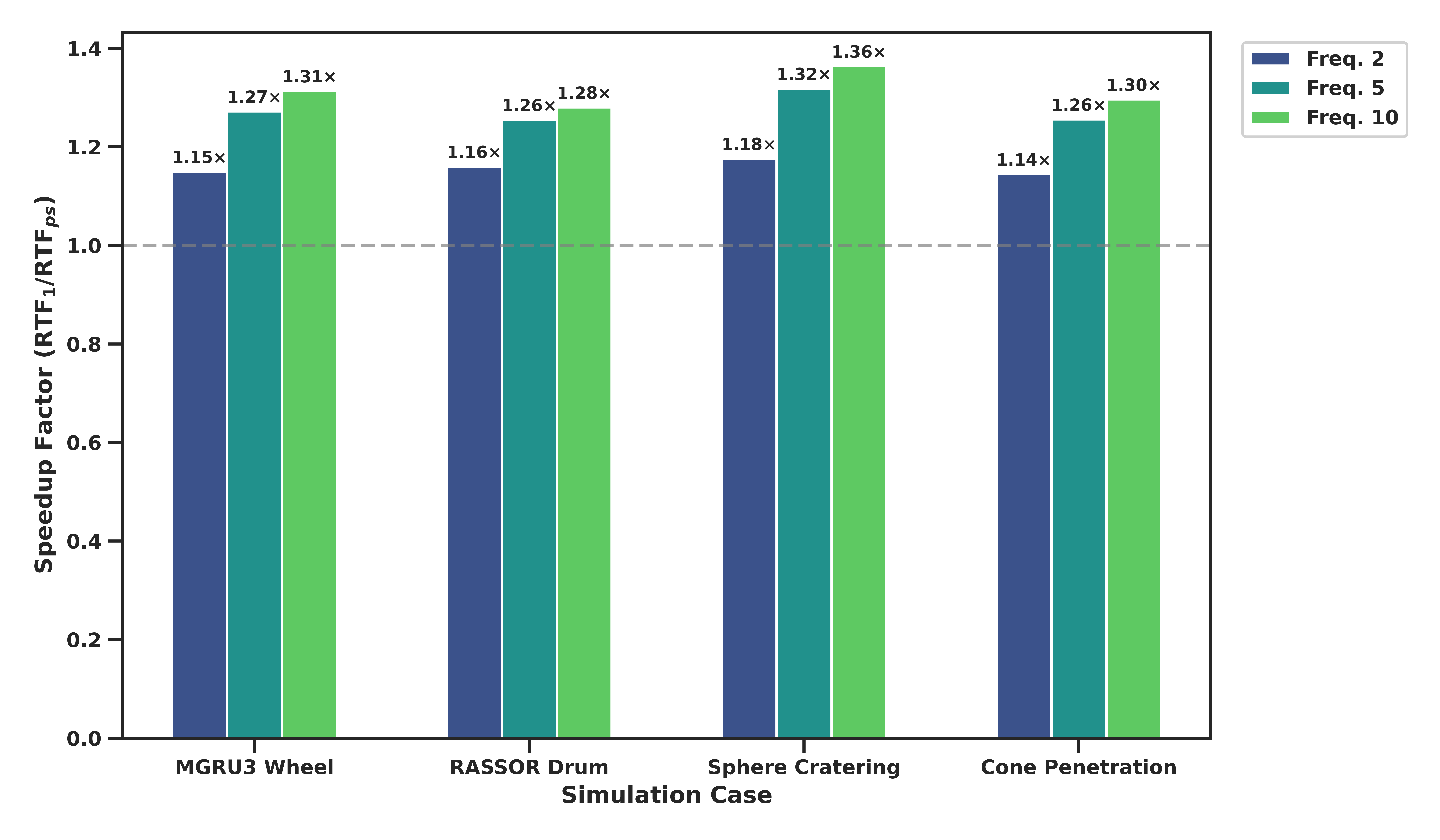}
    \caption{Speedup factor (RTF$_{1}$/RTF$_{ps}$) for different simulations at varying neighbor list update frequencies. Higher values indicate better performance, with the baseline being RTF$_{1}$ (updating the neighbor lists at every time step).}
    \label{fig:ps_freq_speedup_acc}
\end{figure}

Figure~\ref{fig:ps_freq_speedup_acc} presents the speedup factor RTF$_{1}$/RTF$_{ps}$ for the validation benchmarks from Sec.~\ref{sec:validation}, where RTF$_{1}$ represents the RTF for the baseline case $ps_{\text{freq}}=1$.
The results demonstrate that reusing the neighbor lists yields good computational savings across all test cases --- setting $ps_{\text{freq}} = 10$ reduces execution time by 23--27\% (speedup of 1.28$\times$-1.36$\times$). All simulations were run on a single Nvidia RTX 4080Ti GPU with 16 GB of memory.

% =========================================================

\subsection{Active Domains}
Sound travels much slower in loose soil ($\approx$100--300 \si{m/s}) than in water ($\approx$1480 \si{m/s}) because stress is carried only through weak, pressure-dependent contacts between grains. Consequently, when an object pushes through the soil, the disturbance dies out quickly and the movement stays limited to a relatively thin region around the object~\cite{HeGranularForces2025}.

\begin{figure}[htbp]
    \centering
    \includegraphics[width=0.3\textwidth]{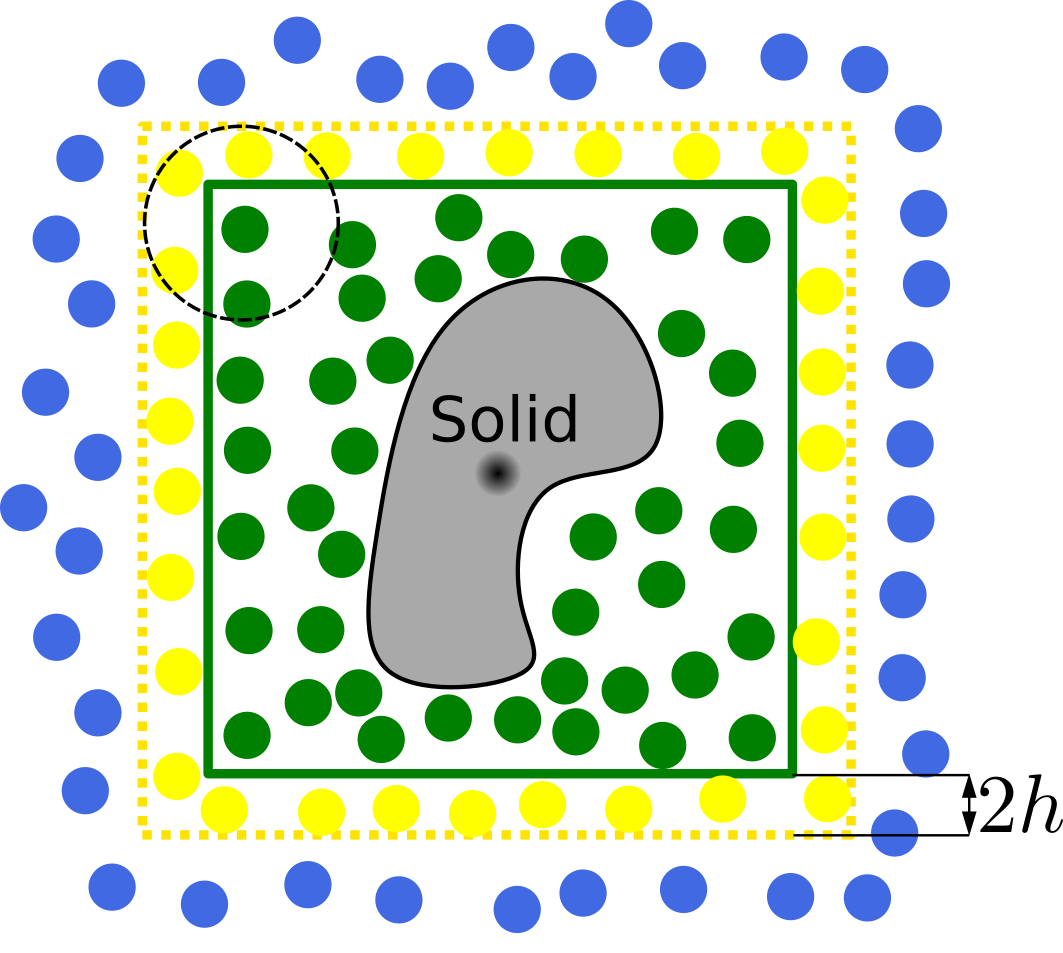}
    \caption{An illustration of the active domain in 2D. The ``active box'' OOBB (green box) moves with the solid body. SPH particles within the ``active box'' are flagged as \textit{Active}. For computing properties of SPH particles at the boundary of the ``active box'', the algorithm considers the contributions from the particles within a distance $2h$ of it (yellow dotted box). These particles (yellow) are flagged as \textit{Extended-Active}. All other particles are set as \textit{Inactive} (blue) and not processed nor stored.}
    \label{fig:active_domain}
\end{figure}

We implement so-called ``active domains'' to selectively activate and deactivate SPH particles in the simulation domain based on their proximity to moving bodies. Users can specify object-oriented bounding boxes (OOBB), called ``active boxes'' associated with any Chrono solid body, that are rigidly attached to the body and move with the body during the simulation. SPH particles within thess active boxes are activated, i.e., they participate in neighbor search and their states are updated. SPH particles outside active boxes are deactivated, they are not included in neighbor search, their states are not updated, and the GPU memory allocated for them is freed. The algorithm is illustrated in Algorithm~\ref{alg:step_with_active_domains} and detailed as follows:

\begin{enumerate}
    \item During initialization, the user specifies the length, width and height of the ``active boxes'' that are positioned at the local origin of each solid body in the simulation. Additionally, the user can also specify an ``active box delay'' ($t_\text{{delay}}$) parameter, which determines the time delay before the active domains are activated.
    \item At relevant time steps (when $t > t_\text{{delay}}$), particle activity is updated based on the position relative to the ``active boxes''. An \texttt{UpdateActivity} kernel assigns one of three flags to each particle: \textit{Active}, \textit{Extended-Active}, or \textit{Inactive}. Particles residing within an ``active box'' are flagged as \textit{Active}. Since SPH calculations for active particles near the box boundary require contributions from neighbors within their kernel support radius, particles outside an ``active box'' but within a distance $2h$ of its boundary are flagged as \textit{Extended-Active} to ensure their data is available for these calculations. All remaining particles are flagged as \textit{Inactive}. See Fig.~\ref{fig:active_domain} for an illustration.
    \item $N_a$ and $N_e$ denote the number of active and extended-active particles, respectively, computed using an \textit{inclusive scan} on the GPU. To optimize memory usage, the primary data arrays are dynamically managed to store data only for the active and extended-active particles, totaling $N_{a+e} = N_a + N_e$ elements required at a given time step. This strategy distinguishes between the arrays' logical size (the $N_{a+e}$ elements currently in use) and their allocated memory capacity (the total memory reserved). When the required size $N_{a+e}$ exceeds the current capacity, that is increased proactively to $N_{a+e} \times G$, where $G$ is a growth factor (empirically set to 1.2). This provides a buffer for possible future array growth and minimizes frequent re-allocations. Following any potential capacity change, the logical size of the arrays is always adjusted to match the exact requirement $N_{a+e}$. To reclaim excessive unused memory, a periodic shrinking mechanism is employed. Every $S_I$ time steps (shrink interval, empirically set to 50), the memory utilization is checked by calculating the ratio of the current size ($N_{a+e}$) to the allocated capacity. If this ratio falls below a shrink threshold $S$ (empirically set to 0.75), the arrays are shrunk in an attempt to reduce the allocated capacity down to the current logical size $N_{a+e}$. The parameter values $G=1.2$, $S=0.75$, and $S_I=50$ were found to provide robust and efficient memory management across all tested simulation scenarios.
    \item Following the activity update and memory allocation, subsequent processing focuses only on the $N_{a+e}$ active and extended-active particles. These particles are sorted for neighbor search and the neighbor lists are constructed using the method described in Sec.~\ref{sec:persistent_neighbor_list}, potentially reusing older lists.
    \item Once the neighbor lists for the active set are available, forces are computed and particle states (such as position and velocity) are updated, considering only these $N_{a+e}$ particles. Consequently, the computational kernels responsible for force calculation and dynamics updates are launched using $N_{a+e}$ threads. This contrasts with the baseline approach, which required launching these kernels with $N$ threads (where $N$ is the total number of particles), thus providing computational savings when $N_{a+e} \ll N$.
\end{enumerate}

% Define additional algorithm functions for Active Domains
\SetKwFunction{KwUpdateActivity}{UpdateActivity}
\SetKwFunction{KwComputeActiveCount}{ComputeActiveCount}
\SetKwFunction{KwManageArrayMemory}{ManageArrayMemory}
\SetKwFunction{KwBuildNeighborList}{BuildNeighborList}
\SetKwFunction{KwComputeForces}{ComputeForces}
\SetKwFunction{KwUpdateStates}{UpdateStates}

\begin{algorithm}[H] % Or allow floating
    \caption{Simulation Step with Active Domains}\label{alg:step_with_active_domains}
    \KwIn{Current state ($pos$, $vel$), time $t$, total particles $N$, parameters ($t_{delay}$, ActiveBox, $ps_{freq}$, $G, S, S_I$, $dt$)}
    \KwOut{Updated state ($pos$, $vel$)}
    % Internal state: neighborList, array_capacity, activity_flags

    \BlankLine
    $N_{process} \leftarrow N$\; \tcp{Default: process all N particles}
    \If{$t > t_{delay}$}{
        \tcp{1. Flag particles based on active boxes}
        $activity\_flags \leftarrow \KwUpdateActivity(pos, ActiveBox)$\;
        \tcp{2. Count Active ($N_a$) and Extended-Active ($N_e$) particles}
        $N_{a+e} \leftarrow \KwComputeActiveCount(activity\_flags)$\;
        \tcp{3. Resize arrays based on $N_{a+e}$ and memory strategy}
        % Handles capacity changes (G, S, S_I) and sets logical size to $N_{a+e}$
        $\KwManageArrayMemory(N_{a+e}, G, S, S_I)$\;
        $N_{process} \leftarrow N_{a+e}$\; \tcp{Subsequent steps use the reduced set}
    }
    \Else{
         \tcp{Activity culling inactive, ensure arrays process N particles}
         % Ensure capacity and logical size are suitable for N particles
         $\KwManageArrayMemory(N, \dots)$\;
         $N_{process} \leftarrow N$\;
    }

    \BlankLine
    \tcp{4. Build/Update Neighbor Lists for $N_{process}$ particles}
    \If{$t \pmod{ps_{freq}} = 0$}{
         $cellStart, cellEnd, offset \leftarrow \KwPreprocessNeighborList(pos[0..N_{process}-1])$\;
         $neighborList \leftarrow \KwCall{Algorithm~\ref{alg:neighbor_list}}{pos[0..N_{process}-1], cellStart, cellEnd, offset}$\;
    }
    % Else: Reuse existing neighborList from previous step where it was built

    \BlankLine
    \tcp{5. Compute Forces \& Update States for $N_{process}$ particles}
    % Kernels launched with N_process threads
    $forces \leftarrow \KwComputeForces(pos, vel, neighborList, N_{process})$\;
    $\KwUpdateStates(pos, vel, forces, N_{process}, dt)$\;

\end{algorithm}

We next evaluate the accuracy and performance of the active domain approach compared to the baseline simulation (without active domains), focusing initially on the MGRU3 and RASSOR test cases. Finally, subsection~\ref{sec:overall_performance_improvements} provides an overall comparison against the previous Chrono SPH implementation~\cite{weiGranularSPH2021}, which rebuilt neighbor lists for the force computation of each equation and at each time step.

\subsubsection{Accuracy of Active Domains}
\label{sec:active_domains_accuracy}
\begin{figure}[htbp]
    \centering
    \begin{subfigure}[b]{0.48\textwidth}
        \centering
        \includegraphics[width=\textwidth]{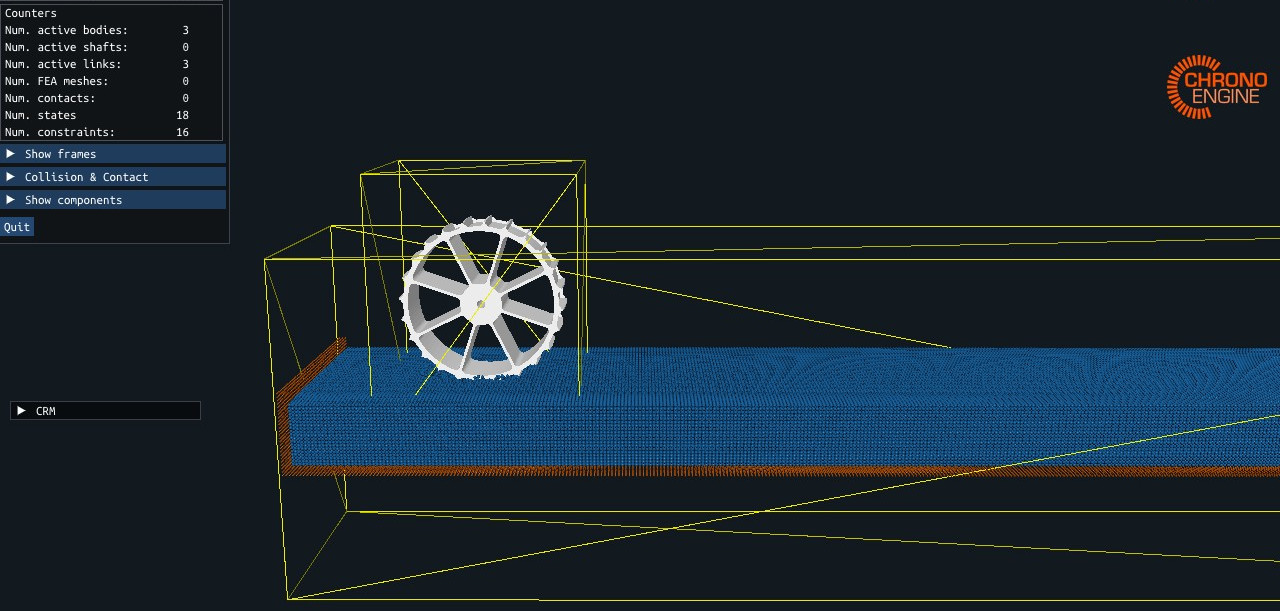}
        \caption{MGRU3 wheel test case. Active Box: $0.6 \times 0.6 \times 0.8$ \si{m}.}
        \label{fig:mgr_active_box}
    \end{subfigure}
    \hfill % Use \hfill for horizontal spacing
    \begin{subfigure}[b]{0.48\textwidth}
        \centering
        \includegraphics[width=\textwidth]{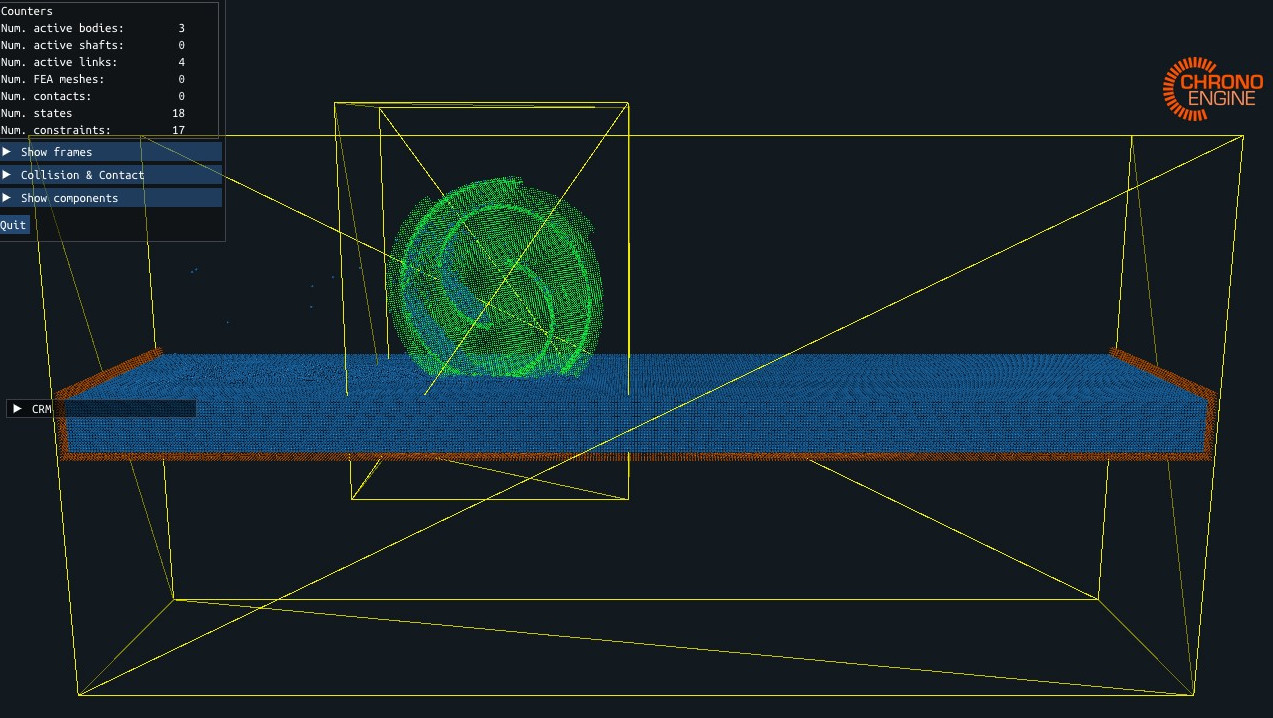}
        \caption{RASSOR drum test case. Active Box: $0.5 \times 0.4 \times 0.7$ \si{m}.}
        \label{fig:rassor_active_box}
    \end{subfigure}
    \caption{Illustration of the active domain partitioning for the validation test cases. The yellow ``active box'' is shown surrounding the primary object, MGRU3 wheel and RASSOR drum, respectively. Particles inside this box (\textit{Active}) and those within $2h$ of its boundary (\textit{Extended-Active}) are processed during the simulation step. Blue spheres represent SPH particles, while brown spheres are boundary BCE markers. The larger yellow box depicts the computational domain of the simulation.}
    % \SBELcomment{Decide whether to talk about periodic boundaries here?}}
    \label{fig:active_domains}
\end{figure}

We redo the MGRU3 and RASSOR test cases from Sec.~\ref{sec:validation} using the active domain approach and compare results with simulations performed without active domains. For the MGRU3 test case, we set the ``active boxes'' dimension to \SI{0.6}{m} $\times$ \SI{0.6}{m} $\times$ \SI{0.8}{m} (see Fig.~\ref{fig:mgr_active_box}). This size covers the entire terrain depth and includes a significant portion of the terrain width and length around the wheel. We also use a $t_{delay} = 1$ \si{s} to ensure that the terrain has sufficient time to settle before active domains are activated. For the RASSOR test case, we set the size of ``active boxes'' to \SI{0.5}{m} $\times$ \SI{0.4}{m} $\times$ \SI{0.7}{m} (see Fig.~\ref{fig:rassor_active_box}).

\begin{figure}[htbp]
    \centering
    \begin{subfigure}[b]{0.48\textwidth}
        \centering
        \includegraphics[width=\textwidth]{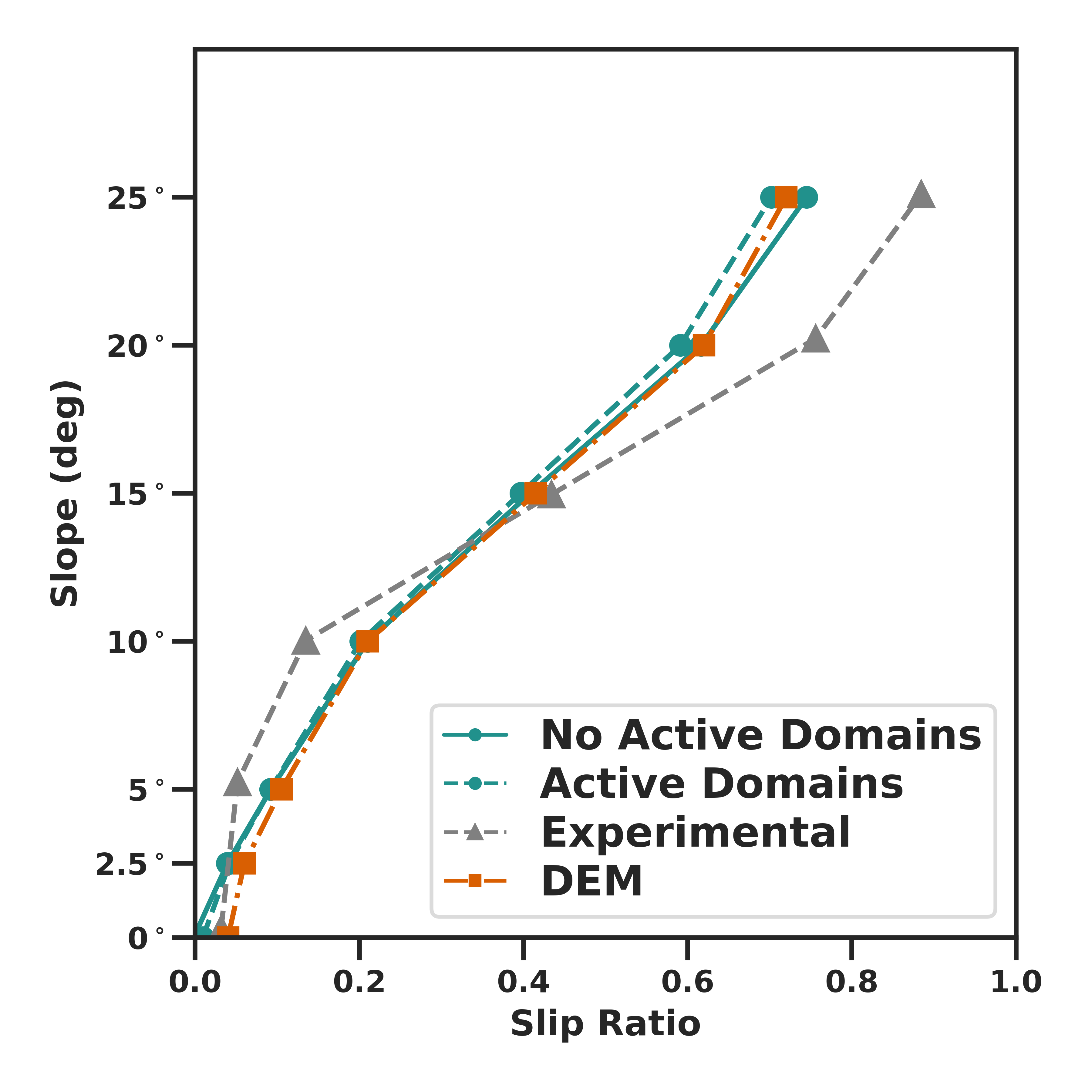}
        \caption{MGRU3 wheel test case.}
        \label{fig:mgr_active_domain_accuracy}
    \end{subfigure}
    \hfill % Use \hfill for horizontal spacing
    \begin{subfigure}[b]{0.48\textwidth}
        \centering
        \includegraphics[width=\textwidth]{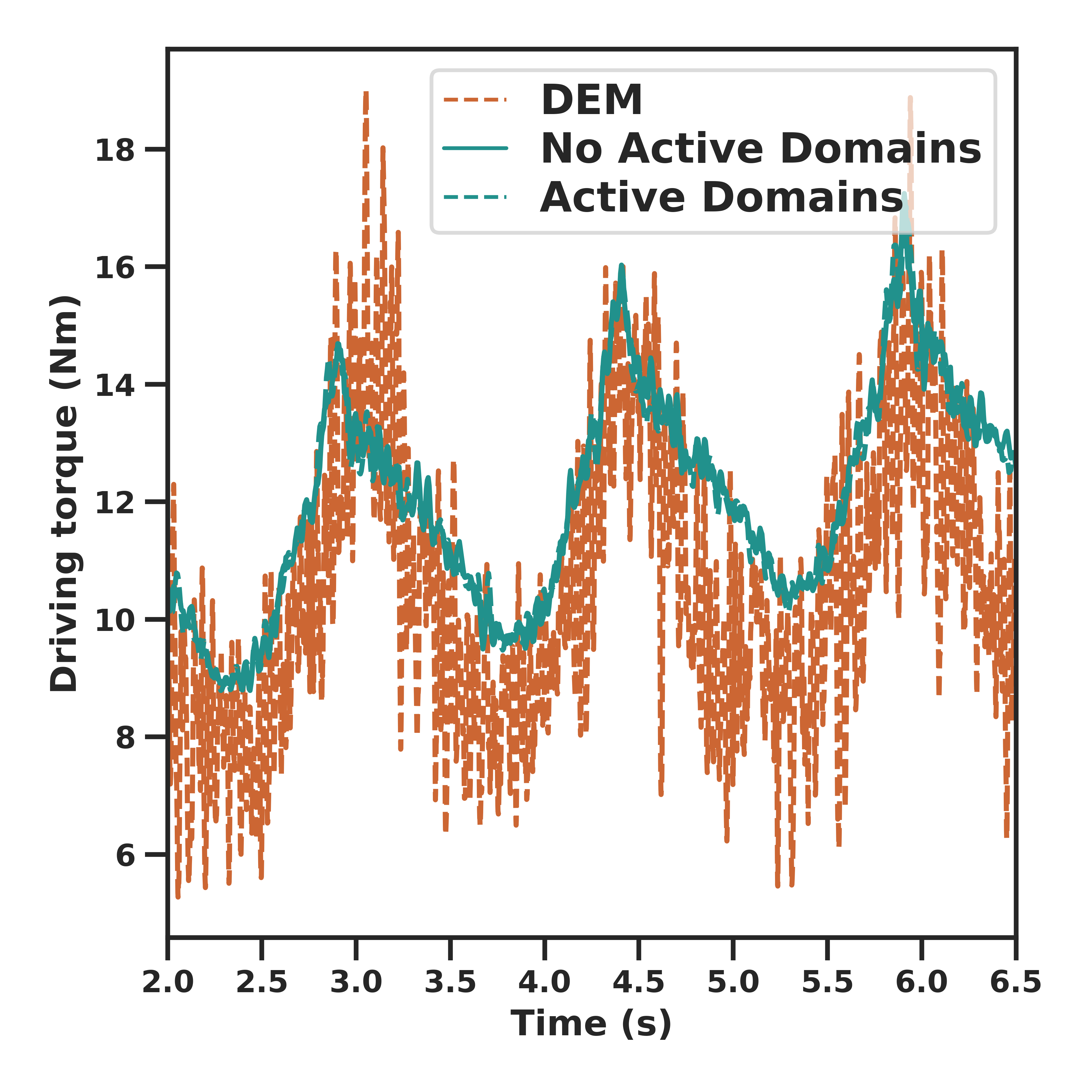}
        \caption{RASSOR drum test case.}
        \label{fig:rassor_active_domain_accuracy}
    \end{subfigure}
    \caption{Comparison of results using active domains (dashed green lines) and without active domains (solid green lines) for the MGRU3 and RASSOR test cases. The results closely match the baseline approach, indicating minimal loss in accuracy due to the active domain optimization. The performance improvements are shown in Table~\ref{tab:active_domains_performance}.}
    \label{fig:active_domains_accuracy} % Combined label for the figure
\end{figure}

We find that using active domains does not introduce any significant loss in accuracy, as shown by the comparison plots in Fig.~\ref{fig:active_domains_accuracy}.

% ===============================

\subsubsection{Performance of Active Domains}
\label{sec:active_domains_performance}
Here we discuss the performance benefits of using active domains for the MGRU3 and RASSOR test cases. To isolate the performance improvement solely due to the active domain optimization, the persistent neighbor list feature was disabled for these runs; i.e., $ps_{\text{freq}} = 1$. Table~\ref{tab:active_domains_performance} shows the resulting performance comparison. We obtain speedups larger than $2$ for both the MGRU3 and RASSOR test cases.

\begin{table}[h]
    \centering
    % Note: Consider using booktabs for nicer rules: \toprule, \midrule, \bottomrule
    \caption{RTF improvement due to active domains for the MGRU3 and RASSOR test cases (with $ps_{\text{freq}} = 1$).}
    \label{tab:active_domains_performance}
    \begin{tabular}{|l|cc|c|}
    \hline
    \multirow{2}{*}{\textbf{Simulation}} & \multicolumn{2}{c|}{\textbf{RTF}} & \multirow{2}{*}{\textbf{Speedup Factor}} \\ \cline{2-3}
     & \textbf{No Active Domains} & \textbf{Active Domains} &  \\ \hline
    MGRU3 Wheel & 59.36 & 20.25 & 2.93$\times$ \\
    RASSOR Drum & 148.66 & 70.49 & 2.11$\times$ \\
    \hline
    \end{tabular}
\end{table}

% ===================================================

\subsection{Overall Performance Improvements}
\label{sec:overall_performance_improvements}
We evaluate the performance of the optimized Chrono::CRM solver against two Chrono options for deformable terrain simulation: (1) the previous SPH implementation based on the granular model in~\cite{weiGranularSPH2021} and (2) the Chrono::SCM semi-empirical terramechanics model~\cite{chronoSCM_JCND_2023}. 

SCM, which employs Bekker-Wong pressure-sinkage relationships~\cite{bekker69} and Janosi-Hanamoto shear stress formulations~\cite{janosi61}, is computationally efficient and thus widely used in Chrono for off-road  mobility studies. However, its reliance on empirical parameters (e.g., soil cohesion $c$, friction angle $\phi$, shear modulus $k$) requires experimental calibration and can limit predictive accuracy across varying soil conditions. 
SCM has difficulties in high-slip scenarios and in simulations involving soil flow. This is because the underlying semi-empirical model, in which the soil is effectively seen as a force element, does not account for soil inertia nor for soil movement within the contact patch. Both CRM and SCM may have difficulty with complex geometries, especially when these exhibit fine features (such as very thin wheel grousers); this is because of the inherent spatial resolution introduced by the inter-particle spacing in CRM and the grid cell size in SCM.

%SCM is challenged in high-slip scenarios (since it does not account for the inertia of the soil), handling deformable wheels (owing to a continuous change of the contact patch shape and area), when dealing with non-trivial grouser geometries, in steering scenarios, in banked traverses and steep climbs (when the pressure in the contact patch is not uniform and non-trivially distributed), and in digging, grading, and bulldozing scenarios (again, because the model does not account for the inertia of the soil, which in SCM is considered a force element only, like an elastic foundation).

\begin{figure}[htbp]
    \centering
    \includegraphics[width=1\textwidth]{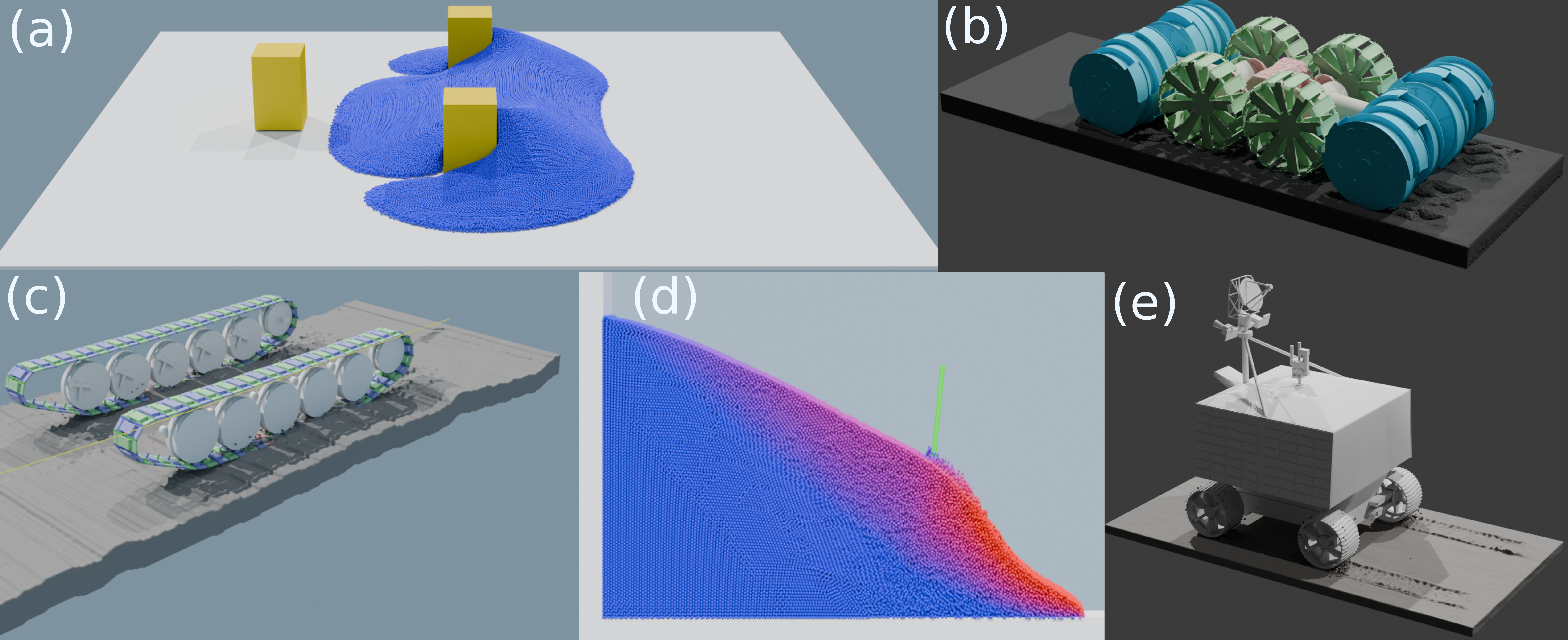}
    \caption{Rendered views of the benchmark test cases used for performance comparison. (a) Baffle Flow, (b) RASSOR, (c) Tracked Vehicle, (d) Flexible Cable, (e) VIPER.}
    \label{fig:performance_collage}
\end{figure}

The following benchmark test cases are used for comparison:
\begin{itemize}
    \item \textbf{Baffle Flow:} Granular material ($0.6 \times 2.4 \times 0.42$ \si{m}) is initialized on a plane ($4.8 \times 4.2$ \si{m}) with three fixed rigid baffles ($0.3 \times 0.3 \times 0.48$ \si{m} each) acting as obstacles. The material starts with an initial velocity of \SI{1.5}{m/s} along the positive x-axis. Simulation runs for 1 second. Parameters: $d_0 = 0.01$ \si{m}, $h = 1.2 d_0$, $\Delta t = 1.0 \times 10^{-4}$ \si{s}. See Fig.~\ref{fig:performance_collage}(a).

    \item \textbf{RASSOR:} A RASSOR rover model (four wheels, two counter-rotating drums) is placed on a rectangular granular terrain ($3.0 \times 1.0 \times 0.1$ \si{m}) patch. The rover moves forward at \SI{0.2}{m/s} with specific wheel rotation, drum oscillation, and drum rotation speeds for 1 second. Parameters: $d_0 = 0.004$ \si{m}, $h = 1.2 d_0$, $\Delta t = 2.0 \times 10^{-4}$ \si{s}. See Fig.~\ref{fig:performance_collage}(b).
   
    \item \textbf{Tracked Vehicle:} An M113 tracked vehicle model follows a predefined \SI{30}{m} long S-shaped path on granular terrain. The terrain geometry is defined by SPH particles loaded from external files. The vehicle targets a speed of \SI{7.0}{m/s} using a PID controller. Parameters: $d_0 = 0.02$ \si{m}, $h = 1.2 d_0$, $\Delta t = 5.0 \times 10^{-4}$ \si{s}. This scenario is also simulated using SCM with the same time step but a terrain mesh resolution of $0.04$ \si{m} (twice the particle spacing $d_0$ used in Chrono::CRM). The simulation is run for five seconds. See Fig.~\ref{fig:performance_collage}(c).

    \item \textbf{Flexible Beam:} A granular column ($1 \times 0.2 \times 1$ \si{m}) initialized in a container ($3 \times 0.2 \times 0.2$ \si{m}) flows and interacts with a flexible beam (height $0.8$ \si{m}, modeled using ANCF~\cite{shabana2020}) positioned $1.7$ \si{m} from the container's back wall. Simulation runs for 2 seconds. Parameters: initial particle spacing $d_0 = 0.01$ \si{m}, smoothing length $h = 1.2 d_0$, time step $\Delta t = 2.5 \times 10^{-4}$ \si{s}. See Fig.~\ref{fig:performance_collage}(d).

    \item \textbf{VIPER:} The rover is positioned on a rectangular granular terrain ($4.0 \times 2.0 \times 0.1$ \si{m}) patch with bulk density \SI{1700}{kg/m^3}. Simulation runs for 1 second. Parameters: $d_0 = 0.01$ \si{m}, $h = 1.2 d_0$, $\Delta t = 2.5 \times 10^{-4}$ \si{s}. This scenario is also simulated using SCM with the same time step but a terrain mesh resolution of $0.02$ \si{m} (twice the particle spacing $d_0$ used in Chrono::CRM). See Fig.~\ref{fig:performance_collage}(e).

\end{itemize}

For the CRM simulations we use an Nvidia RTX 4080 Ti GPU with 16 GB of memory. For the CPU-based Chrono::SCM simulation, we use a 13th Gen Intel(R) Core(TM) i7-13700K CPU with 32 GB of memory. We compare the RTF values for the following configurations:
\begin{itemize}
    \item \textbf{Baseline}: This refers to the previous Chrono SPH implementation based on the work by Wei et al.~\cite{weiGranularSPH2021}. We note that this baseline Chrono SPH code included a simple ``active domains'' implementation - however, that did not include any of the memory management described above and therefore any computational savings were due to early exits from CUDA kernels (for particles found outside an active box). To isolate and highlight the performance gains stemming from the more sophisticated ``active domains'' implementation and the persistent neighbor lists in the current Chrono::CRM code, we decided to deactivate use of active boxes in the baseline case, placing it on the same footing as case (a) below.

    \item \textbf{Chrono::CRM Configurations}: Three variations of the current solver are evaluated:
    \begin{enumerate}[(a)]
        \item \boldmath{$ps = 1$}: Rebuilds the neighbor lists every time step and does not use the active domain optimization. This serves as the baseline performance for the new CRM solver itself.
        \item \boldmath{$ps = 10$}: Reuses the neighbor lists for 10 time steps, but does not employ the active domain optimization.
        \item \boldmath{$ps = 10$} \textbf{- active}: Reuses the neighbor lists and employs the active domain optimization. This configuration represents the fully optimized Chrono::CRM approach evaluated in this study, whose accuracy was validated in previous sections.
    \end{enumerate}

    \item \textbf{SCM Configurations}: Chrono::SCM includes an optimization also called ``active domains'', which limits contact detection computations to SCM grid nodes within the projection of interacting body OOBBs onto the SCM reference plane~\cite{chronoSCM_JCND_2023}. This mechanism is conceptually similar to the active domain approach in Chrono::CRM. We test two SCM versions, \textit{default} and \textit{active}, without and with the SCM active domains optimization enabled, respectively.
    When using the SCM active domains, the OOBBs can more tightly conform to the rigid body geometry; unlike Chrono::CRM they do not need to account for additional interactions with soil outside the active domains.  We therefore set the following SCM active domains:
    \begin{itemize}
       \item For VIPER, four active domains are employed, one for each wheel, with dimensions $1.25D \times W \times 1.25D$, where $D$ is the diameter and $W$ is the width of the rover wheel.
       \item For the Tracked Vehicle, two active domains are used, one corresponding to each track, with dimensions of $10 \times 1.2 \times 2$ \si{m}.
    \end{itemize}
\end{itemize}

Details on the problem size and integration step-size for the five benchmark problems are provided in Table~\ref{tab:particle_counts}. In all these simulations, the same step-size was used for integration of both multibody and fluid dynamics and co-simulation communication was done after each integration step.

\begin{figure}[htbp]
    \centering
    \includegraphics[width=0.8\textwidth]{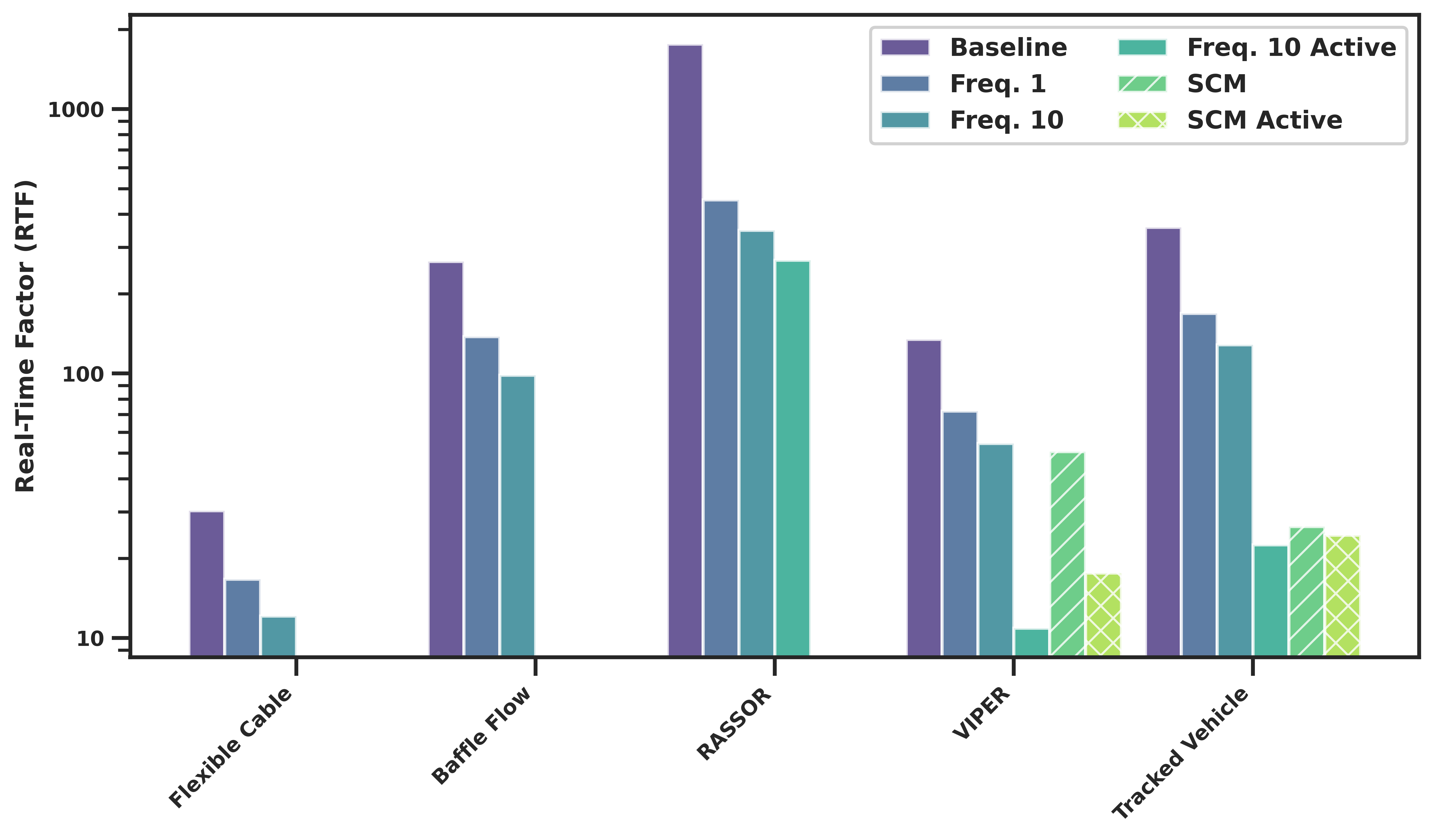}
    \caption{Comparison of RTF performance across benchmarks for different simulation configurations. The RTF-axis uses a logarithmic scale; lower bars indicate better performance (faster execution). Bars represent the Baseline~\cite{weiGranularSPH2021}, Chrono::CRM variants ((a), (b), and (c)), and SCM variants (without and with SCM active domains enabled).}
    \label{fig:overall_performance_improvements}
\end{figure}

The performance results are summarized in Table~\ref{tab:overall_performance_improvements}, which shows the RTF for each configuration and the corresponding speedup factor relative to the baseline. The speedup factor is calculated as $S = RTF_{\text{Baseline}} / RTF_{\text{case}}$, where $RTF_{\text{Baseline}}$ is the RTF of the previous SPH implementation~\cite{weiGranularSPH2021} and $RTF_{\text{case}}$ is the RTF for one of the newly evaluated configurations. A visual comparison of the speedup is provided in Fig.~\ref{fig:overall_performance_improvements}. 

Compared to the baseline SPH implementation~\cite{weiGranularSPH2021}, using the new Chrono::CRM solver with rebuilding the neighbor lists at each step (case (a) above) provides speedups of $1.8\times$ to $3.9\times$ across the test cases. The higher speedups in this comparison ($2.1\times$ for Tracked Vehicle, $3.9\times$ for RASSOR) occur in the simulations with larger particle counts (Table~\ref{tab:particle_counts}). This speedup is predominantly due to computing and storing the neighbor lists rather than recomputing them for each force evaluation.
Reusing the neighbor lists for 10 steps within the new solver (case (b)) increases the speedup relative to the baseline to over $2.5\times$ for all tests. Adding the active domain optimization where applicable (case (c)) results in speedups relative to the baseline exceeding $7\times$ (for RASSOR, VIPER, Tracked Vehicle). The $15.9\times$ speedup for the Tracked Vehicle case is explained by its low ratio of active-to-total particles.

When compared to the Soil Contact Model (SCM), the Chrono::CRM case (c) configuration has lower RTF values (indicating faster execution) than the optimized SCM with active domains for both the VIPER and Tracked Vehicle. We note however, that the comparison of CRM vs. SCM is provided here only as a rough estimate of the relative performance of the two approaches to modeling vehicle interaction with deformable terrain. While SCM simulations could be accelerated using a coarser SCM grid spacing (in these experiments, the SCM grid resolution was set to twice the CRM particle separation), CRM simulations will nonetheless always be more physically accurate. 

\begin{table}[htbp]
    \centering
    % Consider using booktabs for nicer table formatting (\toprule, \midrule, \bottomrule)
    \caption{Particle counts and time step for the benchmark simulations. ``Total Particles'' includes all SPH and BCE particles. ``Active Particles'' indicates the approximate number of particles processed when using the active domain optimization (where applicable).}
    \label{tab:particle_counts}
    \begin{tabular}{|l|r|r|r|}
    \hline
    \textbf{Name} & \textbf{Time Step (s)} & \textbf{Total Particles} & \textbf{Active Particles} \\ \hline % Added units to Time Step
    Flexible Beam & 2.5e-04 & 259,721 & 259,721 \\ %\hline
    Baffle Flow & 1.0e-04 & 1,311,628 & 1,311,628 \\ %\hline
    RASSOR & 2.0e-04 & 6,414,673 & 4,597,696 \\ %\hline
    VIPER & 2.5e-04 & 1,239,664 & 280,682 \\ %\hline
    Tracked Vehicle & 5.0e-04 & 5,854,120 & 492,840 \\ \hline % Check if this count is correct from table 1 data
    \end{tabular}
\end{table}

\begin{table}[htbp]
    \centering
    % Consider using booktabs and siunitx column types for better alignment/spacing
    \caption{Comparison of RTF values (lower is better) across benchmarks. Speedup factors relative to the Baseline~\cite{weiGranularSPH2021}, calculated as $\text{RTF}_\text{Baseline} / \text{RTF}_\text{case}$, are shown in parentheses. For the current CRM implementation, results are provided for neighbor list updates every step, every 10 steps, and every 10 steps with active domains enabled. Where applicable, results for the Chrono::SCM deformable terrain are provided with and without SCM active domains enabled~\cite{chronoSCM_JCND_2023}.}
    \label{tab:overall_performance_improvements}
    \begin{tabular}{|l||r||r|r|r||r|r|}
    \hline
    \multirow{2}{*}{\textbf{Name}} & \multirow{2}{*}{\textbf{Baseline}} & \multicolumn{3}{c||}{\textbf{Chrono::CRM}} & \multicolumn{2}{c|}{\textbf{Chrono::SCM}} \\ \cline{3-7}
    & & \boldmath{$ps = 1$} & \boldmath{$ps = 10$} & \boldmath{$ps = 10$} \textbf{- active} & \textbf{default} & \textbf{active} \\ \hline
    Flexible Beam & 30.3 & 16.7 (1.8x) & 12.1 (2.5x) & - & - & - \\ %\hline
    Baffle Flow & 265.3 & 137.8 (1.9x) & 98.6 (2.7x) & - & - & - \\ %\hline
    RASSOR & 1761.1 & 453.4 (3.9x) & 348.2 (5.1x) & 268.3 (6.6x) & - & - \\ %\hline
    VIPER & 134.8 & 72.1 (1.9x) & 54.4 (2.5x) & 10.9 (12.4x) & 50.7 (2.7x) & 17.6 (7.7x) \\ %\hline
    Tracked Vehicle & 356.9 & 169.0 (2.1x) & 128.4 (2.8x) & 22.5 (15.9x) & 26.4 (13.5x) & 24.5 (14.6x) \\ \hline
    \end{tabular}
\end{table}

\section{Solver Scalability}
\label{sec:scalability}
High-fidelity simulations of large-scale granular materials have traditionally been challenging due to the high computational cost associated with resolving particle interactions. Modeling such systems using DEM, while accurate, often becomes computationally prohibitive for problems involving millions of particles~\cite{fransen2025MLgranular}, as the required particle counts can easily exceed current hardware capabilities.

A notable example is the simulation of VIPER on GRC1 lunar simulant: using a volume \SI{4}{m} $\times$ \SI{2}{m} $\times$ \SI{1}{m} of granular material, a three-second simulation required approximately 11.5 million DEM elements (represented by roughly 35.5 million component spheres) and took 109 hours to compute on two Nvidia A100 GPUs with 40 GB memory each~\cite{ruochunGRC-DEM2023}. In contrast, simulating the same scenario with Chrono::CRM involved approximately 1.2 million SPH particles and completed in 30 seconds on a single Nvidia RTX 4080 Ti GPU with 16 GB of memory (see Sec.~\ref{sec:performance} for more details).

This difference in simulation times highlights the potential of Chrono::CRM as a practical approach for large-scale terramechanics problems. To further assess this potential, we evaluate the scaling properties of the Chrono::CRM solver using a vehicle simulation. For this study, we model a Polaris RZR vehicle using Chrono::Vehicle~\cite{chronoVehicle2019} traversing terrain simulated with Chrono::CRM. The vehicle is equipped with rigid tires (radius \SI{0.33}{m}, width \SI{0.21}{m}). The terrain has a density of \SI{1700}{kg/m^3}, cohesion of \SI{5000}{Pa}, and static friction coefficient $\mu_s = 0.8$. For the SPH discretization, the initial particle spacing is set to $d_0 =$ \SI{0.04}{m} and the smoothing length $h =$ \SI{0.048}{m}. We employ the optimized settings for Chrono::CRM, with a persistent neighbor list update frequency $ps_{\text{freq}}=10$ and active boxes of size \SI{0.4}{m} $\times$ \SI{0.4}{m} $\times$ \SI{0.4}{m} associated with each of the four tires.

\begin{figure}[htbp]
    \centering
    \begin{subfigure}{\textwidth}
        \centering
        \includegraphics[width=1\textwidth]{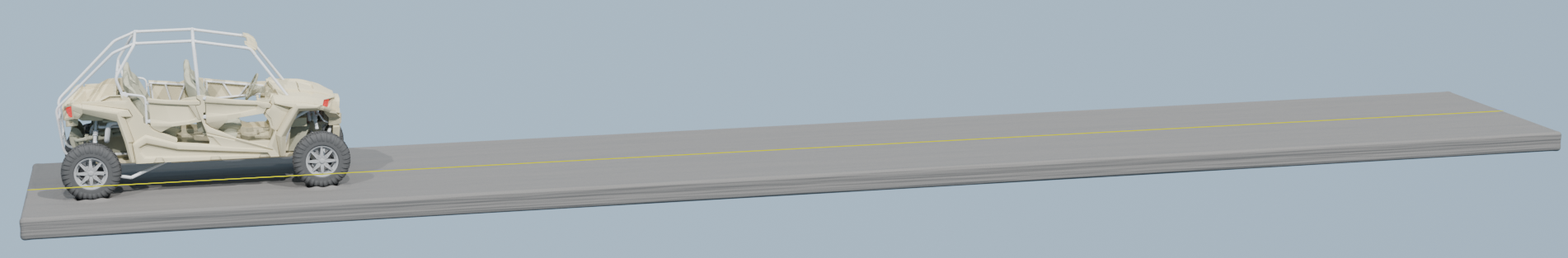} % Ensure this path is correct
        \caption{Rendering of the Polaris RZR vehicle on the \SI{20}{m} long granular terrain strip used in the scaling study. The yellow thin line represents the path followed by the vehicle.}
        \label{fig:scalability_render}
    \end{subfigure}
    \vspace{1em} % Add some vertical space between subfigures
    \begin{subfigure}{\textwidth}
        \centering
        \includegraphics[width=0.8\textwidth]{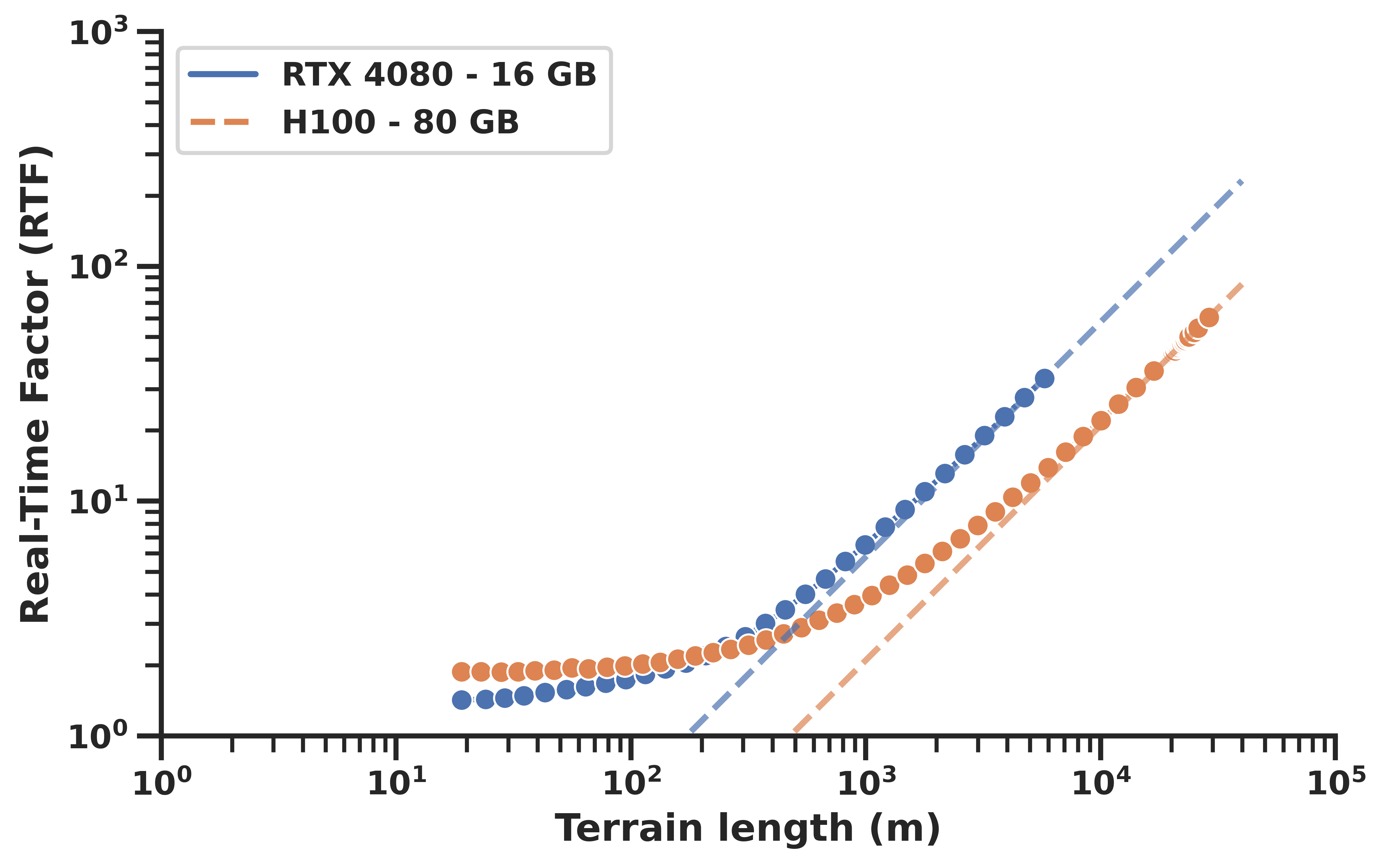} % Ensure this path is correct
        \caption{RTF as a function of terrain length for the Nvidia RTX 4080 Ti (16 GB) and Nvidia H100 (80 GB) GPUs.}
        \label{fig:scalability_plot}
    \end{subfigure}
    \caption{Scaling analysis of the Chrono::CRM solver using a Polaris RZR vehicle simulation. (a) Visual setup for the shortest terrain length. (b) Log-log plot of RTF versus terrain length, demonstrating the solver's performance scalability on different GPUs. The dashed tangent lines indicate linear scaling of RTF with terrain length in the respective operational ranges.}
    \label{fig:scalability}
\end{figure}

The scaling analysis focuses on varying the terrain length while keeping the terrain width fixed at \SI{3}{m} and depth at \SI{0.25}{m}. On an Nvidia RTX 4080 Ti GPU with 16 GB memory, the terrain length is increased from \SI{20}{m} up to \SI{6}{km}. The same experiment is repeated on an Nvidia H100 GPU with 80 GB memory, where the terrain length is scaled from \SI{20}{m} up to \SI{29}{km}. Figure~\ref{fig:scalability_render} shows a rendering of the simulation for the smallest terrain length. The RTF is measured as a function of terrain length for both GPUs, and the results are presented on a log-log plot in Fig.~\ref{fig:scalability_plot} to illustrate the scaling behavior.

The scaling results presented in Fig.~\ref{fig:scalability_plot} demonstrate favorable performance characteristics for Chrono::CRM. On both GPUs, the RTF exhibits sub-linear scaling with terrain length (and hence, number of SPH particles) over a significant range, as indicated by the dashed tangent lines on the log-log plot. This sub-linear scaling is maintained until memory capacity or other system limits are approached. Specifically, using the Nvidia RTX 4080 Ti, simulations remain practical for terrain lengths up to approximately \SI{5.9}{km}, beyond which the simulation runs out of GPU memory. The Nvidia H100, with its larger memory capacity, sustains this efficient scaling for much longer terrains, enabling simulations up to approximately \SI{28}{km} before out of memory errors occur. This indicates that Chrono::CRM can effectively handle large problem sizes, with performance primarily limited by available GPU memory for the tested configurations.

\section{Demonstration of Technology}
\label{sec:demOfTech}
This section demonstrates the versatility of Chrono::CRM through two representative real-world use cases: (1) vehicle performance analysis with flexible tires on deformable terrain, and (2) controller design for autonomous soil leveling (bulldozing). The examples presented are meant to showcase the software's capabilities across diverse application domains. Validating the simulations in this section falls outside the scope of this paper.

\subsection{Deformable Tires on Deformable Terrain}
An experiment was designed to evaluate the performance of a Polaris RZR model equipped with either rigid or flexible tires on both rigid and deformable terrains. The Polaris RZR was simulated using Chrono::Vehicle \cite{chronoVehicle2019}, with Chrono::FEA employed to model the flexible tire, modeled here with ANCF 3423 shell elements~\cite{mikeANCF-comparison2023}. The tires have a radius of \SI{0.33}{m} and a width of \SI{0.33}{m}; the rim radius is \SI{0.13}{m}. 

The FEA mesh for the flexible tires consists of 240 quad shell elements, arranged in a grid with 40 divisions in the circumferential direction and 6 divisions in the transversal direction, and using a representative tire profile (specified as a spline curve). Orthotropic material properties were assigned to the bead, sidewall, and tread of the flexible tire, as listed in Table~\ref{tab:tire_material}.
The rigid tire used a mesh collision geometry based on the same tire profile.

\begin{table}[htbp]
    \centering
    \caption{Orthotropic material properties for the Polaris flexible ANCF tire.}
    \label{tab:tire_material}
    \begin{tabular}{lccc}
    \hline
    \textbf{Property} & \textbf{Bead} & \textbf{Sidewall} & \textbf{Tread} \\
    \hline
    Density (kg/m$^3$) & 1083.33 & 1009.09 & 1151.59 \\
    \hline
    \multicolumn{4}{l}{\textbf{Young modulus (MPa)}} \\
    $E_1$ & 56.93 & 48.48 & 752.28 \\
    $E_2$ & 1299.50 & 6877.04 & 1842.43 \\
    $E_3$ & 56.93 & 48.48 & 81.98 \\
    \hline
    \multicolumn{4}{l}{\textbf{Poisson ratio}} \\
    $\nu_{12}$ & 0.0197 & 0.0032 & 0.0140 \\
    $\nu_{23}$ & 0.7416 & 0.4830 & 0.6983 \\
    $\nu_{31}$ & 0.4500 & 0.4500 & 0.6515 \\
    \hline
    \multicolumn{4}{l}{\textbf{Shear modulus (MPa)}} \\
    $G_{12}$ & 16.34 & 16.34 & 5397.15 \\
    $G_{23}$ & 16.34 & 16.34 & 16.34 \\
    $G_{31}$ & 16.34 & 16.34 & 16.34 \\
    \hline
    \multicolumn{4}{l}{\textbf{Section properties}} \\
    Transversal divisions & 1 & 2 & 3 \\
    Layer thickness (mm) & 9.0 & 3.1 & 8.1 \\
    \hline
    \end{tabular}
\end{table}

%\begin{figure}[htbp]
%    \centering
%    \includegraphics[width=0.4\textwidth]{figs/flex_tire_profile.jpg}
%    \caption{A rendering snapshot showcasing the side and front view's of the flexible tire modeled with ANCF shell elements.}
%    \label{fig:flex_tire_profile}
%\end{figure}

For simulations involving rigid tire-rigid terrain interaction, a smooth contact model was used with a friction coefficient of $0.9$, restitution coefficient of $0.2$, Young modulus of \SI{2e7}{Pa}, and Poisson ratio of $0.2$. The deformable terrain was modeled as CRM soil with friction coefficient $\mu_s = 0.9$, bulk density $\rho=$ \SI{1800}{kg/m^3}, and cohesion $c=$ \SI{1000}{Pa}. Soil particle parameters included an initial spacing $d_0=$ \SI{0.02}{m}, smoothing length $h=$ \SI{0.024}{m}, and artificial viscosity $\gamma_a=$ $0.5$.

The simulated vehicle traversed a terrain patch measuring \SI{50}{m} in length, \SI{3}{m} in width, and \SI{0.3}{m} in height. A controller managed the throttle input through distinct phases: a linear ramp-up during a \SI{1}{s} acceleration phase, followed by a \SI{1.5}{s} cruise phase at a constant \SI{80}{\%} maximum throttle. Finally, a \SI{1.5}{s} coast phase with zero throttle was followed by a braking phase at \SI{70}{\%} brake input. Steering was maintained at zero throughout all phases. The total simulation duration was \SI{5}{s}, with a time step of \SI{5e-4}{s}.

\subsubsection*{Results}
\begin{figure}[htbp]
    \centering
    \includegraphics[width=0.8\textwidth]{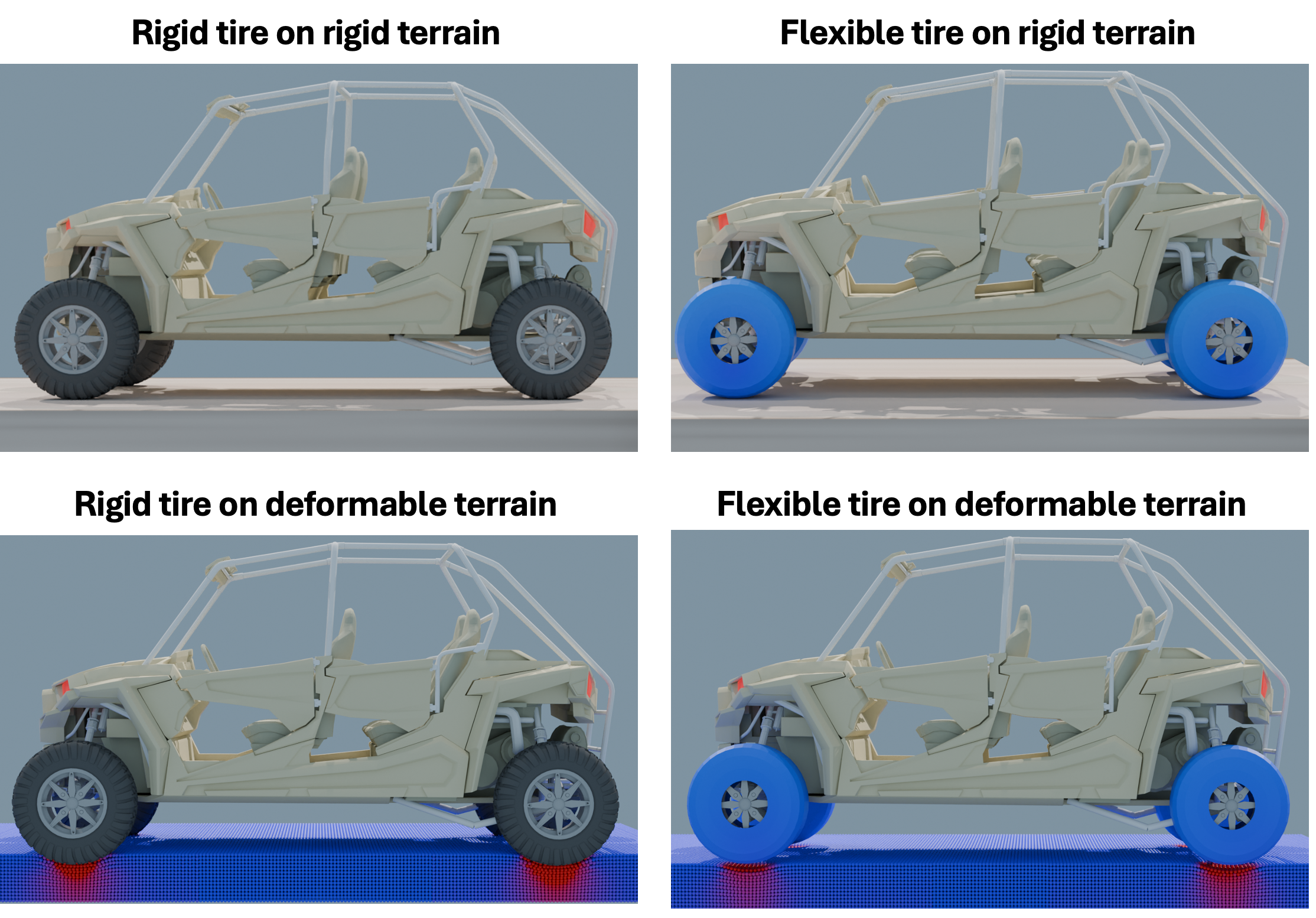}
    \caption{Illustration of the four simulated test scenarios for the Polaris RZR: (Top Left) Rigid tire on rigid terrain, (Top Right) Flexible tire on rigid terrain, (Bottom Left) Rigid tire on deformable terrain, and (Bottom Right) Flexible tire on deformable terrain. Deformable terrain is represented by SPH particles, with color indicating ground pressure (range shown: \SI{0}{Pa} (blue) to \SI{30000}{Pa} (red)). The visualization highlights higher ground pressure concentration under the rigid tire compared to the flexible tire on deformable terrain.}
    \label{fig:four_test_cases}
\end{figure}

\begin{figure}[htbp]
    \centering
    \includegraphics[width=0.8\textwidth]{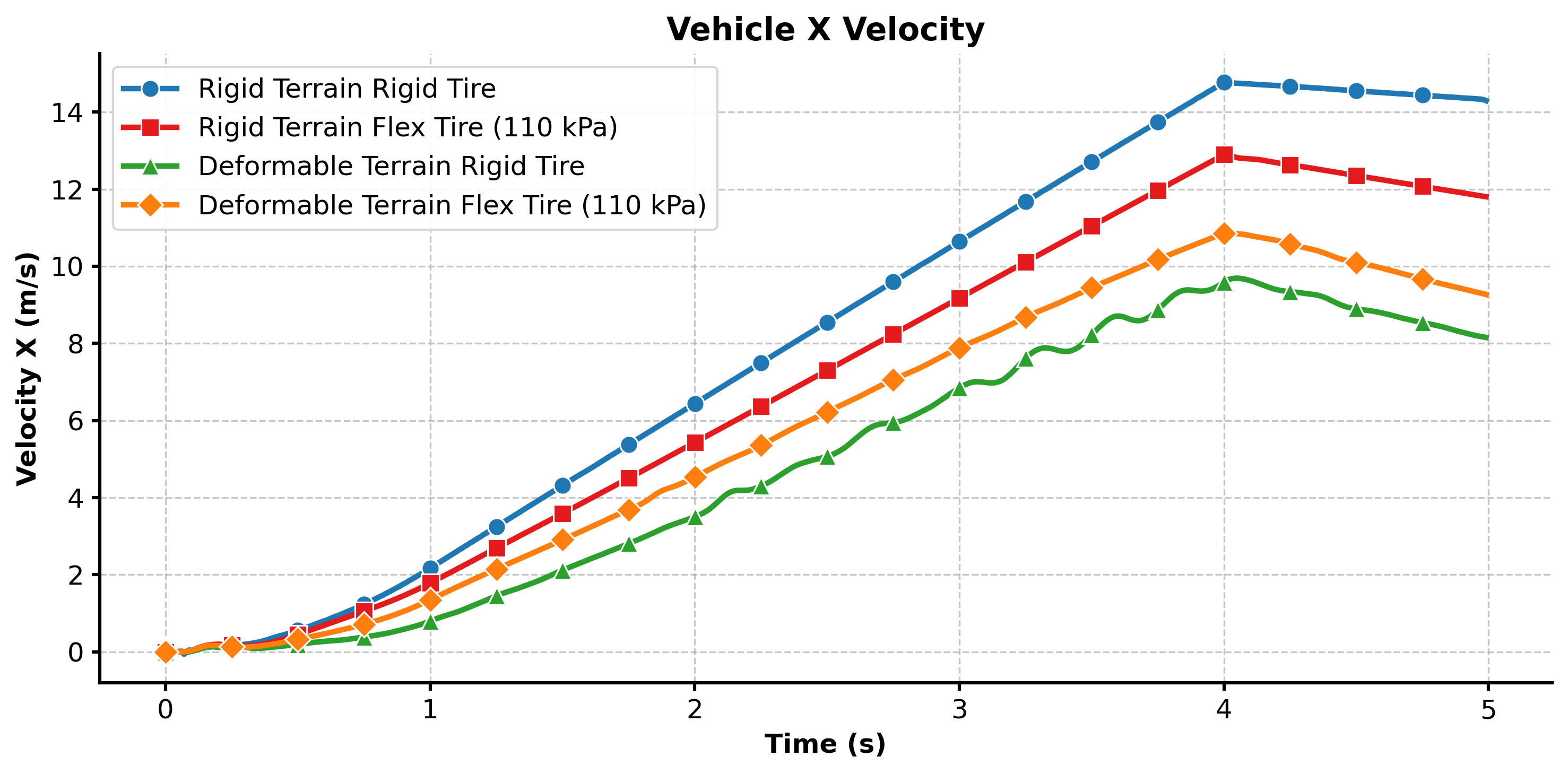}
    \caption{Comparison of vehicle longitudinal velocity over time for the four simulated scenarios: rigid tire on rigid terrain, flexible tire (\SI{110}{kPa}) on rigid terrain, rigid tire on deformable terrain, and flexible tire (\SI{110}{kPa}) on deformable terrain. The plot illustrates that on deformable terrain, the flexible tire achieves a higher peak velocity, while on rigid terrain, the rigid tire results in a higher peak velocity.}
    \label{fig:velocity_comparison}
\end{figure}

Vehicle performance was assessed under four distinct scenarios: (1) rigid tire on rigid terrain, (2) flexible tire on rigid terrain, (3) rigid tire on deformable terrain, and (4) flexible tire on deformable terrain (see Fig.~\ref{fig:four_test_cases}). The flexible tire was maintained at a nominal pressure of \SI{110}{kPa}, corresponding to the recommended inflation pressure for a Polaris RZR tire. The vehicle's longitudinal velocity (x-direction) was recorded for each scenario and is presented in Fig.~\ref{fig:velocity_comparison}.

The simulations demonstrate distinct performance characteristics depending on tire type and terrain. 
In simulations on deformable terrain, Fig.~\ref{fig:four_test_cases} shows that the flexible tires distributes ground pressure more broadly when compared to the concentrated pressure observed under the rigid tire. Quantitatively, this difference in interaction correlates with vehicle velocity (Fig.~\ref{fig:velocity_comparison}). The vehicle modeled with flexible tires achieved a higher maximum velocity. 
Conversely, on rigid terrain, the rigid tire configuration resulted in a greater maximum vehicle velocity. 
These findings highlight the significant interplay between tire flexibility and terrain characteristics in determining vehicle performance. 

Table~\ref{tab:demo1RTF} provides the RTF values for the two co-simulated systems (multibody including the vehicle and tires and, where applicable, CRM for the deformable terrain). We note that the CRM solver completes a step in a significantly faster time than the multibody system simulating a vehicle with flexible tires; this indicates that the CRM terrain simulation on the GPU occurs in the shadow of the Chrono FEA computations on the CPU. At approximately 500, the overall RTF for the case involving flexible tires on CRM deformable terrain is a reasonable value for such a complex simulation, with further improvements expected from planned development in the FEA time integrator.
%Second, we see that the overall RTF of simulating flexible tires on deformable terrain is still reasonable ($\approx 500$), indicating significant speed improvements in the Chrono ANCF solver since~\cite{antonioVehicleTireGranMatSim2017}.

\begin{table}[htbp]
    \centering
    \caption{Subsystem RTF values for the four simulated scenarios.}
    \label{tab:demo1RTF}
    \begin{tabular}{l|rr}
    \hline
    \multirow{2}{*}{\textbf{Scenario}} & \multicolumn{2}{c}{\textbf{RTF}} \\\cline{2-3} 
    & \textbf{Multibody} & \textbf{CRM} \\
    \hline
    Rigid tire on rigid terrain & 0.83 & - \\
    Flexible tire on rigid terrain & 465.3 & - \\
    Rigid tire on deformable terrain & 1.13 & 9.61 \\
    Flexible tire on deformable terrain & 463.2 & 14.61 \\
    \hline
    \end{tabular}
\end{table}

% ==================================================

\subsection{Controller design for autonomous ground leveling}
\label{sec:controllerDesign}
Autonomous ground leveling or grading involves self-operating machinery to achieve required earth-surface contours without direct human intervention. Here, we describe the use of Chrono::CRM to design a control policy to autonomously level a pile of soil using a Chrono model of a Gator vehicle~\cite{Harry2025leveling}. We attach a blade at the front of the Gator to give it the ability to push and level the soil (see Fig.~\ref{fig:gator_blade_motion}). The blade's vertical displacement $d$ and pitch angle $p$ are controlled, while the throttle is kept constant at full throttle. 
%Figure~\ref{fig:gator_combined} provides a rendering of the Gator model with the blade attached and a schematic showing the control inputs. 
%
%\begin{figure}[htbp]
%	\centering
%	\includegraphics[width=0.40\textwidth]{figs/only_gator.png}
%	\includegraphics[width=0.58\textwidth]{figs/bulldozing.png}
%	\caption{Left: Rendering of the Gator model with the blade attached. Right: Schematic showing the control inputs. The pitch $p$ and vertical displacement $d$ of the blade are controlled while the throttle is kept constant at full throttle.}
%	\label{fig:gator_combined}
%\end{figure}

The goal is to design a controller, that, given initial and desired terrain heightmaps, will generate the necessary control inputs to autonomously level the soil and achieve the desired heightmap. 
This is done using a learned model of the input-output mapping, where the inputs are an initial terrain heightmap and a set of controls and the output is the resulting heightmap. This model, whose structure is fully described in~\cite{Harry2025leveling}, is trained using supervised learning on a dataset generated from 1400 Chrono::CRM simulations.
In each simulation, control inputs are randomized by drawing from uniform distributions: $p \sim U(0, \pi/6)$ and $d \sim U(-0.1, 0.05)$. The initial pile height is set to either \SI{0.3}{m} (for 700 simulations) or \SI{0.4}{m} (for the remaining 700). These simulations are conducted on a terrain patch measuring \SI{10}{m} in length and \SI{4}{m} in width. The initial spread of the pile follows a Gaussian distribution with a standard deviation of \SI{0.5}{m}. 
Soil properties are defined by a bulk density \SI{1800}{kg/m^3}, friction coefficient \SI{0.8}{}, and cohesion \SI{5000}{Pa}. For the SPH parameters, we use an initial particle spacing $d_0 = \SI{0.02}{m}$, a smoothing length $h = \SI{0.024}{m}$, and an artificial viscosity $\gamma_a = 0.5$. Each simulation scenario runs for \SI{6}{s} with a time step $\Delta t = \SI{5e-4}{s}$, and data is logged at \SI{20}{Hz}. Generating this complete dataset required approximately 60 hours on an Nvidia A100 GPU, with individual simulations running at an RTF of about 25. In every simulation, the Gator controlled by the action vector $u$, transforms the $h_{\text{initial}}$ heightmap into the $h_{\text{crm}}$ heightmap. The triplet $(h_{\text{initial}}, u, h_{\text{crm}})$ constitutes a labeled sample for training the ML model.

Once the ML-based model is obtained, it is subsequently used ``in the loop'' with  a gradient based trajectory optimization algorithm~\cite{Sukhija2023} to generate the control inputs to autonomously level the soil for a scenario from the training data set simulated with Chrono::CRM. See~\cite{Harry2025leveling} for complete details on the design and implementation of the controller.

\subsubsection*{Results}
\begin{figure}[htbp]
    \centering
    \includegraphics[width=0.8\textwidth]{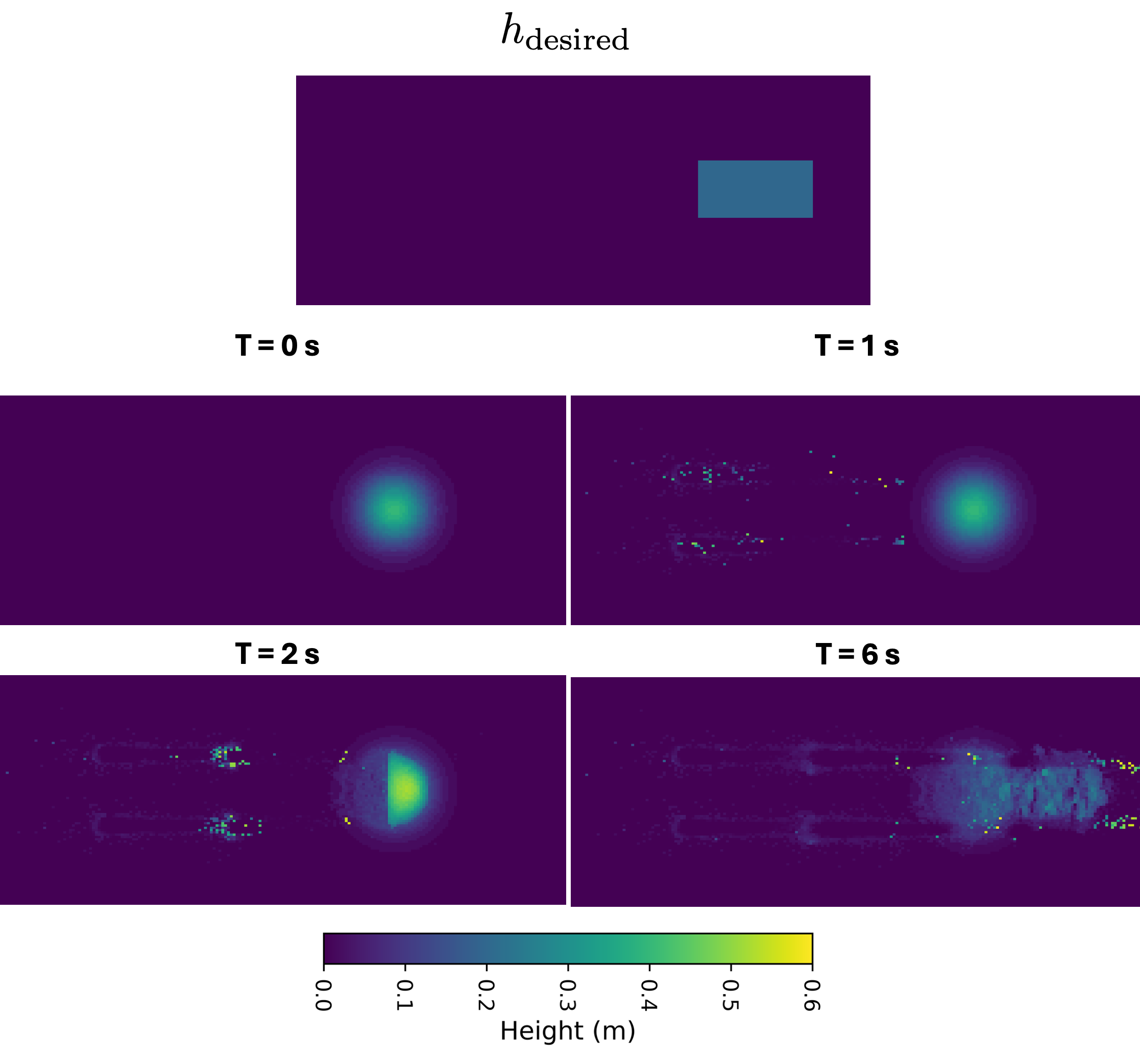}
    \caption{Progression of soil leveling towards a desired state ($h_{\text{desired}}$). The top panel shows the target heightmap. The subsequent panels illustrate the initial heightmap at $T=\SI{0}{s}$ and the evolving actual heightmap at $T=\SI{1}{s}$, $T=\SI{2}{s}$, and $T=\SI{6}{s}$, as achieved by the autonomous controller using the L2 model.}
    \label{fig:height_maps}
\end{figure}

\begin{figure}[htbp]
    \centering
    \includegraphics[width=0.8\textwidth]{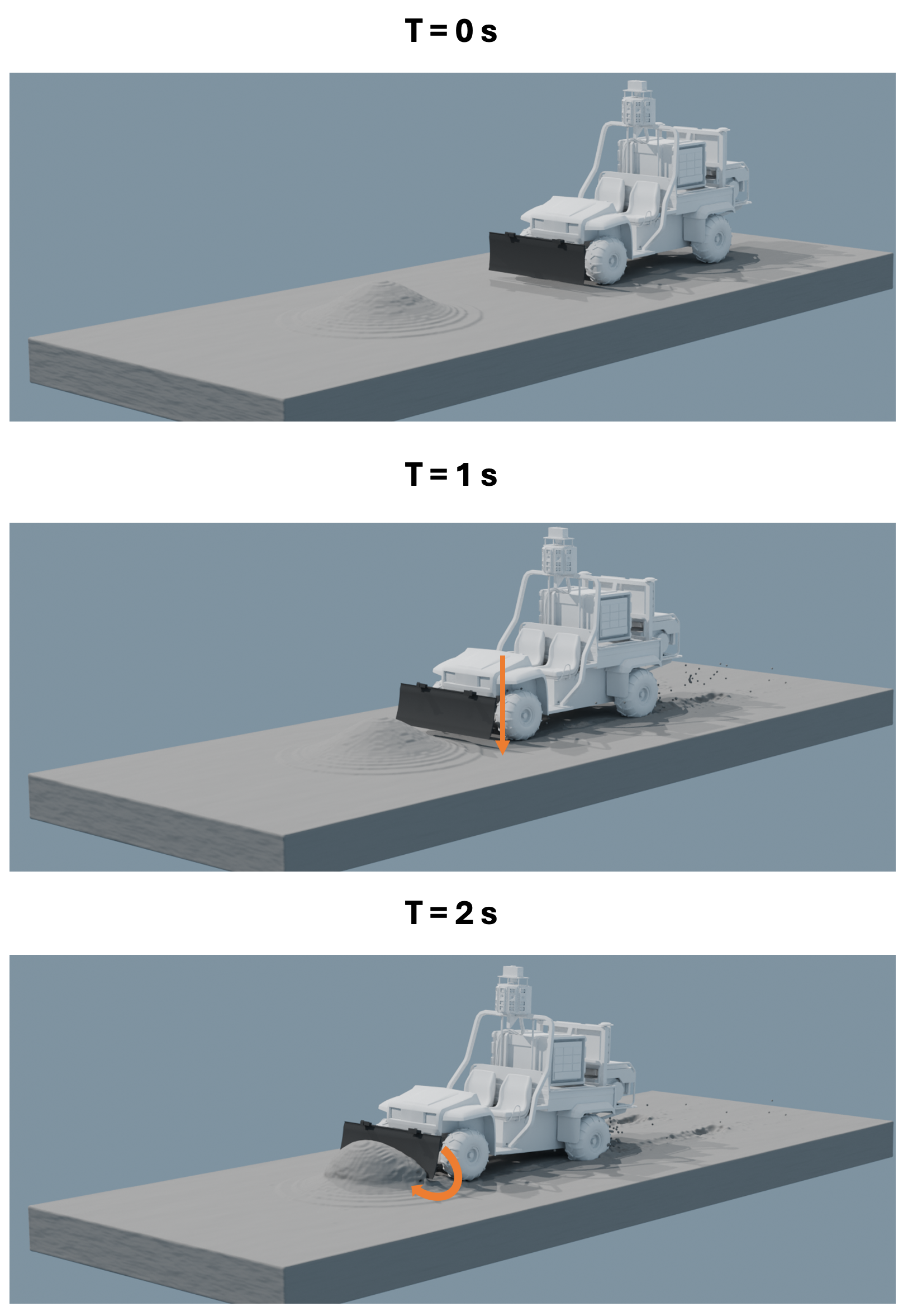}
    \caption{Renderings of the Gator vehicle autonomously leveling a soil pile at different time instances. (Top) Initial state at $T=\SI{0}{s}$. (Middle) At $T=\SI{1}{s}$, the controller has lowered the blade to engage the soil. (Bottom) At $T=\SI{2}{s}$, the controller has rotated the blade upwards, pushing and reshaping the pile. Corresponding heightmaps are shown in Fig.~\ref{fig:height_maps}.}
    \label{fig:gator_blade_motion}
\end{figure}

\begin{figure}[htbp]
    \centering
    \includegraphics[width=0.6\textwidth]{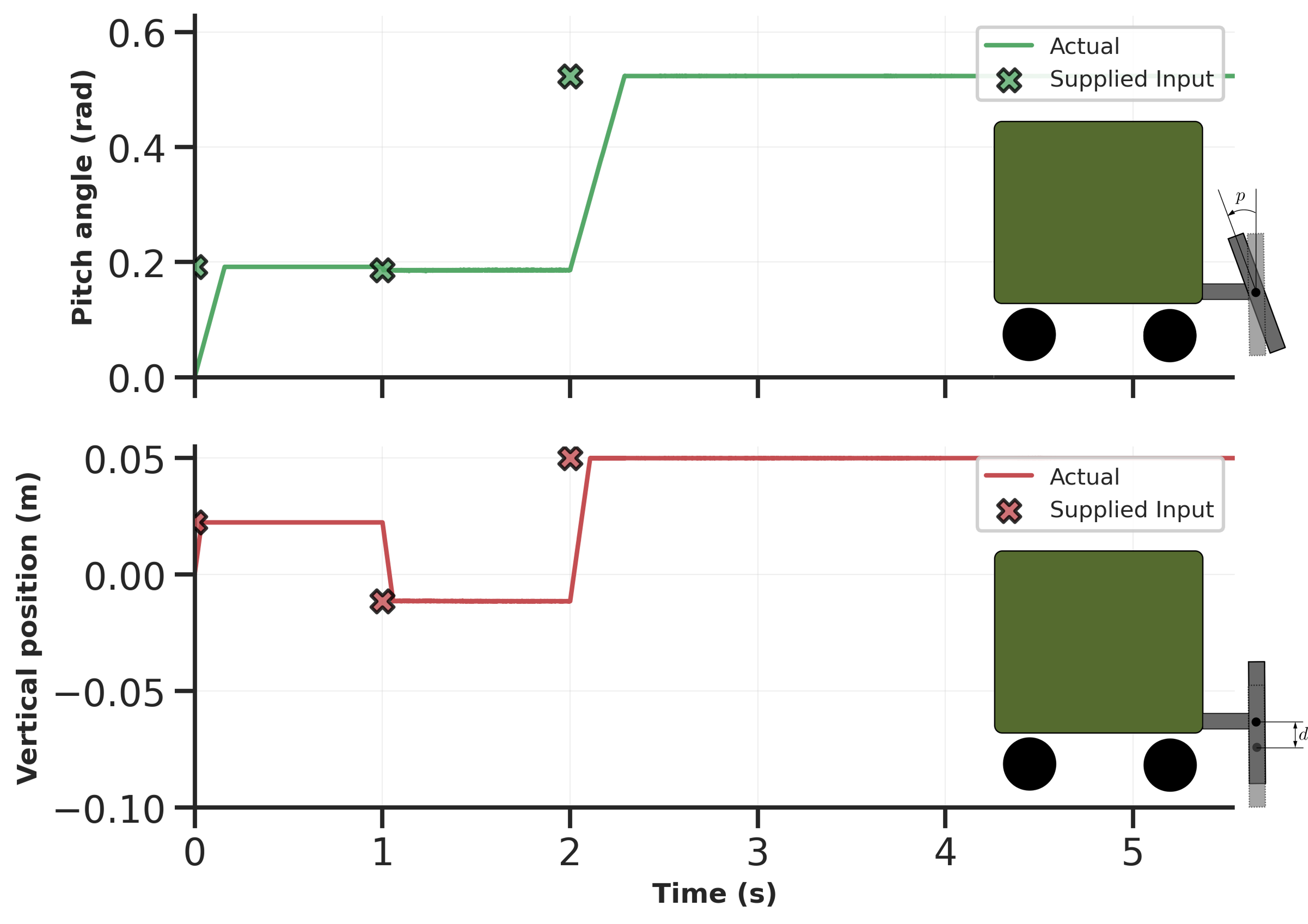}
    \caption{Time evolution of the Gator blade's state during the autonomous leveling task. The plots compare the desired control inputs (pitch angle $p$ and vertical displacement $d$) generated by the controller with the actual achieved states of the blade. These inputs correspond to the vehicle actions shown in Fig.~\ref{fig:gator_blade_motion}.}
    \label{fig:inputs_outputs}
\end{figure}

As a proof of concept, the ML-based model integrated within the gradient-based controller is used to autonomously level a soil pile, starting from an initial heightmap selected from the training dataset. Figure~\ref{fig:height_maps} (top panel) illustrates the desired, flat target heightmap ($h_{\text{desired}}$) alongside the initial state of the soil at $T=\SI{0}{s}$ (second panel, leftmost image). The controller then generates a sequence of actions (blade pitch and vertical displacement), detailed in Fig.~\ref{fig:inputs_outputs}, which plots both the commanded inputs and the blade's achieved states over time. The vehicle's corresponding physical interaction with the soil is visualized in Fig.~\ref{fig:gator_blade_motion}. This figure shows the blade initially lowering to engage the pile around $T=\SI{1}{s}$. Subsequently, around $T=\SI{2}{s}$, the controller commands an upward rotation of the blade, causing it to cut into and redistribute the soil. The resulting progressive change in the terrain is further shown by the sequence of actual heightmaps in Fig.~\ref{fig:height_maps} (panels for $T=\SI{1}{s}$, $T=\SI{2}{s}$, and $T=\SI{6}{s}$), demonstrating the bulldozer's ability to reshape the initial pile towards the desired level profile.

\section{Conclusions}
\label{sec:conclusion}
This paper discusses Chrono::CRM, a general, scalable, and efficient simulation framework for robots and vehicles interacting with deformable terrains. Built upon a continuum model for granular media and tightly integrated with the Chrono multibody dynamics engine via the FSI framework, the SPH-based Chrono::CRM solver provides significant gains in simulation size and speed relative to the initial version detailed in~\cite{weiGranularSPH2021}.

We demonstrated Chrono::CRM's capabilities through a suite of validation tests and benchmarks. The solver's accuracy was established by comparing simulation results against experimental data for sphere cratering, cone penetration, and single-wheel rover tests. Furthermore, digging torque simulations for a RASSOR drum showed favorable agreement with high-fidelity DEM simulations, highlighting Chrono::CRM's utility for complex excavation tasks not amenable to semi-empirical models.

A key focus of this work was enhancing computational performance without compromising physical fidelity. Novel solver optimizations, including the implementation of persistent neighbor lists with controlled update frequencies and an ``active domains'' technique, were introduced. These optimizations were shown to yield good speedups -- up to 3$\times$ from persistent neighbor lists and over 2$\times$ from active domains in relevant test cases -- resulting in overall performance improvements exceeding one order of magnitude (e.g., up to 15.9$\times$ for the tracked vehicle benchmark) compared to previous SPH implementations. These performance gains were achieved with no discernible loss in simulation accuracy across all validated scenarios. Consequently, Chrono::CRM can simulate large-scale vehicle and robotic systems interacting with extensive deformable terrains at speeds comparable to, and in some cases exceeding, semi-empirical methods like Chrono::SCM, while offering a much richer, physics-based, representation of soil behavior.

The scalability of Chrono::CRM was further showcased through simulations involving terrain lengths up to \SI{5.9}{km} on consumer-grade GPUs and up to \SI{28}{km} on high-performance GPUs like the NVIDIA H100, demonstrating favorable scaling of computational cost with problem size until hardware memory limits are reached. This makes high-fidelity, large-scale off-road simulations practically feasible.

Finally, the versatility of Chrono::CRM was illustrated through two distinct case studies: a comparative performance analysis of a Polaris RZR with rigid and flexible tires on deformable terrain, and the design of a control policy for an autonomous Gator undertaking a soil leveling task, where the controller was trained using data generated by Chrono::CRM. These demonstrations underscore the framework's potential to address a wide array of engineering challenges in vehicle design, robotics, and autonomous systems development.

The Chrono::CRM software is released as open source, as a component of the Chrono multi-physics simulator. Future work may involve further validation of the model, extending the material models employed, exploring advanced rendering for real-time visualization of large-scale SPH simulations, and expanding the FSI framework to accommodate more complex multi-physics interactions. The advancements presented herein pave the way for physics-based, efficient, and large-scale simulations of dynamic systems interacting with deformable granular environments.

\section*{Acknowledgements}
We would like to thank Drs. Wei Hu and Lijing Yang for their early contributions to the FSI and CRM Chrono modules. Funding support for this work was provided in part by the National Science Foundation (NSF) under awards CMMI2153855 and OAC2209791.

\printcredits
\section*{Declaration of generative AI and AI-assisted technologies in the writing process}

During the preparation of this work the authors used ChatGPT and Gemini in order to correct grammar and improve readability. After using this tool/service, the authors reviewed and edited the content as needed and take full responsibility for the content of the publication.

%% Using BibTeX for bibliography
\bibliographystyle{cas-model2-names}
\bibliography{all_refs}

\end{document}